\documentclass[preprint,3p,12pt]{elsarticle}
\usepackage{lineno,hyperref}
\usepackage{amsmath,amssymb,epsfig}
\usepackage[utf8]{inputenc}
\usepackage[english]{babel}
\usepackage{mathtools}
\usepackage{helvet}         % selects Helvetica as sans-serif font
\usepackage{courier,stix}        % selects Courier as typewriter font
\usepackage{type1cm}
\usepackage{csquotes}
\usepackage{makeidx}        % allows index generation
\usepackage{graphicx}
\usepackage{multicol}
\usepackage{appendix}
\usepackage{stix}
\usepackage[bottom]{footmisc}% places footnotes at page bottom
\usepackage[dvipsnames]{xcolor}
\usepackage{multicol}
\usepackage[bottom]{footmisc}     
\usepackage{hyperref} 
\usepackage{tikz}
\usepackage{float}  
\usepackage{helvet,overpic} 
\usepackage{subfig}% selects Helvetica as sans-serif font
\usepackage{soul}
\usepackage{amsthm}

\newtheorem{rmk}{Remark}

\newtheorem{assumption}{Assumption}

\usepackage{nomencl}
\usepackage[linesnumbered,ruled,vlined]{algorithm2e}

\makenomenclature

\newcommand{\lineseg}{\vert}

\newcommand{\rectangleblack}{\scalebox{0.5}{\ensuremath \hrectangleblack}}
\begin{document}

\begin{frontmatter}

\title{Singularity Distance Computations for 3-RPR Manipulators Using Intrinsic Metrics}

\author{Aditya Kapilavai and Georg Nawratil}
\address{Institute of Discrete Mathematics and Geometry \& 
Center for Geometry and Computational Design, TU Wien\\
Wiedner Hauptstrasse 8--10, Vienna 1040, Austria.}
%\address{}


%\author[mysecondaryaddress]{Vienna School of Mathematics, Vienna}
\cortext[mycorrespondingauthor]{Corresponding author}
%\ead{akapilavai,nawratil@geometrie.tuwien.ac.at}

\ead{\{akapilavai, nawratil\}@geometrie.tuwien.ac.at}
 
\begin{abstract}
We present an efficient algorithm for computing the closest singular configuration to each non-singular pose of a 3-RPR planar manipulator performing a 1-parametric motion. 
By considering a 3-RPR manipulator as a planar framework,  one can use methods from rigidity theory to compute the singularity distance with respect to an intrinsic metric. 
Such a metric has the advantage over any performance index used for indicating the closeness to singularities, that the obtained value is a distance, which equals the radius of a guaranteed singularity-free sphere in the joint space of the manipulator. 
The proposed method can take different design options into account as the platform/base can be seen as a triangular plate or as a pin-jointed triangular bar structure. 
Moreover, we also allow the additional possibility of pinning down the base/platform triangle to the fixed/moving system thus it cannot be deformed.  
For the resulting nine interpretations, we compute the corresponding intrinsic metrics based on the total elastic strain energy density of the framework using the physical concept of Green-Lagrange strain. The global optimization problem of finding the closest singular configuration with respect to these metrics is solved by using tools from numerical algebraic geometry. The proposed algorithm is demonstrated based on an example, which is also used to compare the obtained intrinsic singularity distances  with the corresponding extrinsic ones. 

\end{abstract}

\end{frontmatter}

\section{Introduction}
\begin{figure}
\centering
\begin{overpic}[width=75mm]{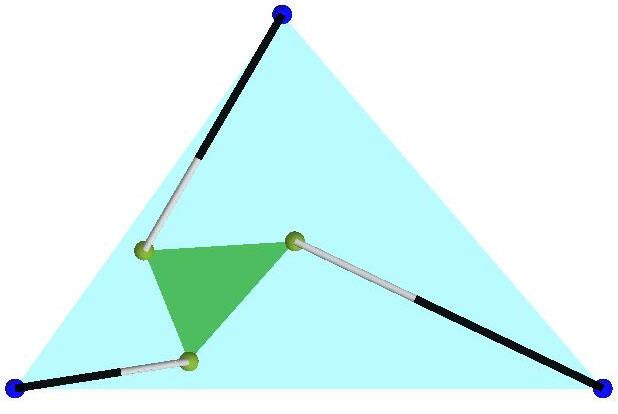}
\put (32.5,6){$\mathbf{k}_{4}$}
\put (-4,2){$\mathbf{k}_{1}$}
\put (-1,6.5){R}
\put (10,8){P}
\put (28,11){R}
\put (100,2){$\mathbf{k}_{2}$}
\put (50,28){$\mathbf{k}_{5}$}
\put (17,23){$\mathbf{k}_{6}$}
\put (45,67){$\mathbf{k}_{3}$}
\end {overpic}
\caption{3-RPR planar parallel manipulator with base anchor points $\mathbf{k}_1,\mathbf{k}_2,\mathbf{k}_3$ and platform anchor points 
$\mathbf{k}_4,\mathbf{k}_5,\mathbf{k}_6$, where $\mathbf{k}_{i}$ 
denotes the coordinate vector of the $i$th attachment with respect to the 
fixed frame.}
\label{3rpr}
\end{figure}
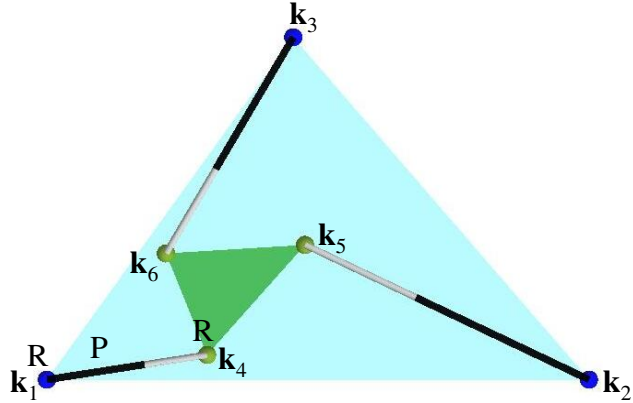

A 3-RPR manipulator (cf. Fig.~\ref{3rpr}) is a three-degree-of-freedom (dof) planar parallel manipulator with two translational dofs and one rotational dof. The base and platform are connected by three legs, where each leg consists of two passive revolute (R) joints connected by an actuated prismatic (P) joint.  
Let $\mathbf{k}_i$ denote the 
coordinate vectors of the base anchor points with respect 
to the fixed frame with coordinates $(x_i,y_i)^T$
for $i=1,2 ,3$. 
The coordinate vectors of the platform anchor points with respect to the moving frame are labeled by  $\mathbf{p}_{j}$ with coordinates $(x_j,y_j)^T$ 
for $j=4,5,6$. Their coordinate vectors with respect to the fixed frame are computed as:
\begin{align}
\mathbf{k}_{j} :=
\mathbf{R} \mathbf{p}_j+ \mathbf{t},
\label{initial}
\end{align}
\noindent where $\mathbf{R}$ is a $2\times2$ rotation matrix and $\mathbf{t}$ is the translation vector. 

Without loss of generality, we can assume that $\mathbf{k}_1$ equals the origin of the fixed frame ($\Rightarrow x_1=y_1=0$) and that  $\mathbf{k}_2$ is located on its $x$-axis ($\Rightarrow y_2=0$). The same assumptions can be made for the moving frame which implies $x_4=y_4=y_5=0$. 
The remaining six coordinates $x_2,x_3,y_3,x_5,x_6,y_6$ can be seen as the design parameters of the base and platform, respectively. Moreover, we assume that the anchor points of the platform as well as the base are not collinear; i.e.\ $y_3y_6\neq 0$. We will comment on the excluded special designs with collinear anchor points in Section \ref{sec:collpb}.

It is well-known that from a line-geometric point of view, a 3-RPR manipulator is in a singular (also known as shaky or infinitesimal flexible) configuration if and only if the carrier lines of the three legs intersect at a common point or are parallel. This condition is equivalent to the fact that the Pl\"ucker coordinates of these lines are linearly dependent, leading to the algebraic characterization in the form of the 
so-called {\it singularity variety} $V={0}$ with 
 \begin{equation}\label{variety}
V = \det\begin{pmatrix}
 \mathbf{k}'_{4}-\mathbf{k}'_{1} & \phantom{-}\mathbf{k}'_{5}-\mathbf{k}'_{2} & \phantom{-}\mathbf{k}'_{6}-\mathbf{k}'_{3} \\ 
\det\left(\mathbf{k}'_{1},\mathbf{k}'_{4}-\mathbf{k}'_{1}\right)& \phantom{-}\det\left(\mathbf{k}'_{2}, \mathbf{k}'_{4}-\mathbf{k}'_{2}\right)& \phantom{-}\det\left(\mathbf{k}'_{3}, \mathbf{k}'_{4}- \mathbf{k}'_{3}\right) 
 \end{pmatrix}.
\end{equation}
The {\it singularity polynomial} $V$ is of degree 4 in the coordinates  $(c_i,d_i)$ of the points $\mathbf{k}_i'$ with respect to the fixed frame.

In singular configurations, the manipulator has at least one uncontrollable instantaneous dof, i.e.\ there is an instantaneous motion of the platform while all P joints are locked. Consequently, minor variations in the manipulator geometry (e.g.\ backlash in revolute joints or uncertainties in the actuation of P joints) 
can significantly affect the resulting manipulator configuration. Another phenomenon that also arises close to singularities is that the prismatic actuator forces can become very large, causing a breakdown of the manipulator \cite{hubert}. Therefore, it is crucial to avoid singularities and their vicinity. 
This is the primary motivation to study the singularity distance computation of 3-RPR robots. Further applications are pointed out in Section \ref{end}.

\subsection{Terminology and notations used for 3-RPR manipulator as frameworks}\label{terminology}
According to~\cite{nawratil2020snappability}, a configuration of a 3-RPR robot can be considered as a planar isostatic framework in the Euclidean plane $\mathbb{E}^{2}$. In this representation, the legs are modeled as bars $\left(\vert\right)$  and the platform/base either as a triangular plate ($\blacktriangle$) or a pin-jointed triangular bar structure ($\vartriangle$). Note that the interpretation of the base/platform 
as a triangular bar structure  ($\vartriangle$) gives the framework
a further possibility of being shaky beside the linear dependence of the three legs, namely if the three base/platform points are collinear. 

From a more abstract point of view, our planar frameworks are composed of a knot set $\mathcal{K}$, where each knot represents either a point or a triangular plate, and a graph $G$ on $\mathcal{K}$. Edges connecting knots of the graph correspond to bars and fix the combinatorial structure of the framework $G(\mathcal{K})$.  
The inner metric of $G(\mathcal{K})$ is determined by assigning lengths to the bars and shapes to the triangular plates, where the latter is equivalent to the assignment of lengths to the three sides of the triangular plate due to the side-side-side theorem. Therefore, the inner metric of all frameworks can be expressed by the 9-dimensional vector  $\mathbf{L}$ containing the distances $\ell_{ij}$ between the end-points $\mathbf{k}_i$ and $\mathbf{k}_j$ of the bars $\vert_{ij}$ with $i<j$ and $(i,j)\in\{(1, 2), (2, 3), (1, 3), (1, 4), (2, 5), (3, 6), (4, 5), (5, 6), (4, 6)\}$.
In general, there exist several embeddings of  $G(\mathcal{K})$ with inner metric $\mathbf{L}$ into $\mathbb{E}^{2}$, which are called realizations and are denoted by $G(\mathbf{K}_1),G(\mathbf{K}_2), \ldots$. 

Let us assume that $G(\mathbf{K})$ with $\mathbf{K}:=(\mathbf{k}_1,\ldots ,\mathbf{k}_6)$ corresponds to our given non-singular 
3-RPR configuration. Now we want to deform $G(\mathbf{K})$ into $G(\mathbf{K}')$ by a continuous increasing deviation from $\mathbf{L}$ 
in a way that $G(\mathbf{K}')$ is singular and that $\mathbf{L}'$ is as close as possible to $\mathbf{L}$, where the difference in the inner metric is measured on basis of the total elastic strain energy density of the framework (cf. Section~ \ref{intrinsic}). Therefore, 
we speak of a singularity distance computation with respect to an intrinsic metric. In contrast, if the singularity distance computation is based on the minimization of $\mathbf{K}'$ to $\mathbf{K}$ with respect to metrics induced by the one of the embedding space $\mathbb{E}^{2}$, we speak of extrinsic singularity distances, which are studied in detail in~ \cite{nawratil2019} and \cite{akapilavai2022}. 

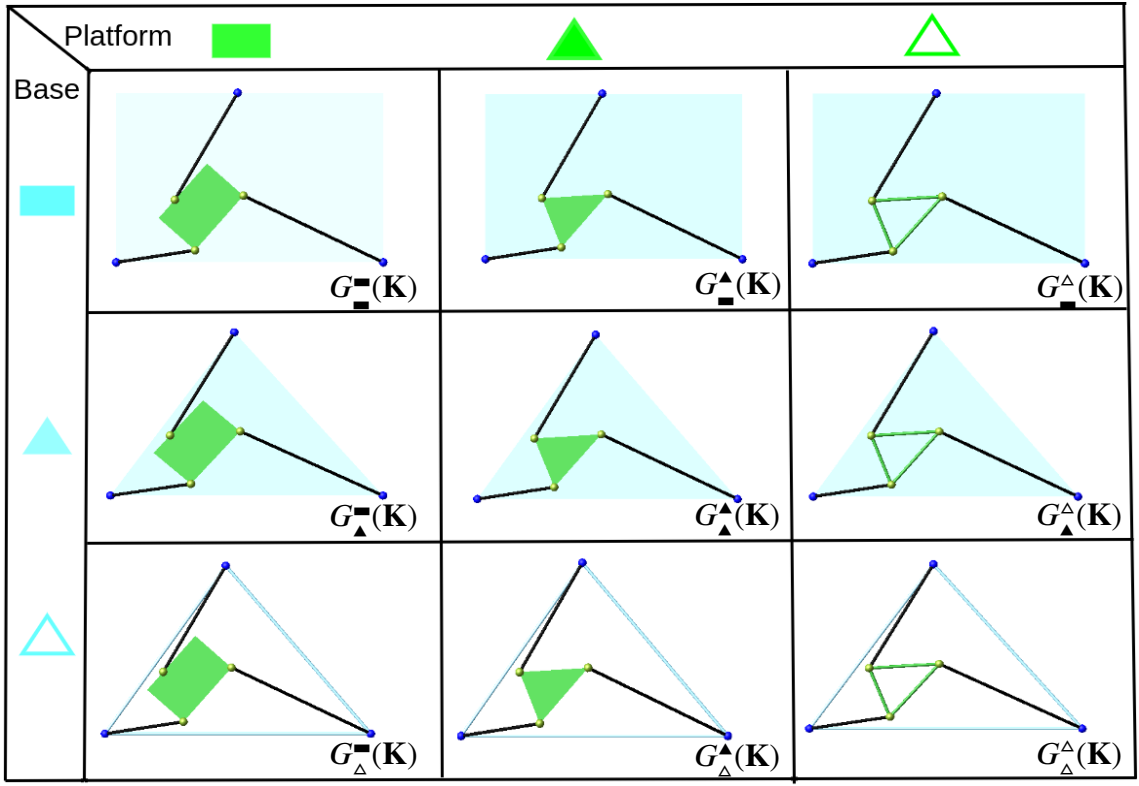
\begin{figure}[t]
    \begin{center}
       \begin{overpic}[width=15cm]{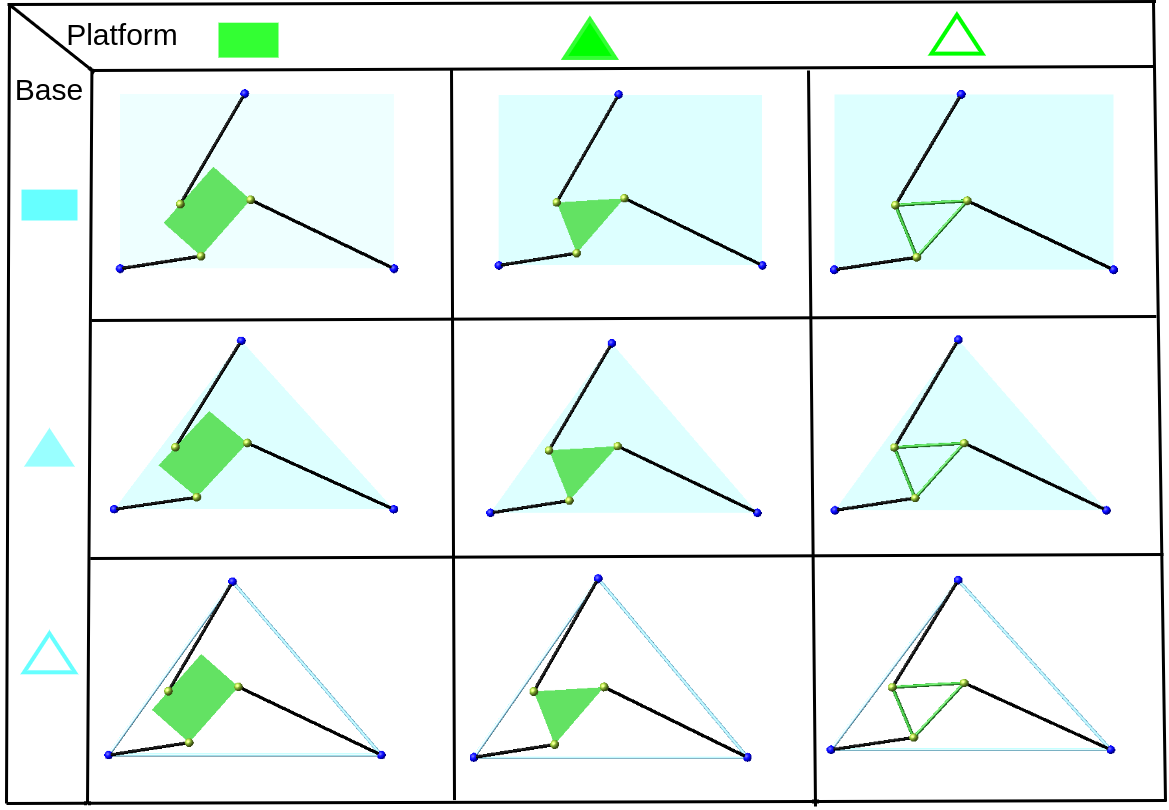}
       \put(29,43.2){$G_{\rectangleblack}^{\rectangleblack}(\mathbf{K})$}
         \put(61,43.5){$G^{\blacktriangle}_{\rectangleblack} (\mathbf{K})$}
          \put(91,43.2){$G^{\vartriangle}_{\rectangleblack} (\mathbf{K})$}
           \put(29,23){$G_{\blacktriangle}^{\rectangleblack} (\mathbf{K})$}
           \put(61,23.1){$G^{\blacktriangle}_{\blacktriangle} (\mathbf{K})$}
              \put(91,23.1){$G_{\blacktriangle}^{\vartriangle} (\mathbf{K})$}
              \put(29,2){$G_{\vartriangle}^{\rectangleblack}(\mathbf{K})$}
            \put(61,1.8){$G^{\blacktriangle}_{\vartriangle} (\mathbf{K})$}
            \put(91,2){$G^{\vartriangle}_{\vartriangle} (\mathbf{K})$}
        \end{overpic}
      \caption{Illustration of the nine considered interpretations of a 3-RPR  manipulator as planar frameworks, which are 
      arranged in a $3\times 3$ matrix. The latter is symmetric with respect to the exchange of the platform and the base, which can be achieved by considering the inverse motion of the manipulator.}
      \label{inter}
    \end{center}
\end{figure}

In the physical world, the change of the inner metric is realized by deformable material. We also take the option into account that the base or platform is made of an undeformable material, which is indicated by the symbol 
$\hrectangleblack$. 
Thus we end up with three possibilities ($\blacktriangle, \vartriangle, \hrectangleblack$)
for the platform/base, which results in a total of the following nine interpretations displayed in Fig.~\ref{inter}:
\begin{equation}\label{nine_int}
G_{\rectangleblack}^{\rectangleblack}(\mathbf{K}), G_{\rectangleblack}^{\blacktriangle}(\mathbf{K}), G_{\blacktriangle}^{\rectangleblack}(\mathbf{K}),
G^{\vartriangle}_{\rectangleblack} (\mathbf{K}),   
G^{\rectangleblack}_{\vartriangle} (\mathbf{K}),
G_{\blacktriangle}^{\blacktriangle}(\mathbf{K}), G_{\blacktriangle}^{\vartriangle}(\mathbf{K}), G_{\vartriangle}^{\blacktriangle}(\mathbf{K}),  G_{\vartriangle}^{\vartriangle}(\mathbf{K}). 
\end{equation}

\subsection{Review on intrinsic singularity distance computations}\label{sec:review}

In \cite{zein}, a procedure is presented for the determination of a maximal singularity-free cube in the joint space centered in the point, which corresponds to the given non-singular configuration. However, half of the edge length $e$ of this cube (which equals the infinity norm in the joint space) is not very well suited as a closeness index due to the fact that the mapping from the configuration space to the joint space is 6 to 1 (cf.\ \cite{husty3rpr}).  In general, not all configurations, which correspond to a point on the singularity variety in the joint space are singular. Therefore, it can be the case that even in a non-singular configuration $e$ equals zero as already pointed out by the second author in~\cite{nawratil2019}. 
 Note that this index belongs to the interpretation 
$G_{\rectangleblack}^{\rectangleblack}(\mathbf{K})$.
For this interpretation (applied to the spatial analog of the 3-RPR manipulator) two further intrinsic metrics were presented by the second author \cite[Eqs.\ (40) and (44)]{NAWRATIL2022104510}. 

For the interpretations $G_{\rectangleblack}^{\blacktriangle}(\mathbf{K})$ and 
$G_{\rectangleblack}^{\vartriangle}(\mathbf{K})$
intrinsic metrics for 3-RPR manipulators were already presented by the second author in \cite{nawratil2020snappability}. In the paper at hand we use the metric approach of \cite{nawratil2020snappability} and \cite{NAWRATIL2022104510}, which is adapted and extended to the other interpretations of Fig.~\ref{inter}. Our primary focus is on the algorithmic computation of these 
singularity distances along a 1-dimensional motion performed by the platform of the 3-RPR robot. 
In detail the structure of the paper is as follows:
Section \ref{intrinsic}, summarizes the results of \cite{nawratil2020snappability} and presents for each of the nine interpretations given in Eq.\ (\ref{nine_int}) a corresponding intrinsic metric. 
In Section \ref{generalcase} we formulate the constrained optimization problems for computing the resulting nine singularity distances including the closest singular configurations.  
Section \ref{results} is devoted to the computational pipeline for solving these optimization problems, which is validated on the basis of a concrete example in Section \ref{sec:example}. The latter section also contains a comparison of (a) the presented method with already existing intrinsic metrics and (b) the presented intrinsic singularity distances with the corresponding extrinsic ones given in \cite{akapilavai2022}. In Section~\ref{end} we conclude the study, give some final remarks, and point out possible applications and future research.

%%%%%%%%%%%%%%%%%%%%%%%%%%%%%%%%%%%%%%%%%%

\section{Intrinsic metric formulation}\label{intrinsic}

\subsection{Strain energy functions between geometric elements}
 In this section, we provide a summary of the approach presented in \cite{nawratil2020snappability}, which is the base for the formulation of the intrinsic metrics for the presented nine interpretations. 
  To begin, let us recap the necessary assumptions made in \cite{nawratil2020snappability}.
 
\begin{assumption}\label{ass:1}
All the bars and triangular plates of the framework are made of the same homogeneous isotropic material, deforming at constant volume; i.e.\ the Poisson ratio equals $\tfrac{1}{2}$. Additionally, all bars have the same cross-sectional area. Finally, pin-jointed triangular bar structures and corresponding triangular plates are made of the same amount of material. 
\end{assumption}

\subsubsection{Green-Lagrange strain energy for bars}

The formula for the elastic  Green-Lagrange (GL) strain energy between an undeformed bar $\vert_{ij}$ and its deformation $\vert'_{ij}$ is given by:
\begin{equation}
U\left(\vert_{ij},\vert'_{ij}\right)=\frac{EA}{8\ell(\mathbf{k}_{i},\mathbf{k}_{j})^3}\left(\ell(\mathbf{k}'_{i},\mathbf{k}'_{j})^{2}-\ell(\mathbf{k}_{i},\mathbf{k}_{j})^2 \right)^2,
\label{metricdp3}
\end{equation}  
where $A$ is the cross sectional-area of the undeformed bar $\vert_{ij}$ and $E$ denotes the Young modulus.
Moreover, the bar lengths of the undeformed bar  $\vert_{ij}$ 
and the deformed bar  $\vert'_{ij}$ are computed by
\begin{equation}\label{length}
\ell(\mathbf{k}_{i},\mathbf{k}_{j})=\sqrt{(\mathbf{k}_{i}-\mathbf{k}_{j})(\mathbf{k}_{i}-\mathbf{k}_{j})}
\quad \text{and} \quad
\ell(\mathbf{k}'_{i},\mathbf{k}'_{j})=\sqrt{(\mathbf{k}'_{i}-\mathbf{k}'_{j})(\mathbf{k}'_{i}-\mathbf{k}'_{j})}. 
\end{equation}

\subsubsection{Green-Lagrange strain energy for triangular plates}

For that, we consider the $2 \times 2$ matrix $\mathbf{F}$ of the affine transformation with
\begin{equation}\label{triangle1}
\mathbf{k}_{j}-\mathbf{k}_{i} \mapsto  \mathbf{k}'_{j}-\mathbf{k}'_{i}  \quad \textnormal{and} \quad
\mathbf{k}_{k}-\mathbf{k}_{i} \mapsto  \mathbf{k}'_{k}-\mathbf{k}'_{i} 
\end{equation}
from the undeformed triangular plate $\blacktriangle_{ijk}$ with $i<j<k$ to the deformed triangular plate $\blacktriangle'_{ijk}$, where the index set $\{i,j,k\}$ can either be $\{1,2,3\}$ or $\{4,5,6\}$. Based on this matrix $\mathbf{F}$ the Green-Lagrange (GL) normal strains $\epsilon_{x}$ and $\epsilon_{y}$ and the GL shear strain $\gamma_{xy}$ can be computed as:
\begin{equation}
\begin{pmatrix}
\epsilon_{x} & \gamma_{xy}  \\ \gamma_{xy} & \epsilon_{y} 
\end{pmatrix}
=\frac{1}{2}(\mathbf{F^{T}}\mathbf{F}-\mathbf{I}).
\end{equation}
We reassemble these quantities in the vector $\mathbf{q} = \left(\epsilon_{x}, \epsilon_{y}, 2\gamma_{xy} \right)^{T}$. Using this notation,the resulting total elastic strain energy $U$ of  the  deformed triangular plate $\blacktriangle'_{ijk}$ with respect to the undeformed state 
 $\blacktriangle_{ijk}$ can be calculated as follows: 
\begin{align}\label{energy:plate}
U\left(\blacktriangle_{ijk}, \blacktriangle'_{ijk}\right)=E\text{Vol}_{\blacktriangle_{ijk}}\mathbf{q}^T\mathbf{S}\,\mathbf{q} \quad \text{with} \quad
\mathbf{S}:=\frac{1}{6}
\begin{pmatrix}
4 & 2 & 0 \\
2 & 4 & 0 \\
0 & 0 & 1 \\
\end{pmatrix}
\end{align}
and  $\text{Vol}_{\blacktriangle_{ijk}}$ denoting the volume of $\blacktriangle_{ijk}$ with
\begin{equation}\label{eq:volume}
\text{Vol}_{\blacktriangle_{ijk}}=A(\ell(\mathbf{k}_{i},\mathbf{k}_{j})+\ell(\mathbf{k}_{i},\mathbf{k}_{k})+\ell(\mathbf{k}_{j},\mathbf{k}_{k})).
\end{equation}
Note that the energy expression of the plate only contains even powers of the deformed edge lengths of the triangle (cf.\ Lemma 1 of \cite{nawratil2020snappability} and \cite{NAWRATIL2022104510}).
This concludes the essential required formulations for setting up the intrinsic metrics for the nine interpretations done next.

\subsection{Intrinsic metrics for the nine interpretations}\label{sec:metrics}
The total elastic GL strain energy, denoted as  $U_\star^\circ(\mathbf{K},\mathbf{K}')$ with $\circ,\star\in\left\{\blacktriangle, \vartriangle, \tikz[baseline=0.2ex]\draw [fill=black,black](0,0.2) rectangle (0.25,0.36ex);\right\}$, corresponds to of one of the nine framework interpretations given in Fig.~\ref{inter}. This energy is a result of summing up the energies of the involved bars and triangular plates, specified in (cf.\ Eqs.\ (\ref{metricdp3}) and (\ref{energy:plate})) respectively. Its density $D_\star^\circ(\mathbf{K},\mathbf{K}')$ is obtained by dividing $U_\star^\circ(\mathbf{K},\mathbf{K}')$ by the total volume of the framework, which reads for the nine interpretations as follows:

\begin{align}
    D_{\rectangleblack}^{\rectangleblack}(\mathbf{K},\mathbf{K}') &=
    \frac{1}{\Omega_{I_1}} \sum_{(i,j) \in I_1} {U\left( \lineseg_{ij}, \lineseg_{ij}^{'}\right)},
    \label{In1} \\
    D^{\blacktriangle}_{\rectangleblack} (\mathbf{K},\mathbf{K}') &=\frac{1}{\Omega_{I_2}} \left[ \sum_{(i,j) \in I_1}  {U\left( \lineseg_{ij}, \lineseg_{ij}^{'}\right)} + {U\left(\blacktriangle_{456}, \blacktriangle_{456}'\right)}\right],
    \label{In2}\\
    D_{\rectangleblack}^{\vartriangle}(\mathbf{K},\mathbf{K}') &=\frac{1}{\Omega_{I_2}} \sum_{(i,j) \in I_2} U\left(\lineseg_{ij},\lineseg'_{ij}\right),
    \label{In3} \\
    D^{\rectangleblack}_{\blacktriangle} (\mathbf{K},\mathbf{K}') &=\frac{1}{\Omega_{I_3}} \left[ \sum_{(i,j) \in I_1}  {U\left( \lineseg_{i,j}, \lineseg_{ij}^{'}\right)} + {U\left(\blacktriangle_{123}, \blacktriangle_{123}'\right)}\right],
    \label{In4} \\
D_{\vartriangle}^{\rectangleblack}(\mathbf{K},\mathbf{K}') &= 
    \frac{1}{\Omega_{I_3}} \sum_{(i,j) \in I_3} {U\left( \lineseg_{ij}, \lineseg_{ij}^{'}\right)},
    \label{In5} \\
 D_{\blacktriangle}^{\blacktriangle}(\mathbf{K},\mathbf{K}') &=
    \frac{1}{\Omega_{I_4}} \left[{ \sum_{(i,j)\in I_1}  U \left(\lineseg_{ij}, \lineseg_{ij}^{'}\right)} +  U\left(\blacktriangle_{123}, \blacktriangle_{123}'\right) + U\left(\blacktriangle_{456}, \blacktriangle_{456}'\right) \right],
     \label{In6} \\
    D_{\blacktriangle}^{\vartriangle}(\mathbf{K},\mathbf{K}') &= \frac{1}{\Omega_{I_4}}\left[{ \sum_{(i,j)\in I_2} U\left( \lineseg_{ij}, \lineseg_{ij}^{'}\right)}+{U\left(\blacktriangle_{123}, \blacktriangle_{123}'\right)}\right],
    \label{In7} \\
    D_{\vartriangle}^{\blacktriangle}(\mathbf{K},\mathbf{K}') &= \frac{1}{\Omega_{I_4}}\left[{ \sum_{(i,j)\in I_3} U\left( \lineseg_{ij}, \lineseg_{i,j}^{'}\right)}+{U\left(\blacktriangle_{456}, \blacktriangle_{456}'\right)}\right],
     \label{In8} \\
    D_{\vartriangle}^{\vartriangle}(\mathbf{K},\mathbf{K}') &= 
    \frac{1}{\Omega_{I_4}} \sum_{(i,j) \in I_4} {U\left( \lineseg_{ij}, \lineseg_{ij}^{'}\right)},
     \label{In9}
\end{align}

\noindent where $\Omega_I$  
denotes the  cross-sectional area $A$ times the summation of bar lengths $\ell_{ij}$ for
$(i,j)\in I$ with 
 \begin{align}
    I_1 &= \{(1, 4), (2,5), (3,6)\}, \label{index1} \\
    I_2 &= \{(1, 4), (2, 5), (3, 6), (4, 5), (4, 6), (5, 6)\}, \label{index2}\\
    I_3 &= \{(1, 4), (2, 5), (3, 6), (1, 2), (2, 3), (1, 3)\}, \label{index3} \\
     I_4 &= \{(1, 2), (2, 3), (1, 3), (1, 4), (2, 5), (3, 6), (4, 5), (5, 6), (4, 6)\} \label{index4}.
 \end{align}
 \noindent
 
It should be noted that all  $D_\star^\circ(\mathbf{K},\mathbf{K}')$ are independent of 
the bar's cross-sectional area $A$ (as it cancels out), and that the Young modulus $E$ can be factored out. As $E$ plays the role of a scaling factor, we can set it equal to one. Taking this into account, the density functions serve as our distance functions, 
which only rely on the inner metric of the 
framework and not on its embedding into the Euclidean plane. Therefore, it would be more precise to write 
$D_\star^\circ(\mathbf{L},\mathbf{L}')$ instead of $D_\star^\circ(\mathbf{K},\mathbf{K}')$. However, we stick to the latter notation as we solve the optimization problem in the embedding space, which is explained in detail in the next section.

\section{The constrained optimization problems for computing the singularity distance}\label{generalcase}

\subsection{Closest configuration on the singularity variety}\label{sec:singvar}

According to \cite[Section 5]{NAWRATIL2022104510}, the singularity distance can be computed by finding the real point $\mathbf{K}'$ on the {\it singularity variety} $V=0$ defined in Eq.\ (\ref{variety}) that minimizes the 
value $D_\star^\circ(\mathbf{K},\mathbf{K}')$,  where $G_\star^\circ(\mathbf{K})$ represents the given undeformed realization of the framework. 
In addition, there should exist a 1-parametric deformation of $G_\star^\circ(\mathbf{K})$ into $G_\star^\circ(\mathbf{K}')$ such that the deformation energy density has to increase monotonically. 

 The following explains how the configuration $G_{\star}^{\circ}(\mathbf{K}')$ can be computed:

\begin{itemize}
\item
$D_\star^\circ(\mathbf{K},\mathbf{K}')$ 
 with $\circ,\star\in\left\{\blacktriangle, \vartriangle,\hrectangleblack \right\}$ but $(\circ,\star)\neq (\hrectangleblack,\hrectangleblack)$:
 For these eight framework interpretations, the Lagrangian of the contained optimization problem reads as:
 \begin{equation}\label{eq:lagrange}
     L:=D_\star^\circ(\mathbf{K},\mathbf{K}')+\lambda V,
 \end{equation}
where $\lambda$ denotes the Lagrange multiplier.
$G_{\star}^{\circ}(\mathbf{K}')$  corresponds to one of the critical points of this function $L$; i.e.\ 
solutions of the system of partial derivatives of 
$L$ with respect to $\lambda$ and the free
coordinates $(c_i,d_i)$ of the points $\mathbf{k}_i'$ for 
$i=1,\ldots ,6$. The number of unknowns (cf.\ Table \ref{tabel1})
depends on the framework interpretation, which is discussed next:

As we are dealing with an intrinsic metric, we can always assume without loss of generality that $\mathbf{k}_1'$ equals the origin of the fixed frame and $\mathbf{k}_2'$ is located on its $x$-axis; i.e.\  $c_1=d_1=d_2=0$, which will be the case for the rest of the paper. Therefore, 10 unknowns remain for the four cases $D_\star^\circ(\mathbf{K},\mathbf{K}')$ 
 with $\circ,\star\in\left\{\blacktriangle, \vartriangle \right\}$.

\begin{itemize}
\item[--] Moreover, if only the 
base is made of undeformable material; i.e. $G_{\rectangleblack}^{\circ}(\mathbf{K}')$ with 
$\circ\in\left\{\blacktriangle, \vartriangle \right\}$
then we can set in addition $c_2=x_2$, $c_3=x_3$ and $d_3=y_3$. Therefore we remain with 7 unknowns for these two cases.

\item[--] If only the platform is made of undeformable material, i.e.\ 
$G_{\star}^{\rectangleblack}(\mathbf{K}')$ with $\star\in\left\{\blacktriangle, \vartriangle \right\}$
we consider the inverse motion (cf.\ caption of Fig.\ \ref{inter}) and we end up with the same two cases 
mentioned before. 

\end{itemize}

There is a problem with the Lagrangian formulation of Eq.\ (\ref{eq:lagrange}), namely that the singular points of the singularity variety $V=0$ are not covered by this approach. 
Consequently, it becomes necessary to handle these points separately. In \cite{akapilavai2022}, the authors have already characterized these singular points geometrically by one of the following four configurations:
\begin{enumerate}
\item Three legs of the manipulator are collinear.
\item  Two legs are collinear and one leg degenerates to a point.
\item Two legs degenerate to points.
\item One leg degenerates to a point and the carrier lines of the remaining two legs pass through that point.
\end{enumerate}
 We restrict ourselves to the singular points, which correspond to case 1,  as in practice no leg can have zero length; i.e.\ cases 2-4 are not of interest.  
In order to include these singular points into our computation of the singularity distance, we parametrize this set by setting also $d_3=\ldots=d_6=0$ thus the 
configuration is parameterized by $c_2,\ldots,c_6$.
Now we minimize $D_\star^\circ(\mathbf{K},\mathbf{K}')$ without any side-condition; i.e.\
\begin{equation}\label{eq:f}
L=D_\star^\circ(\mathbf{K},\mathbf{K}').
\end{equation}
If the base points are collinear then case 1 can also appear for $\star=\hrectangleblack$. In this case, we set again $c_2=x_2$ and $c_3=x_3$, respectively. 
Clearly, for $\circ=\hrectangleblack$ the same considerations hold regarding the inverse motion. \item
$D_{\rectangleblack}^{\rectangleblack}(\mathbf{K},\mathbf{K}')$:
For the determination of
$G_{\rectangleblack}^{\rectangleblack}(\mathbf{K}')$ we can set again  $c_2=x_2$, $c_3=x_3$ and $d_3=y_3$ and achieve 
the Euclidean motion of the platform by using the so-called point-based representation \cite{kapilavai2020homotopy}; i.e.\ the coordinates of $\mathbf{k}'_{6}$ can be written as
\begin{equation}
\begin{pmatrix}
    c_6\\
    d_6
\end{pmatrix} =\begin{pmatrix}
\frac{(c_{5}-c_{4})x_{6}+(d_{4}-d_{5})y_{6}+c_{4}x_{5}}{x_{5}}\\
\frac{(d_{5}-d_{4})x_{6}+(c_{5}-c_{4})y_{6}+d_{4}x_{5}}{x_{5}}
\end{pmatrix},
\label{PBR2}
\end{equation}
in dependency of $\mathbf{k}'_{4}$ and $\mathbf{k}'_{5}$, which  can freely be chosen under consideration that the condition $E=0$ holds with 
\begin{equation}
E=\|\mathbf{k}'_{5}-\mathbf{k}'_{4}\|^{2}-\|\mathbf{p}_{5}-\mathbf{p}_{4}\|^2. 
\label{platform1}
\end{equation}
As in this case, the parameters $c_4,c_5,d_4,d_5$ are restricted by an additional side condition we have to take the partial derivatives of the Lagrangian function
\begin{equation}\label{eq:lagrangeplus}
 L:= D_{\rectangleblack}^{\rectangleblack}(\mathbf{K},\mathbf{K}')+\kappa E+\lambda V,
\end{equation}
where $\kappa$ and $\lambda$ are the Lagrange multipliers.

In \cite{akapilavai2022}, it was also shown that the singular points of the constrained variety of this optimization problem correspond to the same four configurations as mentioned above. 
Now, case 1 can only occur if the given base points are collinear, as well as the platform points. If this holds, we set again $d_3=\ldots=d_6=0$ and $c_2=x_2$ as well as $c_3=x_3$. The parameterization is completed by
\begin{equation}\label{pmsign}
c_5=c_4\pm x_5, \quad c_6=c_4\pm x_6.
\end{equation}
Now we minimize $D_\star^\circ(\mathbf{K},\mathbf{K}')$ without any side-condition; i.e.\ 
\begin{equation}\label{eq:fplus}
L=D_{\rectangleblack}^{\rectangleblack}(\mathbf{K},\mathbf{K}')
\end{equation}
which only depends on the unknown $c_4$ but has to be done twice, due to the $\pm$ sign in Eq.\ (\ref{pmsign}). 
\end{itemize}

As already mentioned at the end of the first paragraph in Section \ref{terminology}, we get additional singular configurations if the platform/base is interpreted as a triangular bar structure. The optimization problem for computing the closest configurations with collinear platform/base anchor points is discussed in the next subsection. 

\begin{table}[h]
\caption{Summary of the  computational data  with respect to $L$ of Eqs.\ (\ref{eq:lagrange}) (\ref{eq:lagrangeplus}), (\ref{eq:lagrange1}) and  (\ref{eq:lagrange2}).}
    \centering
\begin{tabular}{|l||l|l|l|}
\hline
Interpretation &  \begin{tabular}[c]{@{}l@{}} unknowns\end{tabular}&      \begin{tabular}[c]{@{}l@{}} $\#$ unknowns \end{tabular} \\ \hline
$G_{\rectangleblack}^{\rectangleblack}(\mathbf{K'})$ & \begin{tabular}[c]{@{}l@{}} $c_4,c_5,d_4,d_5,\kappa,\lambda$   \end{tabular} &  6 \\ \hline
$G^{\blacktriangle}_{\rectangleblack} (\mathbf{K'})$, $G_{\rectangleblack}^{\vartriangle}(\mathbf{K'})$,
inversions of
$ G^{\rectangleblack}_{\blacktriangle} (\mathbf{K'})$, $G_{\vartriangle}^{\rectangleblack}(\mathbf{K'})$
& \begin{tabular}[c]{@{}l@{}} $c_4,c_5,c_6,d_4,d_5,d_6,\lambda$  \end{tabular}& 7 \\ \hline

\begin{tabular}[c]{@{}l@{}}    $G_{\blacktriangle}^{\blacktriangle}(\mathbf{K'})$, $G_{\vartriangle}^{\vartriangle}(\mathbf{K'})$,  $G_{\blacktriangle}^{\vartriangle}(\mathbf{K'})$, 
 $G_{\vartriangle}^{\blacktriangle}(\mathbf{K'})$ 
    \end{tabular} &  \begin{tabular}[c]{@{}l@{}}     $c_2,c_3,c_4,c_5,c_6,d_3,d_4,d_5,d_6,\lambda$  \end{tabular}  & 10  \\ \hline
\end{tabular}
\label{tabel1}
\end{table}

\begin{table}[h]
\caption{Summary of the  computational data  with respect to  $L$ of Eqs.\ (\ref{eq:f}) and (\ref{eq:fplus}). 
}
    \centering
\begin{tabular}{|l||l|l|l|}
\hline
Interpretation &  \begin{tabular}[c]{@{}l@{}} unknowns\end{tabular}&      \begin{tabular}[c]{@{}l@{}} $\#$ unknowns \end{tabular} \\ \hline
$G_{\rectangleblack}^{\rectangleblack}(\mathbf{K'})$ &  $c_4$    &  1 \\ \hline
$G^{\blacktriangle}_{\rectangleblack} (\mathbf{K'})$, $G_{\rectangleblack}^{\vartriangle}(\mathbf{K'})$,
inversions of $ G^{\rectangleblack}_{\blacktriangle} (\mathbf{K'})$, $G_{\vartriangle}^{\rectangleblack}(\mathbf{K'})$ & \ $c_4,c_5, c_6$  & 3 \\ \hline

$G_{\blacktriangle}^{\blacktriangle}(\mathbf{K'})$, $G_{\vartriangle}^{\vartriangle}(\mathbf{K'})$,  $G_{\blacktriangle}^{\vartriangle}(\mathbf{K'})$, inversion of $G_{\vartriangle}^{\blacktriangle}(\mathbf{K'})$ &  $c_2,c_3,c_4,c_5,c_6$   & 5  \\ \hline
\end{tabular}
\label{tabel1b}
\end{table}

\subsection{Closest configuration on the collinearity variety}\label{collinearity}

If the base or platform is interpreted as triangular bar-structure $(\vartriangle)$, the additional singularities can be characterized 
algebraically by the condition $C_B=0$ and $C_P=0$, respectively, with
\begin{equation}
\mathrm{C}_B =\det\begin{pmatrix}
1 & 1 & 1 \\ 
0 & c_{2} &  c_{3} \\
0 & 0 &  d_{3} \\
\end{pmatrix},\quad
\mathrm{C}_P =\det\begin{pmatrix}
1 & 1 & 1 \\ 
c_{4} & c_{5} &  c_{6} \\
d_{4} & d_{5} &  d_{6} \\
\end{pmatrix}.
\label{coll}
\end{equation}
Note that the so-called {\it collinearity varieties} 
$C_B=0$ and $C_P=0$ are both quadratic. 

\begin{itemize}
    \item 
$D_\vartriangle^\circ(\mathbf{K},\mathbf{K}')$ 
 with $\circ\in\left\{\blacktriangle, \vartriangle \right\}$:
 For these two framework interpretations (and for $D^{\vartriangle}_\star(\mathbf{K},\mathbf{K}')$ 
 with $\star\in\left\{\blacktriangle, \vartriangle \right\}$ regarding the inverse motion with side condition $C_{P}$) the Lagrangian of the contained optimization problem reads as:
 \begin{equation}
L:=D_\vartriangle^\circ(\mathbf{K},\mathbf{K}')+\lambda C_B,
 \label{eq:lagrange1}
 \end{equation}
where $\lambda$ denotes the Lagrange multiplier. 
{The number of unknowns within this optimization problem equals $10$ (cf.\ Table ~\ref{tabel1}).}

Once again, the singular points of $C_B=0$ are not considered within the formulation of Eq.\ (\ref{eq:lagrange1}), 
which corresponds to configurations where the base degenerates into a point. Therefore, one has to study this case separately by parametrizing it as follows: We set 
$c_2=c_3=d_3=0$ and can further assume $d_4=0$. Then we have to minimize $D_\vartriangle^\circ(\mathbf{K},\mathbf{K}')$ (and for $D^{\vartriangle}_\star(\mathbf{K},\mathbf{K}')$ 
 with $\star\in\left\{\blacktriangle, \vartriangle \right\}$ regarding the inverse motion) without any side condition; i.e.\
\begin{equation}
L:=D_\vartriangle^\circ(\mathbf{K},\mathbf{K}').
 \label{eq:lagrange4}
 \end{equation}
{The number of unknowns within this optimization problem equals $5$ (cf.\ Table ~\ref{spcv}).}

\begin{table}[]
\caption{
Computational data  with respect to  $L$ of Eqs.\ (\ref{eq:lagrange4}) and (\ref{eq:point1}) with constrained varieties $C_{B}=0$ and $C_{P}=0$ .}
\centering
\begin{tabular}{|l||l|l|}
\hline
Interpretation & unknowns & $\#$ unknowns \\ \hline
 $G_{\vartriangle}^{\blacktriangle}(\mathbf{K'})$, $G_{\vartriangle}^{\vartriangle}(\mathbf{K'})$,  inversion of $G^{\vartriangle}_{\blacktriangle}(\mathbf{K'})$, $G_{\vartriangle}^{\vartriangle}(\mathbf{K'})$   &     $c_{4}, c_{5}, c_{6},d_{5}, d_{6}, $  &      $5$    \\ \hline
  $G_{\rectangleblack}^{\vartriangle}(\mathbf{K'})$ inversion of  $G^{\rectangleblack}_{\vartriangle}(\mathbf{K'})$                &   $c,d$      & 2 \\  \hline
\end{tabular}
\label{spcv}
\end{table}

\item
    $D_{\rectangleblack}^{\vartriangle}(\mathbf{K},\mathbf{K}')$: For this framework interpretation (and for $D^{\rectangleblack}_{\vartriangle}(\mathbf{K},\mathbf{K}')$ regarding the inverse motion) the Lagrangian of the contained optimization problem reads as:
 \begin{equation}\label{eq:lagrange2}
     L:=D_{\rectangleblack}^{\vartriangle}(\mathbf{K},\mathbf{K}')+\lambda C_P,
 \end{equation}
where we can set again $c_2=x_2$, $c_3=x_3$ and $d_3=y_3$. The number of unknowns within this optimization problem equals $7$ (cf.\ Table \ref{tabel1}).

Again the singular points of $C_P=0$ are not considered within the formulation of Eq.\ (\ref{eq:lagrange2}),  which correspond to configurations where the platform degenerates into a point. Therefore, one has to study this case separately by parametrizing it as follows. In addition to $c_2=x_2$, $c_3=x_3$ and $d_3=y_3$ we set
$c:=c_4=c_5=c_6$ and $d:=d_4=d_5=d_6$. Then we have to minimize $D_{\rectangleblack}^{\vartriangle}(\mathbf{K},\mathbf{K}')$ without any side-condition; i.e.\ 
\begin{equation}\label{eq:point1}
 L:=D_{\rectangleblack}^{\vartriangle}(\mathbf{K},\mathbf{K}').
\end{equation}
The number of unknowns for this optimization problem is summarized in Table ~\ref{spcv}.
\end{itemize}

\subsection{Identification of the closest singular configuration}\label{non}

For a chosen metric $D_\star^\circ(\mathbf{K},\mathbf{K}')$, we solve all the mentioned optimization problems formulated in the last two subsections and filter out the local minima from the set of critical points by checking the eigenvalues of the (bordered) Hessian matrix in the (constrained) unconstrained case (cf.\ \cite{borderhessian}). Then, we merge the obtained local minima of the respective minimization problems within a set  $\mathcal{C}$.  

Let $G_\star^\circ(\mathbf{K}'_1),G_\star^\circ(\mathbf{K}'_2), \ldots$ denote the entries of  $\mathcal{C}$ 
in ascending order with respect to the intrinsic metric $D_\star^\circ(\mathbf{K},\mathbf{K}_i')$. 
But it is not said that
 $G_\star^\circ(\mathbf{K}'_1)$ is the closest singularity to the
 realization 
$G_{\star}^{\circ}(\mathbf{K})$ is associated with the given 3-RPR robot configuration as we are dealing with an intrinsic metric. It can also be the closest singularity to another realization of that framework. Therefore, we have to 
check if a continuous change of $\mathbf{L}$ into $\mathbf{L}'_1$ implies a continuous deformation of $G_{\star}^{\circ}(\mathbf{K})$ into $G_{\star}^{\circ}(\mathbf{K}'_1)$. 
According to \cite{NAWRATIL2022104510}  this can be done by considering the following change of the inner metric 
\begin{equation}\label{eq:path}
\sqrt{\ell(\mathbf{k}'_{1,i},\mathbf{k}'_{1,j})^{2}+{h}\left(\ell(\mathbf{k}_{i},\mathbf{k}_{j})^{2}-\ell(\mathbf{k}'_{1,i},\mathbf{k}'_{1,j})^{2}\right)}, \end{equation} 
where $h$ is the deformation parameter (homotopy parameter) running from one to zero. 
Now we track the path of the realization $G_{\star}^{\circ}(\mathbf{K})$ implied by that homotopy, which ends up in a configuration $G_{\star}^{\circ}(\mathbf{K}'')$. If 
$G_{\star}^{\circ}(\mathbf{K}'')=G_{\star}^{\circ}(\mathbf{K}'_1)$ holds true, we have found the closest singularity to $G_{\star}^{\circ}(\mathbf{K})$; otherwise we repeat the complete procedure with respect to the next point $G_\star^\circ(\mathbf{K}'_2)$ in the sorted list $\mathcal{C}$ until we find the closest singular configuration.

\subsection{Comments on the singularity distance computation}\label{sec:erratum}

In \cite{NAWRATIL2022104510}, 
the singularity distance was obtained based on the computation of saddle points of the distance function $D_{\star}^{\circ}(\mathbf{K},\mathbf{K}')$ only. The reasoning for this procedure was given in \cite[Theorem 5]{NAWRATIL2022104510}, which turned out to be not entirely correct due to the following reason:

A singularity can also be seen as a configuration where at least two of the realizations coincide. Therefore, in a singular configuration $G_{\star}^{\circ}(\mathbf{K}')$, at least two of the paths of Eq.\ (\ref{eq:path}) meet, where one starts in the given realization $G_{\star}^{\circ}(\mathbf{K})$ and the other $k>0$ paths
in the realizations $G_{\star}^{\circ}(\mathbf{K}_1),\ldots, G_{\star}^{\circ}(\mathbf{K}_k)$. 
Now, in the proof of \cite[Theorem 5]{NAWRATIL2022104510}, the argument was used that 
this circumstance already implies that the framework can snap over $G_{\star}^{\circ}(\mathbf{K}')$ from the configuration $G_{\star}^{\circ}(\mathbf{K})$ into another 
realization $G_{\star}^{\circ}(\mathbf{K}_1)$,\ldots, $G_{\star}^{\circ}(\mathbf{K}_k)$. This is true for ordinary singularities ($k=1$), but for higher-order singularities ($k>1$) this has not been the case. E.g.\ for $k=2$, the corresponding two realizations $G_{\star}^{\circ}(\mathbf{K}_1), G_{\star}^{\circ}(\mathbf{K}_2)$ can be conjugate complex, where the corresponding conjugate complex paths exactly meet in the real configuration  $G_{\star}^{\circ}(\mathbf{K}')$. Therefore, $G_{\star}^{\circ}(\mathbf{K}')$ is not a saddle point of the distance function $D_{\star}^{\circ}(\mathbf{K},\mathbf{K}')$. This is demonstrated (see Fig.\ \ref{zeinsingular}(a)) within the illustrating example discussed in Section \ref{sec:example}. 

Summarizing, one can say that Theorem 5 of \cite{NAWRATIL2022104510}, can only be used for a secure computation of the distance to singularities with odd $k$ (which include the ordinary singularities).

\subsection{How to handle manipulators with collinear platform/base}\label{sec:collpb}

For the interpretation of the platform/base as 
$\blacktriangle$ or $\vartriangle$, the applied transformation is an affine one, in contrast to the interpretation $\tikz[baseline=0.2ex]\draw [fill=black,black](0,0.2) rectangle (0.25,0.36ex);$ implying an Euclidean motion. When the triangular platform/base degenerates into a straight bar made of deformable/undeformable material, we can use the same transformations. It should only be pointed out that the affine transformation of a line segment reduces to an equiform motion. Keeping this in mind, it is straightforward to adapt the concepts presented in the paper to 3-RPR manipulators possessing a collinear platform/base.

\section{Computational procedure}\label{results}
In this section, we present a pipeline for
computing the closest singular configuration along a one-parametric motion, which is discretized into $n$ poses, where $n$ can be defined by the user. 

For all the optimization problems discussed in Section \ref{generalcase}, we end up with a certain set $\mathcal{D}$ of algebraic equations, which is in most cases no longer solvable in a reasonable time using symbolic methods (e.g.\ G\"oberner basis method) implemented in computer algebraic system (e.g.~software \texttt{Maple}). For this purpose, we use algorithms from numerical algebraic geometry implemented in the  freeware~\texttt{Bertini}\footnote{Reasons for choosing \texttt{Bertini} are given in~\cite[Sec.\ 1]{kapilavai2020homotopy}.}.

In Subsection \ref{sec:proc_main}, we will focus on the main optimization problem of finding the closest regular point $\mathbf{K}'$ on the singularity variety $V=0$ to a given non-singular configuration $\mathbf{K}$ with respect to the nine presented intrinsic metrics. 
In Subsection \ref{sec:proc_main4}, we will proceed with computing the distance to the closest regular point on the collinearity varieties $C_B=0$ and $C_P=0$. 
Subsections \ref{sec:proc_special1} and \ref{sec:proc_special} are devoted to the determination of the distance to the closest singular point on the constrained varieties.

\subsection{Computing the closest regular point on the singularity variety}\label{sec:proc_main}

\begin{figure}[t]
    \begin{center}
      \includegraphics[width=15cm]{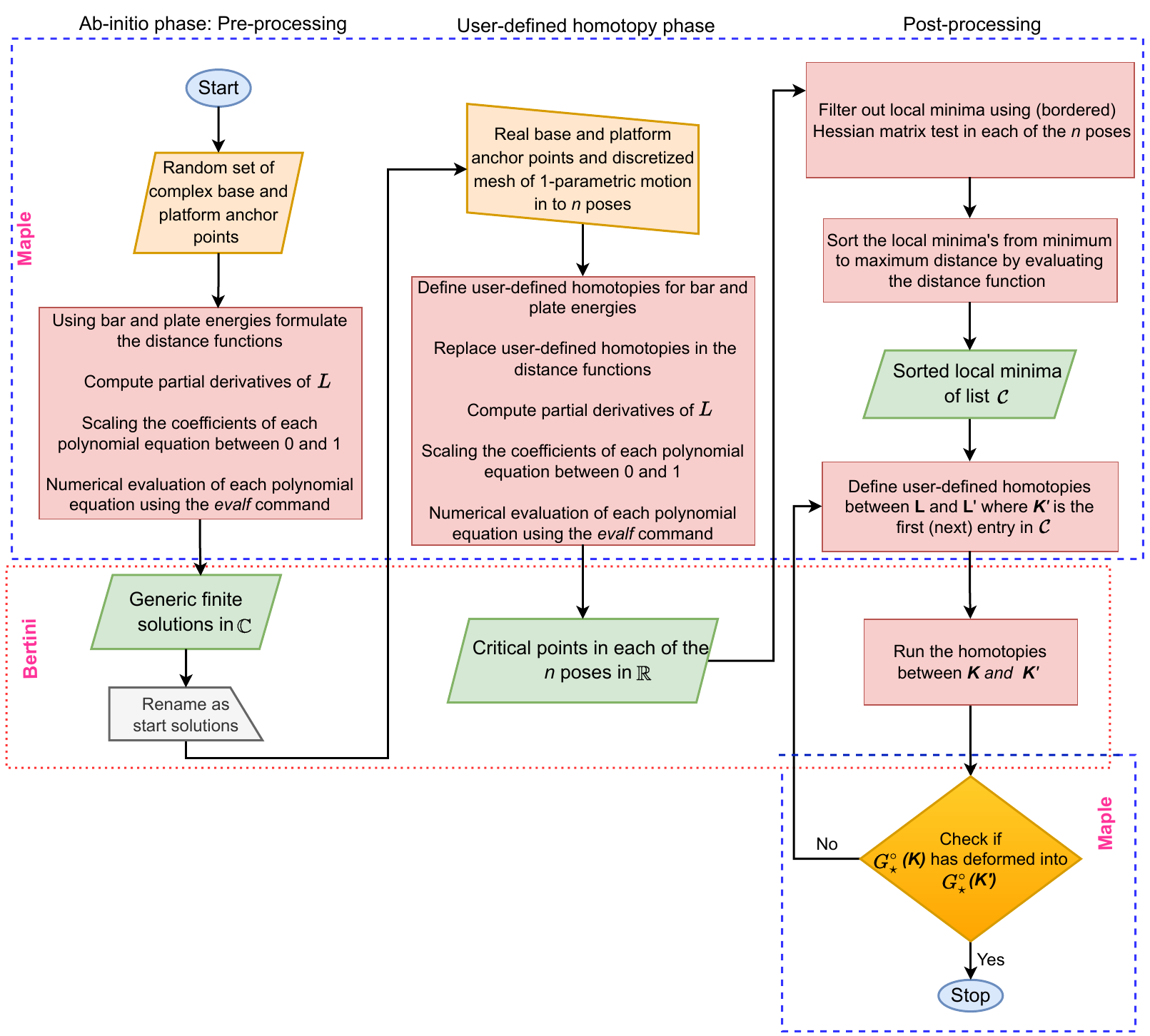}
      \caption{Operational flowchart for the homotopy-based computational pipeline.}
      \label{flow}
    \end{center}
\end{figure}

The overall operational procedure for the computation of $\mathbf{K}'$ can be divided into three phases, as illustrated in the flowchart provided in Fig.\ref{flow}. The first phase, the ab-initio phase, begins with a generic complex manipulator configuration, and through the utilization of \texttt{Bertini}, the solution set of the resulting polynomial system $\mathcal{D}$ is computed. These solutions serve as the starting points for user-defined homotopies within \texttt{Bertini}, which are then traced to the solution sets of the $n$ poses. Both the ab-initio phase and the user-defined homotopy phase entail input file generation for \texttt{Bertini} through \texttt{Maple}. The second phase, the user-defined homotopy phase, involves the tracking of user-defined homotopies to the solution sets of the $n$ poses using the \texttt{Maple-Bertini} interface. Finally, in the post-processing phase, the \texttt{Maple-Bertini} interface is utilized again to identify the closest singular configuration, as described in Section\ref{non}. Detailed discussions of these three phases are provided in the following subsections.

\subsubsection{Ab-Initio phase}\label{settings}
We start with a generic framework by picking randomly all nine remaining
coordinates (besides $c_1=d_1=d_2=0$) with respect to the fixed frame from the set of complex numbers $\mathbb{C}$. 
 Therefore, we denote the resulting vectors of the anchor points with respect to the fixed frame by $\mathbf{k}^{\mathbb{C}}_i$ for $i=1,\ldots,6$ and consequently the complex configuration by $G_{\star}^{\circ}(\mathbf{K}^{\mathbb{C}})$.

For this input, we want to solve the square system $\mathcal{D}$ of non-homogeneous polynomial equations, which is called the {\it original system}. 
 We use the computer algebra system software \texttt{Maple} as a pre-processor to generate the system $\mathcal{D}$ and for writing the input file for \texttt{Bertini}{\footnote{All the  results reported in this article are computed using $\mathtt{Bertini 1.6v}$}}, where the following points have to be considered:

\begin{itemize}
 \item \texttt{Bertini} throws an error message if the coefficients of individual equations are too large. To avoid this we divided each equation through the maximal absolute value of appearing coefficients to roughly have the same magnitude between $0$ and $1$ (cf.\ \cite[Sec.\ 7.7]{bates2013numerically}).
 \item
 We evaluate individual equations using the $\mathtt{evalf}$ function in \texttt{Maple} which generates individual equations with floating point arithmetic (up to a certain number of digits depending on the precision used for computation). This approach is taken because the authors of \texttt{Bertini} implemented consciously the evaluation of the square root only over $\mathbb{R}$ to avoid numerical instabilities near the branch cut \cite{personal}.
 
\begin{rmk}
 HC.jl~\cite{julia} can evaluate square root over $\mathbb{C}$, which eliminates the numerical evaluation of scaled equations using $\mathtt{evalf}$ function in $\mathtt{Maple}$. 
\hfill $\diamond$ 
\end{rmk}
\end{itemize}

Using this input file, \texttt{Bertini} computes the solution set of the original system by means of homotopy continuation. However, we ensure that we do not miss any finite solution with this numerical approach; otherwise, we cannot guarantee that the further computations, which are based on this solution set, lead to the sought-after global minima. 
Due to the degree and size of $\mathcal{D}$ we are not able to get information (number of isolated solutions, existence of higher dimensional components) about its zero set by using symbolic methods (e.g.\ Gr\"obner basis method within \texttt{Maple}). 
Therefore, we proceed with the following two-step approach:
\begin{enumerate}
    \item 
{\bf Higher dimensional components:}
First, we check if the solution set of $\mathcal{D}$ contains higher dimensional components by invoking $\mathtt{Track Type 1}$ (see \cite[Sec 8.2]{bates2013numerically}) in the \texttt{Bertini} input file.
Moreover, we use the following {configuration settings}\footnote{\texttt{Bertini} allows numerous configuration settings (for details we refer to~\cite[p. 307-321]{bates2013numerically}), which in general have to be tuned manually to adapt the computations to the problem at hand.} in the input file, which is used throughout all the computations reported in this article. 
\begin{itemize}
     \item Several predictor-corrector methods are implemented, but we use the default method, which involves a $4^{th}$ order Runge-Kutta method for prediction and Newton iterations for correction. Additionally, power series are used for the end game.
    \item  We use default adaptive precision settings for all the computations, but we added the command 
    $\mathtt{Security Level 1}$ to ensure that the paths are not truncated based on approximations of the endpoints. 
\end{itemize}  
After tracking paths using $\mathtt{Track Type 1}$ we can conclude for the optimization problems given in Eq.\ (\ref{eq:lagrangeplus}) and Eq.~(\ref{eq:lagrange}) with $\circ$ or $\star$  being $\hrectangleblack$, that the solution sets are zero-dimensional. 
For the remaining optimization problem given in  Eq.~(\ref{eq:lagrange}) with $\circ,\star\in\left\{\blacktriangle, \vartriangle \right\}$ we are not able to comment on the existence of higher dimensional components as the computations exceed our computational limits\footnote{\label{note1}
The computations were performed in parallel using a total of 64 threads using AMD Ryzen 7 2700X, 3.7 GHz processor with 63\,973\,484 kB RAM.}.

\item
{\bf Bounds on the number of isolated solutions:}
To ensure that we have obtained all isolated solutions, it is crucial that all paths of the homotopy continuation are tracked without any error or warning message in \texttt{Bertini}. To increase the likelihood of achieving such a ``perfect tracking'', we are aiming to find a partition of the variables (into $m$ groups) such that the resulting \textit{$m$-homogeneous homotopy} has a low number of paths.
According to \cite{sommese2005numerical}, this number of paths to be tracked cannot be lower than the so-called 
BKK bound.\footnote{Note that the polyhedral homotopy is not implemented in $\mathtt{Bertini 1.6v}$ but it is available in \texttt{HC.jl}~\cite{julia}.} 
To find the minimal $m$-homogeneous B\'{e}zout number, we studied all the possible partitions given in ~\cite[p. 72]{bates2013numerically} for the listed number of unknowns in Table~\ref{tabel1}.

For the optimization problems given in  Eq.\ (\ref{eq:lagrange}) with $\circ$ or $\star$  being $\hrectangleblack$ we can see from Table \ref{bounds}
that  the minimal B\'{e}zout bound is implied by $m=1$. For the optimization problem given in Eq.\ (\ref{eq:lagrangeplus}) the minimal B\'{e}zout bound is obtained for $m=2$, where the corresponding grouping of variables is given in Table \ref{bounds}.
For the cases mentioned so far, we succeeded with ``perfect tracking''
using a total degree homotopy ($m=1$) and a multi-homogeneous homotopy ($m=2$), respectively.
 The number of finite isolated solutions for these optimization problems is given in Table~\ref{rootcount}. 

On the other hand, for the optimization problems given in Eq.\ (\ref{eq:lagrange}) with $\circ, \star \in \left\{\blacktriangle, \vartriangle \right\}$, the minimal Bézout bound is also obtained for $m=2$. However, in these cases, we are unable to achieve a ``perfect tracking'' of paths.
The tracking tolerances used for the multi-homogeneous homotopy run are set to 1e-8, but we encountered some failures in tracking all the paths successfully. 
In the numerical example provided in \cite[Section 5.5.2]{bates2013numerically}, a comparison is made between multi-homogeneous and regeneration homotopies. The example demonstrates that the regeneration algorithm is more efficient but misses singular solutions. In order to find all isolated solutions, we employ a regeneration algorithm using the cascade approach (for more details we refer to \cite[pages 89 and 168]{bates2013numerically}). But also with this method, we are not able to perform ``perfect tracking'' as can be seen in the number of path failures given in Table ~\ref{rootcount}. 
 Therefore, the problem of computing all the generic finite solutions without encountering any path failures remains unresolved.
\end{enumerate}

\begin{table}
\caption{Comparison of numbers of paths to be tracked by polyhedral and $m$-homogeneous homotopies Eqs.\ (\ref{eq:lagrange}) and (\ref{eq:lagrangeplus}).}
\centering
\begin{tabular}{|l||l|l|l|l|}
\hline
Interpretation &  \begin{tabular}[c]{@{}l@{}}   BKK \\ bound \end{tabular}  & $m$ & \begin{tabular}[c]{@{}l@{}} $ {m}$-homogeneous \\   B\'{e}zout bound \end{tabular} & grouping \\ \hline
 $G_{\rectangleblack}^{\rectangleblack}(\mathbf{K'})$              &        150          &  2 &      {324}     & \begin{tabular}[c]{@{}l@{}} $\left\{c_{4}, d_{4},c_{5},d_{5} \right\}$, \\ $\left\{\kappa, \lambda \right\} $  \end{tabular} \\ \hline
 \begin{tabular}[c]{@{}l@{}}  $G^{\blacktriangle}_{\rectangleblack} (\mathbf{K'}),   G_{\rectangleblack}^{\vartriangle}(\mathbf{K'})$, \\  inversions of $G^{\rectangleblack}_{\blacktriangle} (\mathbf{K'}), G_{\vartriangle}^{\rectangleblack}(\mathbf{K'})$      \end{tabular}        &            {1\,845}      &   1 &  2\,187         &  $\left\{c_{4}, d_{4},c_{5},d_{5}, c_{6},d_{6}, \lambda \right\}$     \\ \hline
  \begin{tabular}[c]{@{}l@{}}  
 $G_{\blacktriangle}^{\blacktriangle}(\mathbf{K'})$, $G_{\vartriangle}^{\blacktriangle}(\mathbf{K'})$, \\ $G_{\vartriangle}^{\blacktriangle}(\mathbf{K'})$, $G_{\vartriangle}^{\vartriangle}(\mathbf{K'})$            \end{tabular}      &     148\,992           &   2 &  236\,196        &  \begin{tabular}[c]{@{}l@{}}   $\left\{c_{2}, c_{3}, d_{3}, c_{4}, d_{4},c_{5},d_{5}, c_{6},d_{6}\right\}$, \\ $\left\{\lambda \right\}$   \end{tabular}     \\  \hline
\end{tabular}
\label{bounds}
\end{table}

\begin{table}[]
\caption{Summary of generic finite solutions obtained from the ab-initio phase to $L$ of Eqs.\ (\ref{eq:lagrange}) and (\ref{eq:lagrangeplus}).}
\centering
\begin{tabular}{|l||l|l|}
\hline
Interpretation &  \begin{tabular}[c]{@{}l@{}} $\#$ finite solutions \\ over $\mathbb{C}$ (\texttt{Bertini})\end{tabular} & $\#$ path failures \\ \hline
 $G_{\rectangleblack}^{\rectangleblack}(\mathbf{K'})$              &        76          &        0      \\ \hline
  $G^{\blacktriangle}_{\rectangleblack} (\mathbf{K'}),   G_{\rectangleblack}^{\vartriangle}(\mathbf{K'})$, inversions of $G^{\rectangleblack}_{\blacktriangle} (\mathbf{K'}), G_{\vartriangle}^{\rectangleblack}(\mathbf{K'})$               &            858      &  0            \\ \hline
 $G_{\blacktriangle}^{\blacktriangle}(\mathbf{K'})$              &       {15\,552}           &   2\,613            \\ \hline
 $G_{\vartriangle}^{\blacktriangle}(\mathbf{K'})$, 
 inversion of $G^{\vartriangle}_{\blacktriangle}(\mathbf{K'})$
              &        {17\,608}          &       {1\,167}        \\ \hline
  $G_{\vartriangle}^{\vartriangle}(\mathbf{K'})$             &        {17\,824}          &     {301}          \\ \hline
\end{tabular}
\label{rootcount}
\end{table}

\subsubsection{User-defined homotopy phase}\label{secondstep} 

The input for this step is the manipulator's geometry given by the real values $x_2,x_3,y_3,x_5,x_6,y_6$ and a 1-parametric motion with parameter $\phi\in[u;v]$ given as :
\begin{equation}\label{motion_phi}
\mathbf{k}_{j}(\phi) =
\mathbf{R}(\phi) \mathbf{p}_j+ \mathbf{t}(\phi)
\quad \text{for} \quad j=4,5,6 
\end{equation}
according to Eq.~(\ref{initial}). We discretize the interval of $\phi$ into a user-defined number $n$ of poses. The following procedure has to be done for each of these $n$ poses.

For the construction of the user-defined homotopy of the polynomial system, $\mathcal{D}$ we are using the following three strategies:

\begin{itemize}
\item
User-defined homotopy of bar energy: The bar energy $U\left(\vert_{ij},\vert'_{ij}\right)$ of Eq.\ (\ref{metricdp3}) is transformed by replacing the lengths $\ell\left(\mathbf{k}_{i}^\mathbb{C},\mathbf{k}_{j}^\mathbb{C}\right)$  of the generic complex configuration 
$\mathbf{K}^{\mathbb{C}}$ by
\begin{equation}\label{user1}
\ell(\mathbf{k}_{i},\mathbf{k}_{j}) +{h} \left[\ell\left(\mathbf{k}_{i}^\mathbb{C},\mathbf{k}_{j}^\mathbb{C}\right)-\ell(\mathbf{k}_{i},\mathbf{k}_{j})\right],
\end{equation}
where 
$\ell(\mathbf{k}_{i},\mathbf{k}_{j})$ is the corresponding length of the considered pose and
$h$ the homotopy parameter running from $1$ to $0$. 
\item
User-defined homotopy of plate energy: For the homotopic transformation of the plate energy $U\left(\blacktriangle_{ijk}, \blacktriangle'_{ijk}\right)$ of Eq.\ (\ref{energy:plate}) we replace the vectors 
$\mathbf{k}_{j}^\mathbb{C}-\mathbf{k}_{i}^\mathbb{C}$
and 
$\mathbf{k}_{k}^\mathbb{C}-\mathbf{k}_{i}^\mathbb{C}$
used for the computation of the matrix $\mathbf{F}$ (cf.\ Eq.\ (\ref{triangle1})) by
\begin{equation}\label{triangle}
(\mathbf{k}_{j}-\mathbf{k}_{i})+{h}\left[\left(\mathbf{k}_{j}^\mathbb{C}-\mathbf{k}_{i}^\mathbb{C}\right)-(\mathbf{k}_{j}-\mathbf{k}_{i})
\right]\quad \text{and} \quad
(\mathbf{k}_{k}-\mathbf{k}_{i})+{h}\left[\left(\mathbf{k}_{k}^\mathbb{C}-\mathbf{k}_{i}^\mathbb{C}\right)-(\mathbf{k}_{k}-\mathbf{k}_{i})
\right],
\end{equation}
respectively. In addition, we use the replacements given in Eq.\ (\ref{user1}) for the homotopic transformation of $\text{Vol}_{\blacktriangle_{ijk}}$ given in Eq.\ (\ref{eq:volume}).
\item
User-defined homotopy in case of undeformed base: 
In this case, we make a homotopy between the base points; i.e.\
\begin{equation}
  \mathbf{k}_{i}^\mathbb{C} \mapsto \mathbf{k}_{i}({h}):=
{{\mathbf{k}}}_{i} + {h}\left(\mathbf{k}_{i}^\mathbb{C} - {{\mathbf{k}}}_{i}\right)
\quad \text{for} \quad i=1,2,3
\end{equation}
which implies a homotopy for the deformed lengths of the 
three legs in Eq. (\ref{metricdp3}); i.e.\ 
$\ell(\mathbf{k}_{i}',\mathbf{k}_{i+3}')$ is replaces by
\begin{equation}\label{homtopy_deform}
    \ell\left(\mathbf{k}_{i}({h}),\mathbf{k}_{i+3}'\right) \quad \text{for} \quad  i=1,2,3. 
\end{equation}
\end{itemize}

We can apply the above homotopies to the intrinsic metrics given in Eqs.\ (\ref{In1}-\ref{In9}) which also effects the Lagrangian functions $L$ given in Eq.\ (\ref{eq:lagrange}) and Eq.\ (\ref{eq:lagrangeplus}), respectively. Their partial derivatives yield the user-defined homotopy of the polynomial system $\mathcal{D}$, which is passed on to \texttt{Bertini} to track the finite isolated solutions of the ab-initio phase to the given manipulator configuration. 
If there are path failures during this tracking, one has to tighten the tracking tolerances and recompute for the failed paths\footnote{Currently, the failed paths are recomputed manually.}.

\paragraph{{\bf Discussion of alternative approaches involving \texttt{Paramotopy}}} 
Based on an input file, the software \texttt{Paramotopy} \cite{bates2018paramotopy}, executes a two-step process for solving the system of polynomial equations at a generic parameter point. In the first step, \texttt{Bertini} is called to find the solutions to the equations. Subsequently, in the second step, \texttt{Paramotopy} tracks these solutions to all $n$ parameter values by again invoking \texttt{Bertini}. However, this approach presents certain limitations when applied to our optimization problems: 

The undeformed leg lengths $\ell({\mathbf{k}_{i}},{\mathbf{k}_{j}})$ with $(i,j) \in I_{1}$ given in Eq.~(\ref{length}) depend on the pose of the manipulator. For executing \texttt{Paramotopy} in every of the $n$ discretized poses along the given one parametric motion, one has to make the reparametrization 
\begin{equation}
  \phi= v-(v-w)(1-h),
\label{complex}
\end{equation}
 where $v$ and $w$ are the lower and upper bounds of $\phi$'s interval. It can be noticed that the homotopy parameter $h$ appears under the square root of these undeformed leg lengths. Note that the parameter homotopy approach only works if the parameterized system of equations is analytic in the homotopy parameter.
 To circumvent this problem, one can replace the corresponding undeformed leg lengths with dummy variables $\Lambda_{ij}$ and add the following three equations 
  \begin{equation}\label{extra}    \ell({\mathbf{k}_{i}},{\mathbf{k}_{j}})^2-\Lambda_{ij}^2=0
    \end{equation}
 for $(i,j) \in I_{1}$ to the set $\mathcal{D}$ of equations. Instead of solving for the unknowns listed in Table \ref{tabel1}, one ends up with additional three equations formulated by Eq.~(\ref{extra}) for the optimization problems of Eq.~(\ref{eq:lagrange}) and Eq.~(\ref{eq:lagrangeplus}), which blows up the system. 
 
In addition to this drawback, users are required to switch between different program interfaces and 
create various input format files for both Bertini and Paramotopy. This can be cumbersome and time-consuming, leading to a less efficient workflow. Moreover, the entire process generates numerous files during its execution, contributing to the accumulation of data in the computer memory and potentially causing memory overflow issues.

 Further, if there are path failures during step 1 of \texttt{Paramotopy}, one has to tighten tolerances and do a recomputation until no path failures appear, which can be computationally very expensive. This can be avoided by using the idea presented in \cite[Section 5]{akapilavai2022}. 

\subsubsection{Post-processing phase}\label{postprocess}
In the post-processing step, we import for each pose the real solutions obtained from \texttt{Bertini} to \texttt{Maple}. 
Then we filter out the local minima  by checking the eigenvalues of the (bordered) Hessian matrix (cf.\ \cite{borderhessian}) and sort them from minimum to maximum by evaluating the corresponding 
metric $D_\star^\circ(\mathbf{K},\mathbf{K}')$, which yields the ordered list  $G_\star^\circ(\mathbf{K}'_1),G_\star^\circ(\mathbf{K}'_2), \ldots$.

Now we run the user-defined homotopies for the identification of the closest singular configuration as described in Section~\ref{non}. Using the notation of this subsection the condition $G_{\star}^{\circ}(\mathbf{K}'')=G_{\star}^{\circ}(\mathbf{K}'_i)$ is computationally checked by
\begin{equation}
(\mathbf{K}''-\mathbf{K}'_i)^T(\mathbf{K}''-\mathbf{K}'_i)<\delta.
    \label{eq:metric10}   
\end{equation}
Note that the threshold $\delta$ in Eq.~(\ref{eq:metric10}) has to be chosen in dependence of the computer precision used in \texttt{Maple} and \texttt{Bertini}, respectively.

\begin{rmk}
Note that due to numerical inaccuracies during the computational process $\mathbf{K}''$ may drift slightly to the complex domain. In this case one either replace  Eq.\ (\ref{eq:metric10}) by
%\begin{equation}
$(\mathbf{K}''-\mathbf{K}'_i)^T(\overline{\mathbf{K}}''-\mathbf{K}'_i)<\delta$
 %   \label{eq:metric10_new}   
%\end{equation}
or one increases the precision of the computation until $\mathbf{K}''$ is real.
\hfill $\diamond$
\end{rmk}

\subsection{Computing the closest regular point on the collinearity variety}\label{sec:proc_main4}

In this section, we discuss the computations related to the optimization problems given in Eqs.\ (\ref{eq:lagrange1}) and (\ref{eq:lagrange2}).
 Once again, when using $\mathtt{Track Type 1}$ to optimize the problems given in Eq. (\ref{eq:lagrange1}),  we are unable to comment on the existence of higher-dimensional components due to computational limitations\footref{note1}. However, for the optimization problem given in Eq.\ (\ref{eq:lagrange2}), we can conclude that the solution set is zero-dimensional. Only for this case, we are also able to perform a ``perfect tracking'' (cf.\ Table \ref{collinearityresult}) by means of a  multi-homogeneous homotopy approach using the same grouping as specified in the $5^{th}$ column of Table \ref{bounds}. 
Using the obtained generic solutions we can compute 
 the closest regular point on the collinearity variety applying the procedure described in the Subsections
\ref{secondstep} and \ref{postprocess}. 

\begin{table}[h!]
\caption{Summary of generic finite solutions obtained from the ab-initio phase to $L$ of Eqs.\ (\ref{eq:lagrange1}) and (\ref{eq:lagrange2}).}
\centering
\begin{tabular}{|l||l|l|l|}
\hline
Interpretation &
  \begin{tabular}[c]{@{}l@{}} $\#$ finite solutions\\over $\mathbb{C}$(\texttt{Bertini})\end{tabular} & $\#$ path failures 
 \\ \hline
$G_{\rectangleblack}^{\vartriangle}(\mathbf{K}')$, inversion of $G_\vartriangle^{\rectangleblack}(\mathbf{K}')$  & 206 & 0 \\ \hline
$G_{\vartriangle}^{\blacktriangle}(\mathbf{K}')$, inversion of $G_{\blacktriangle}^{\vartriangle}(\mathbf{K}')$  
   & 2\,696     &  1\,863    \\  \hline
$G^{\vartriangle}_{\vartriangle}(\mathbf{K}')$ and its inversion &  3\,598     &  549   \\  \hline
\end{tabular}
\label{collinearityresult}
\end{table}

But there is maybe no need for doing these computations as we can give lower bounds for the distance to regular points of the collinearity variety. Only in the case where these bounds are below the distances computed in Section \ref{sec:proc_main} with respect to the corresponding metric, we have to perform the aforementioned computations. In the following, we describe how these lower bounds can be obtained.

\subsubsection{Lower bounds for  the distance to regular points of the collinearity variety}\label{boundcollinear}

The minimum deformation energy of the pin-jointed triangular bar structure $\vartriangle_{ijk}$ deforming into a line-segment is denoted by $U_{\text{min}}\left(\vartriangle_{ijk}, \text{coll}(\vartriangle_{ijk}')\right)$ with $(i,j,k)=(1,2,3)$ or $(i,j,k)=(4,5,6)$, respectively. The computation is straightforward; i.e.\ we compute the partial derivatives of 
\begin{equation}\label{eq:triener}
U\left(\vert_{ij},\vert'_{ij}\right) +
U\left(\vert_{ik},\vert'_{ik}\right) +
U\left(\vert_{jk},\vert'_{jk}\right),
\end{equation}
with respect to $c_j$ and $c_k$ as we can set
$c_i=d_i=d_j=d_k=0$ without loss of generality. 
Direct computations show that both equations split up into a line and a conic section considered in the  $(c_j,c_k)$-pane, thus in total nine critical points exist over $\mathbb{C}$.
One of them implies the lowest value for  Eq.\ (\ref{eq:triener}) which we define as
$U_{\text{min}}\left(\vartriangle_{ijk}, \text{coll}(\vartriangle_{ijk}')\right)$. 

Now the lower bounds can easily be obtained by assuming that besides the deformation of  $\vartriangle_{ijk}$  into a line segment no further structural components are deformed. This implies the following expressions:

\begin{align}
    B_{\rectangleblack}^{\textnormal{coll}(\vartriangle)}(\mathbf{K},\mathbf{K}') &=\frac{1}{\Omega_{I_2}} \left[U_\text{min}\left(\vartriangle_{456}, \textnormal{coll}(\vartriangle_{456}')\right)\right],
    \label{b1} \\
    B_{\textnormal{coll}(\vartriangle)}^{\rectangleblack}(\mathbf{K},\mathbf{K}') &= 
    \frac{1}{\Omega_{I_3}}\left[ U_\text{min}\left(\vartriangle_{123}, \text{coll}(\vartriangle_{123}')\right)\right],
    \label{b2} \\
    B_{\blacktriangle}^{\textnormal{coll}(\vartriangle)}(\mathbf{K},\mathbf{K}') &= B_{\vartriangle}^{\textnormal{coll}(\vartriangle)}(\mathbf{K},\mathbf{K}')=\frac{1}{\Omega_{I_4}}\left[U_\text{min}\left(\vartriangle_{456}, \textnormal{coll}(\vartriangle_{456}')\right)\right],
    \label{b3} \\
    B_{\textnormal{coll}(\vartriangle)}^{\blacktriangle}(\mathbf{K},\mathbf{K}') &= B_{\textnormal{coll}(\vartriangle)}^{\vartriangle}(\mathbf{K},\mathbf{K}')=\frac{1}{\Omega_{I_4}}\left[U_\text{min}\left(\vartriangle_{123}, \textnormal{coll}(\vartriangle_{123}')\right)\right],
     \label{b4} %\\
\end{align}

where $\Omega_{I_2},  \Omega_{I_3},  \Omega_{I_4}$ are given by Eqs.\ (\ref{index2}) -- (\ref{index4}). Note that in Eqs.\ (\ref{b3}--\ref{b4}) we also obtain lower bounds for the interpretations where we do not have any guaranteed information about the set of finite solutions in the ab-initio phase.

\subsection{Computing the closest singular point of the singularity variety}\label{sec:proc_special1}
In this section, we discuss the distance computations for singular points on the singularity variety $V=0$, formulated in Section~\ref{sec:singvar}.
When using $\mathtt{Track Type 1}$ to optimize the problems presented in Eqs.\ (\ref{eq:f}) and (\ref{eq:fplus}), which are related to singular points on $V=0$, we find that the solution sets are zero-dimensional for the respective number of unknowns listed in Table~\ref{tabel1b}. We have successfully achieved ``perfect path tracking''  for these optimization problems by employing a total degree homotopy. Table \ref{spsvrootcount} provides the generic finite solutions for these optimization problems.
These solutions can then be used to proceed as described in Subsections \ref{secondstep} and \ref{postprocess}.

\begin{table}[h!]
\caption{Summary of generic finite solutions obtained from the ab-initio phase for Eqs.~(\ref{eq:f}) and (\ref{eq:fplus}). 
}
    \centering
\begin{tabular}{|l||l|l|l|}
\hline
Interpretation &  \begin{tabular}[c]{@{}l@{}} $\#$ finite solutions \\ over $\mathbb{C} (\mathbb{R})$ (\texttt{Bertini})\end{tabular} & $\#$ path failures \\ \hline
$G_{\rectangleblack}^{\rectangleblack}(\mathbf{K'})$ &     3 (0) & 0  \\ \hline
$G^{\blacktriangle}_{\rectangleblack} (\mathbf{K'})$, $G_{\rectangleblack}^{\vartriangle}(\mathbf{K'})$,
inversions of
$ G^{\rectangleblack}_{\blacktriangle} (\mathbf{K'})$, $G_{\vartriangle}^{\rectangleblack}(\mathbf{K'})$
& 27 (0) & 0  \\ \hline

  $G_{\blacktriangle}^{\blacktriangle}(\mathbf{K'})$, $G_{\vartriangle}^{\vartriangle}(\mathbf{K'})$,  $G_{\blacktriangle}^{\vartriangle}(\mathbf{K'})$, inversion of 
 $G_{\vartriangle}^{\blacktriangle}(\mathbf{K'})$ 
    &   242 (1) & 0 \\ \hline
\end{tabular}
\label{spsvrootcount}
\end{table}

Note that these computations only need to be performed if the lower bound $B_{\textnormal{coll}(\star)}^{\textnormal{coll}(\circ)}(\mathbf{K},\mathbf{K}')$ for the distance to singular points of the singularity variety is below the minimum of distances obtained in Sections \ref{sec:proc_main} and \ref{sec:proc_main4}
with respect to the metrics $D_{\star}^{\circ}(\mathbf{K},\mathbf{K}')$. In the following, we will describe how these lower bounds are computed.

\subsubsection{Lower bounds for  the distance to singular points of singularity variety}\label{test}
The minimum deformation energy of the triangular plate $\blacktriangle_{ijk}$ deforming into a line segment is denoted by $U_{\text{min}}\left(\blacktriangle_{ijk}, \text{coll}(\blacktriangle_{ijk}')\right)$. By direct computations, it can easily be shown that this energy equals: 
\begin{equation}
  U_{\text{min}}\left(\blacktriangle_{ijk}, \text{coll}(\blacktriangle_{ijk}')\right)= \tfrac{1}{8}\text{Vol}_{\blacktriangle_{ijk}},
  \label{boundplate}
\end{equation}
with $\text{Vol}_{\blacktriangle_{ijk}}$ from Eq.\ (\ref{eq:volume}).
Based on  Eq.~(\ref{eq:triener}) and Eq.~(\ref{boundplate}) the lower bounds $B_{\text{coll}(\star)}^{\text{coll}(\circ)}$ with $\circ, \star \in \left\{\blacktriangle, \vartriangle \right\}$ can be computed as follows:

\begin{align}
    B_{\textnormal{coll}(\blacktriangle)}^{\textnormal{coll}(\blacktriangle)}(\mathbf{K},\mathbf{K}') &=
    \frac{1}{\Omega_{I_4}} \left[
    U_\text{min}\left(\blacktriangle_{123}, coll(\blacktriangle_{123}')\right) +
    U_\text{min}\left(\blacktriangle_{456}, \text{coll}(\blacktriangle_{456}')\right)
    \right],
     \label{b7} \\
 B_{\textnormal{coll}(\blacktriangle)}^{\textnormal{coll}(\vartriangle)}(\mathbf{K},\mathbf{K}') &=
    \frac{1}{\Omega_{I_4}} \left[
    U_\text{min}\left(\blacktriangle_{123}, \text{coll}(\blacktriangle_{123}')\right) +
    U_\text{min}\left(\vartriangle_{456}, \text{coll}(\vartriangle_{456}')\right)
    \right],
     \label{b8} \\
B^{\textnormal{coll}(\blacktriangle)}_{\textnormal{coll}(\vartriangle)}(\mathbf{K},\mathbf{K}') &=
    \frac{1}{\Omega_{I_4}} \left[
    U_\text{min}\left(\blacktriangle_{456}, \textnormal{coll}(\blacktriangle_{456}')\right) +
    U_\text{min}\left(\vartriangle_{123}, \textnormal{coll}(\vartriangle_{123}')\right)
    \right],
     \label{b9} \\
   B_{\textnormal{coll}(\vartriangle)}^{\textnormal{coll}(\vartriangle)}(\mathbf{K},\mathbf{K}') &= 
    \frac{1}{\Omega_{I_4}} \left[
    U_\text{min}\left(\vartriangle_{123}, \textnormal{coll}(\vartriangle_{123}')\right) +
    U_\text{min}\left(\vartriangle_{456}, \textnormal{coll}(\vartriangle_{456}')\right)
    \right],
     \label{b10} 
\end{align}

with $\Omega_{I_4}$ from Eq.~(\ref{index4}).

\subsection{Computing the closest singular point of the collinearity variety}\label{sec:proc_special}
The distance computations for singular points on the collinearity variety $C_{P}=0$ (resp. $C_{B}=0$) are discussed in this section. 
Using the $\mathtt{Track Type 1}$ method, it has been observed that the corresponding optimization problems formulated in  Eqs.\ (\ref{eq:lagrange4}) and (\ref{eq:point1}) have zero-dimensional solution sets for the listed unknowns in Table ~\ref{spcv}. By employing a total degree homotopy, ``perfect path tracking'' have been achieved for these cases.
Table~\ref{solutionscollinear} provides the generic finite solutions for these optimization problems. 
These solutions can then be used to proceed as described in Subsections \ref{secondstep} and \ref{postprocess}.

\begin{table}[h!]
\caption{Summary of generic 
       finite solutions obtained from the ab-initio phase to $L$ of Eqs.~(\ref{eq:lagrange4}) and (\ref{eq:point1}).
}
\centering
\begin{tabular}{|l||l|l|}
\hline
Interpretation &  \begin{tabular}[c]{@{}l@{}} $\#$ finite solutions \\ over $\mathbb{C} (\mathbb{R})$ (\texttt{Bertini})\end{tabular} &  $\#$ path failures  \\ \hline
 $G_{\vartriangle}^{\blacktriangle}(\mathbf{K'})$, $G_{\vartriangle}^{\vartriangle}(\mathbf{K'})$, inversion of $G^{\vartriangle}_{\blacktriangle}(\mathbf{K'})$, $G_{\vartriangle}^{\vartriangle}(\mathbf{K'})$   &   90(1)     &   0     \\ \hline
$G_{\rectangleblack}^{\vartriangle}(\mathbf{K'})$ inversion of  $G^{\rectangleblack}_{\vartriangle}(\mathbf{K'})$                      &      5(0)       &      0    \\ \hline 
 \end{tabular}
\label{solutionscollinear}
\end{table}

 For the distance to singular points of the collinearity variety, we can also give lower bounds denoted by $B_{\circ}^{\bullet}(\mathbf{K},\mathbf{K}')$ and $B^{\star}_{\bullet}(\mathbf{K},\mathbf{K}')$   with $\circ, \star \in \left\{\blacktriangle, \vartriangle, \rectangleblack \right\}$. 
Further computations need to be carried out only in cases where these lower bounds fall below the minimum of distances obtained in Sections \ref{sec:proc_main}, \ref{sec:proc_main4} and \ref{sec:proc_special1} with respect to the corresponding metric. Next, we discuss the computation of the mentioned lower bounds.

\begin{rmk}
If one is interested in computing the closest singular configuration for a 3-RPR manipulator, then the generic finite solutions can be used, which were achieved by ``perfect path tracking'' (cf.\ Tables \ref{rootcount}--\ref{solutionscollinear}). In this case, it is sufficient to run only Sections \ref{secondstep} and \ref{postprocess} of the computation. The generic finite solutions and complex coordinates chosen in Section ~\ref{settings} are provided as supplementary material~\cite{codes}.  \hfill $\diamond$
\end{rmk}

\subsubsection{Lower bounds for  the distance to singular points of collinearity variety}\label{test2}
By setting $\ell(\mathbf{k}'_{i},\mathbf{k}'_{j})$ equal to zero in Eq.\ (\ref{metricdp3}) we obtain the following lower bounds for the cases where the platform and base, respectively, degenerate into a point: 

\begin{align}
  B_{\rectangleblack}^{\bullet}(\mathbf{K},\mathbf{K}') & =\frac{1}{8\Omega_{I_2}}
 \sum_{(i,j)\in I_5 }\ell(\mathbf{k}_{i},\mathbf{k}_{j}),
  && B^{\rectangleblack}_{\bullet}(\mathbf{K},\mathbf{K}') =\frac{1}{8\Omega_{I_3}}
 \sum_{(i,j)\in I_6 }\ell(\mathbf{k}_{i},\mathbf{k}_{j})
 \label{cb1} \\
 B_{\star}^{\bullet}(\mathbf{K},\mathbf{K}') &= \frac{1}{8\Omega_{I_4}}
 \sum_{(i,j)\in I_5 }\ell(\mathbf{k}_{i},\mathbf{k}_{j}),
 &&   B_{\bullet}^{\circ}(\mathbf{K},\mathbf{K}') =\frac{1}{8\Omega_{I_4}}
 \sum_{(i,j)\in I_6 }\ell(\mathbf{k}_{i},\mathbf{k}_{j})
 \label{cb3} 
\end{align}

\noindent with
$ I_5 = \{(4, 5), (4, 6), (5, 6)\} $, 
$I_6 = \{(1, 2), (2, 3), (1, 3)\}$
and 
$\circ, \star \in \left\{\blacktriangle, \vartriangle \right\}$.

\section{Results and Discussion}\label{sec:example}
The presented computational procedure for determining singularity distances is illustrated by the following numerical example, which was also used for the singularity distance computations with respect to extrinsic metrics  \cite{akapilavai2022}. 
 The base and platform anchor points, as well as the one-parametric motion, are defined as follows:
\begin{equation}
\mathbf{k}_{1}=\mathbf{p}_{4}=(0,0), \quad \mathbf{k}_{2}=(11,0), \quad \mathbf{p}_{5}=(3,0), \quad \mathbf{k}_{3}=(5,7),   \quad \mathbf{p}_{6}=(1,2)
\label{design}
\end{equation}
and
\begin{equation}
\mathbf{k}_{i} =
\underbrace{\begin{pmatrix}
\cos{\phi} & -\sin{\phi} \\
\sin{\phi} & \cos{\phi}
\end{pmatrix}}_\mathbf{R}
\mathbf{p}_{i}
+
\underbrace{\frac{1}{2}
\begin{pmatrix}
{11}-6\sin{\phi} \\
{3}-{3}\cos{\phi}
\end{pmatrix}}_\mathbf{t} \quad \text{for} \quad {i}={4,5,6}.
\label{parameter}
\end{equation}
without considering any motion range limitations on P and R joints into account.
Note that the one-parametric motion of the manipulator exhibits two singular configurations at $\phi=0$ and $\phi\approx 3.0356972$ radians.

We discretize the motion parameter $\phi$ into 90 poses within the interval of $0$ to $2\pi$. By defining the required user-defined homotopies from Section ~\ref{secondstep}, we obtain a system of polynomial equations dependent on the homotopy parameter $h$. 
We apply the \texttt{evalf} function in \texttt{Maple} to compute the coefficients of these equations in floating-point arithmetic with a precision of 20 digits using the \texttt{Digits} command. Their solution is done by an adaptive precision path tracking algorithm implemented in  \texttt{Bertini} with a default precision of 94 bits. 
Further, to avoid tracking infinitely long paths, we invoke \texttt{User homotopy:2} in the input files (for details, see~\cite[Sec. 7.5.2]{bates2013numerically}). In order to minimize path failures, we tighten the tracking tolerances to $10e-8$, ensuring a closer tracking of each path. We use generic finite solutions of the ab-initio phase as the \texttt{start solutions} and track critical points for each of the 90 discretized poses. The resulting solution set for each pose is post-processed following the steps outlined in Section~\ref{postprocess}. Finally, for checking the condition given by Eq.~(\ref{eq:metric10}), we set $\delta= 10e-8$.

\begin{figure}[h!]
    \begin{center}
       \begin{overpic}[width=15cm]{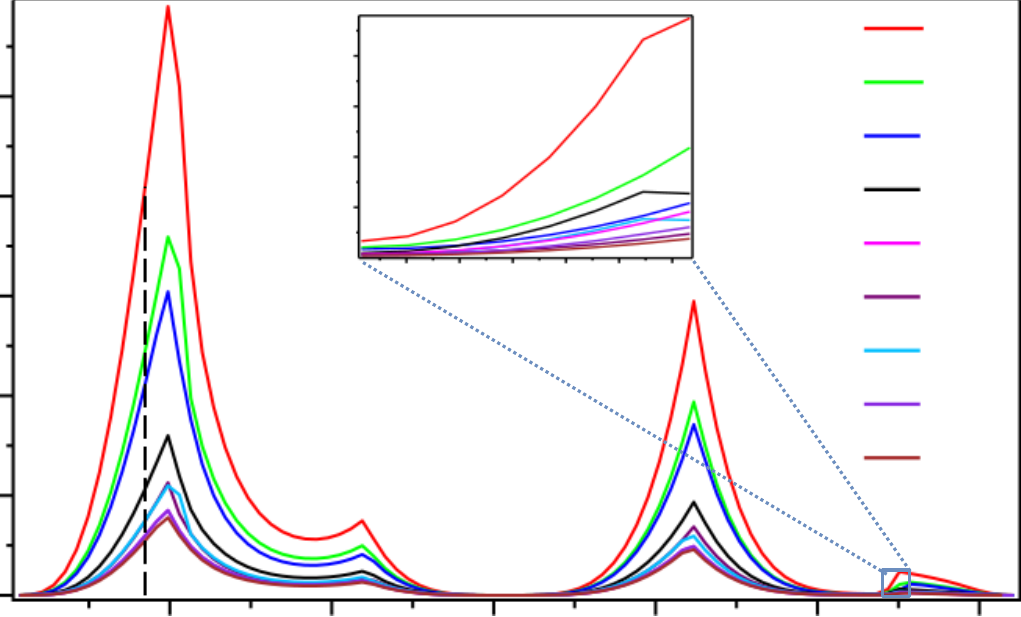}
        \put(85,-1.5){$\phi$ (rad)}
         \put(-8,30){\makebox(0,0){\rotatebox{90}{Distance}}}
       \put (91,57){$G_{\rectangleblack}^{\rectangleblack}(\mathbf{K'})$}
         \put (91,52){$G_{\rectangleblack}^{\blacktriangle}(\mathbf{K'})$}
          \put (91,46){$G_{\rectangleblack}^{\vartriangle}(\mathbf{K'})$}
          \put (91,41){$G^{\rectangleblack}_{\blacktriangle}(\mathbf{K'})$}
           \put (91,35.5){$G^{\blacktriangle}_{\blacktriangle}(\mathbf{K'})$}
           \put (91,31){$G^{\vartriangle}_{\blacktriangle}(\mathbf{K'})$}
           \put (91,25){$G^{\rectangleblack}_{\vartriangle}(\mathbf{K'})$}
            \put (91,20){$G^{\blacktriangle}_{\vartriangle}(\mathbf{K'})$}
            \put (91,15){$G^{\vartriangle}_{\vartriangle}(\mathbf{K'})$}
      \put (-1.7,1){0}
       \put (16,-2.5){1}  
        \put (32,-2.5){2}
          \put (48,-2.5){3}
          \put (64,-2.5){4}
          \put (80,-2.5){5}
          \put (96,-2.5){6}
           \put (96,-2.5){6}
            
            \put (-5.5,50){$0.01$}
            \put (-6.5,40){$0.008$}
            \put (-6.5,30.5){$0.006$}
            \put (-6.5,21){$0.004$}
             \put (-6.5,11){$0.002$}
            \put (27,39.5){$0.0001$}
            \put (27,43.5){$0.0002$}
             \put (27,48.5){$0.0003$}
             \put (27,54){$0.0004$}
             \begin{footnotesize}
              \put (39,32.5){$5.4$}
                \put (43,32.5){$5.42$}
                 \put (48.5,32.5){$5.44$}
                 \put (54,32.5){$5.46$}
                  \put (59.6,32.5){$5.48$}
                  \put (64.2,32.5){$5.50$}
                   \end{footnotesize}
        \end{overpic} \medskip
     \caption{Distances to the closest regular point on  $V=0$  with respect to the different intrinsic metrics.}
      \label{measure2}
    \end{center}
\end{figure}

Fig.~\ref{measure2} compares the distances to the closest regular points on the singularity variety $V=0$ obtained by the presented nine intrinsic metrics. Furthermore, we inserted eight additional discrete poses in the interval between $\phi = 5.38306606$ and $\phi = 5.5066118442$ radians to smooth out the appearance of the graph in this region (see the zoomed-in blue box in Fig.~\ref{measure2}).
It is worth noting that for the interpretations $G_{\star}^{\circ}(\mathbf{K}')$ $\circ, \star \in \left\{\blacktriangle, \vartriangle \right\}$ 
the computations were based on the generic finite solutions with path failures (cf.\ Table~\ref{rootcount}). 
 Therefore, for these cases, we cannot guarantee that the obtained intrinsic distances to the regular points of $V=0$  are global minima. However, the trend of the corresponding graphs in Fig.~\ref{measure2} aligns with the rest of the cases.  

 Note that the cusps
in the graphs of Fig.~\ref{measure2} belong to the cut loci of the intrinsic distance functions; i.e.\ in the corresponding configurations the closest singular configuration is not determined uniquely (there exist at least two closest singular configurations).

\begin{figure}[h!]
\begin{center}
\begin{overpic}
    [width=45mm]{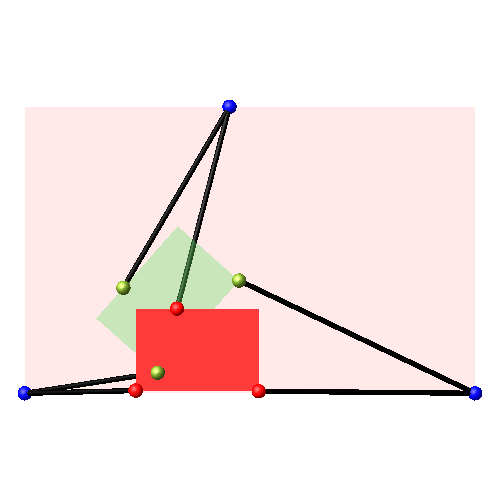}
\put (24,4){(a) $G_{\rectangleblack}^{\rectangleblack}(\mathbf{K,K'})$}    
\end{overpic}
\quad
\begin{overpic}
    [height=45mm]{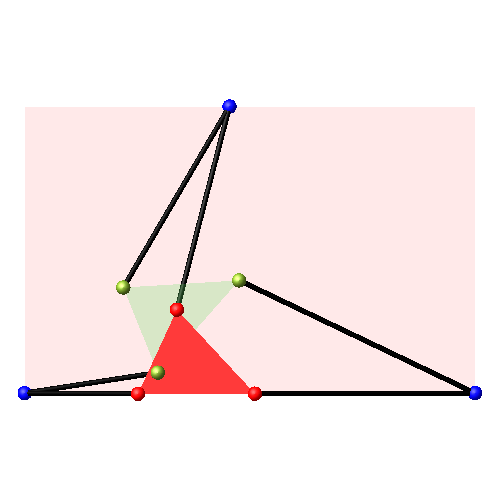}
    \end{overpic}
    \put (-80,6){(b) $G_{\rectangleblack}^{\blacktriangle}(\mathbf{K,K'})$} 
\quad
\begin{overpic}
    [height=45mm]{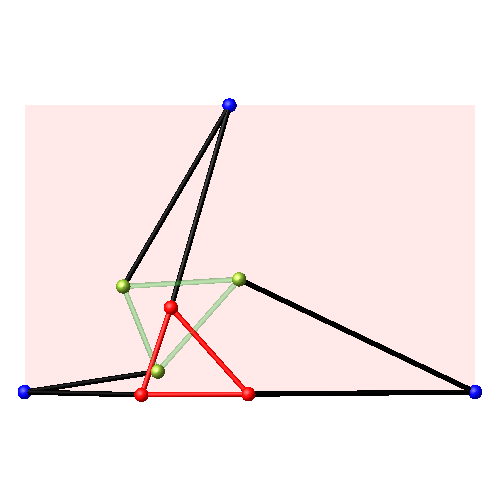}
\end{overpic} 
 \put (-80,6){(c) $G_{\rectangleblack}^{\vartriangle}(\mathbf{K,K'})$} 
\newline
\begin{overpic}
[height=45mm]{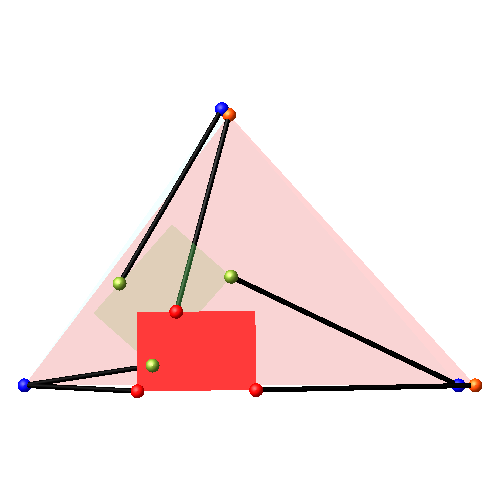}
\end{overpic}
    \put (-80,0){(d) $G^{\rectangleblack}_{\blacktriangle}(\mathbf{K,K'})$}   
\quad
\begin{overpic}
    [height=45mm]{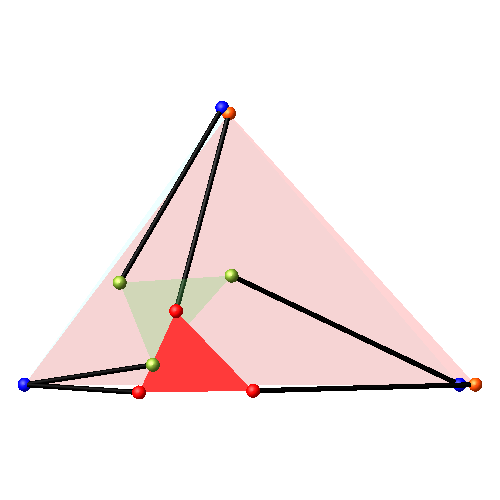}
   \put (40,0){(e) $G_{\blacktriangle}^{\blacktriangle}(\mathbf{K,K'})$}   
\end{overpic}
\quad
\begin{overpic}
    [height=45mm]{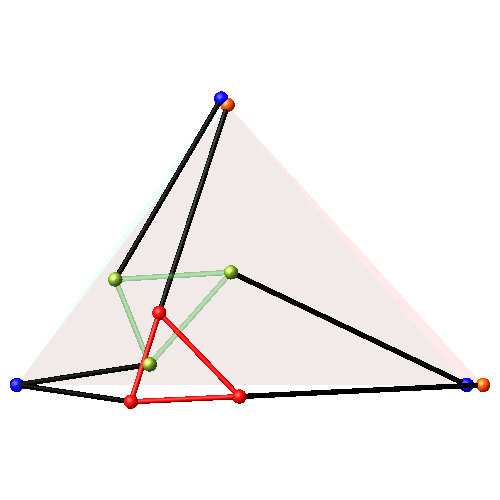}
     \put (40,0){(f) $G_{\blacktriangle}^{\vartriangle}(\mathbf{K,K'})$} 
\end{overpic} 
\newline
\begin{overpic}
    [height=45mm]{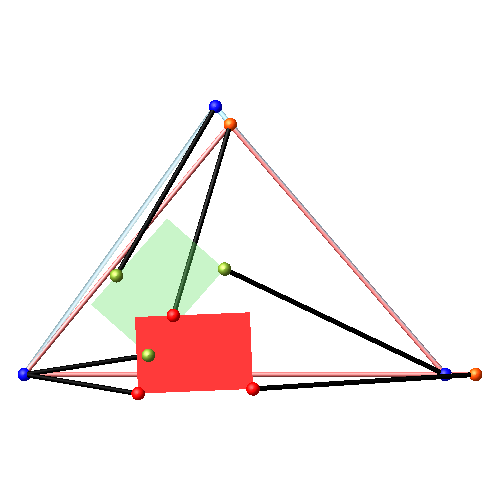}
   \put (30,0){(g) $G^{\rectangleblack}_{\vartriangle}(\mathbf{K'})$}   
\end{overpic}
\quad
\begin{overpic}
    [height=45mm]{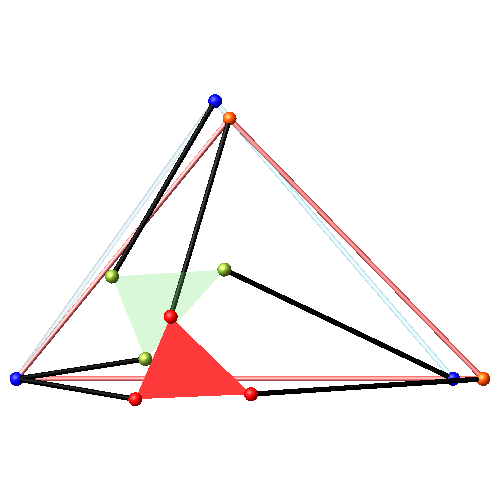}
 \put (30,0){(h) $G^{\blacktriangle}_{\vartriangle}(\mathbf{K,K'})$}    
\end{overpic}
\quad
\begin{overpic}
    [height=45mm]{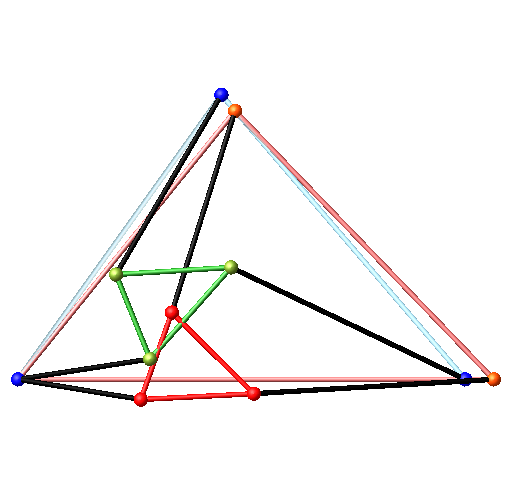}
 \put (30,0){(i) $G^{\vartriangle}_{\vartriangle}(\mathbf{K,K'})$}     
\end{overpic}
\end{center}
\caption{
 Given configuration (green) for $\phi=0.8471710528$ radians and the singular configuration (red), which corresponds to the closest regular point on $V=0$.
}
\label{fig:my_label}
\end{figure}

 In Fig.~\ref{fig:my_label}, the
 singular configurations of Fig.~\ref{measure2} are illustrated for the pose $\phi=0.8471710528$ (indicated by the black dashed line). Animations showing the singular configuration along the one-parametric motion are provided as supplementary material \cite{codes}.

\begin{figure}[t]
\begin{center}
\begin{overpic}
   [height=40mm]{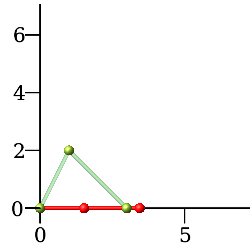} 
    \begin{small}
    \put (2,-12){(a) $U_{\text{min}}\left(\vartriangle_{4,5,6}, \text{coll}(\vartriangle_{4,5,6}')\right)$}
    \put (5,24){$\mathbf{p}'_{4}$}
    \put (30,5){$\mathbf{p}'_{6}$}
   \put (55,5){$\mathbf{p}'_{5}$}
     \end{small}
\end{overpic}
\,
\begin{overpic}
     [height=40mm]{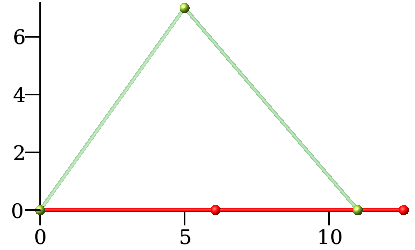}
     \begin{small}
      \put (18,-7){(b) $U_{\text{min}}\left(\vartriangle_{1,2,3}, \text{coll}(\vartriangle_{1,2,3}')\right)$}
    \put (3,14){$\mathbf{k}'_{1}$}
    \put (49,2){$\mathbf{k}'_{3}$}
     \put (94,2){$\mathbf{k}'_{2}$}
     \end{small}
\end{overpic}
\,
\begin{overpic}
    [width=43mm]{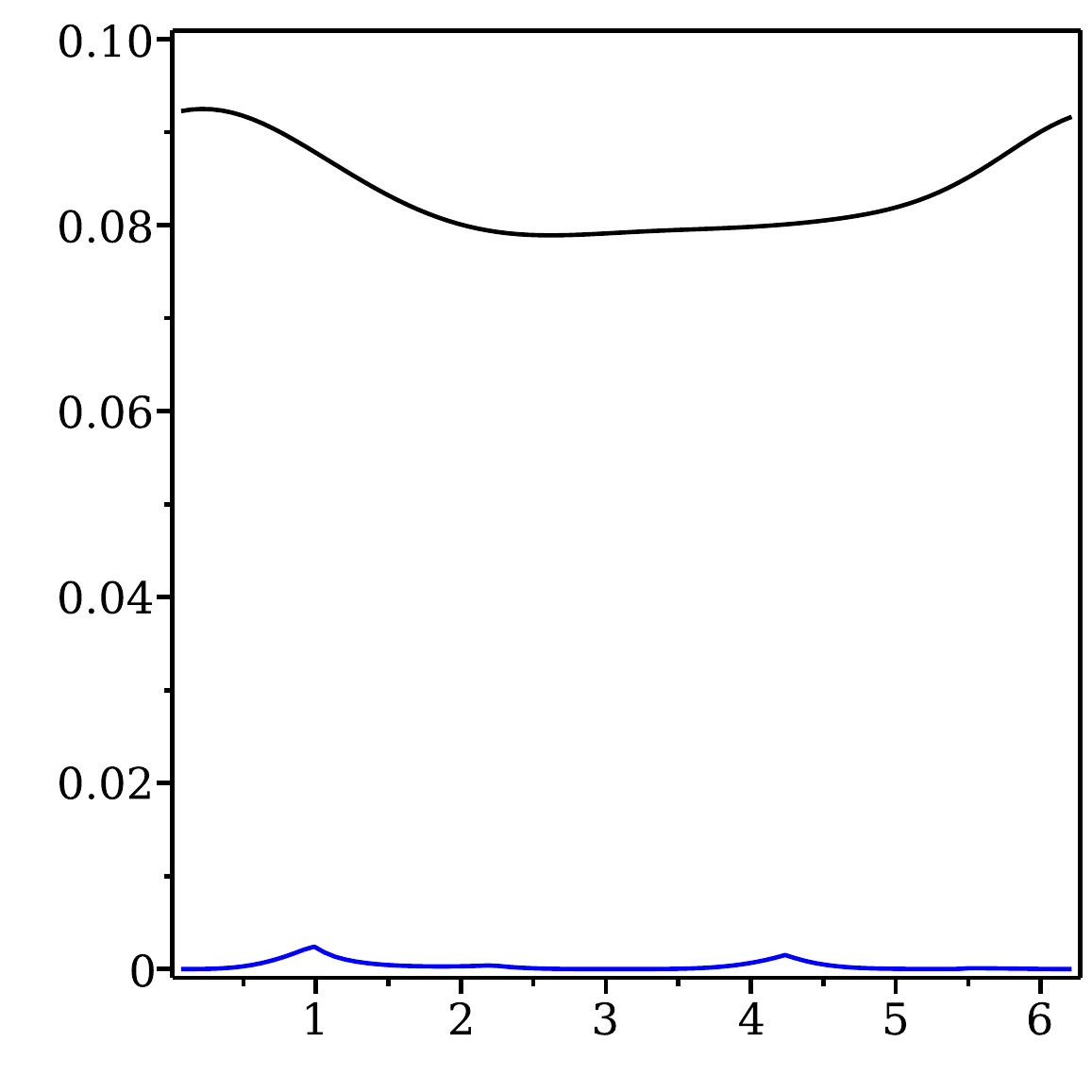}
      \begin{small}
     \put(48,-11) {(c)}   
     \put(32,86){$B_{\text{coll}(\blacktriangle)}^{\text{coll}(\blacktriangle)}((\mathbf{K}, \mathbf{K}')$}
      \end{small}
    \begin{scriptsize}
    \put(0,50){\makebox(0,0){\rotatebox{90}{Distance}}}
     \put(80,-1){$\phi$ (rad)}
\end{scriptsize}
\end{overpic}
\end{center}
\bigskip
\caption{The collinear configurations (red) of the given (green) pin-jointed bar structure associated with the platform (a) and base (b), respectively, causing the minimum deformation energy.
(c) Comparison of the distance to the closest regular point on $V=0$ (blue) and the lower bound for the distance to the closest singular point on the singularity variety (black).}
\label{bounds1}
\end{figure}

\paragraph{Lower bound tests} 
The lower bounds discussed in Subsections \ref{boundcollinear}, \ref{test}, and \ref{test2} have to be computed for each of the $90$ discretized poses.
For this we need the minimum plate energies $U_{\text{min}}\left(\blacktriangle_{ijk}, \text{coll}(\blacktriangle_{ijk}')\right)$ 
for  $(i,j,k)=(1,2,3)$ and $(i,j,k)=(4,5,6)$, which 
can easily be obtained by plugging the values of the example under consideration into Eq.\ (\ref{boundplate}). For the platform and base, we obtain the values $1.008061$ and $3.602733$, respectively.
Moreover, we need the minimum energies $U_{\text{min}}\left(\vartriangle_{ijk}, \text{coll}(\vartriangle_{ijk}')\right)$ of 
the platform and the base interpreted as pin-jointed triangle bar structure, which can be determined by following the procedure described in Section \ref{boundcollinear}. Doing this we get:
\begin{align}
U_{\text{min}}\left(\vartriangle_{123}, \text{coll}(\vartriangle_{123}')\right) &= \tfrac{356411+30267\sqrt{85}-(6149\sqrt{85}-32456)\sqrt{74}}{617764}, \\
U_{\text{min}}\left(\vartriangle_{456}, \text{coll}(\vartriangle_{456}')\right)& = \tfrac{291+127\sqrt{5}-(129\sqrt{5}-197)\sqrt{2}}{2018},
\end{align}
where the corresponding configurations are illustrated in  Fig.\ \ref{bounds1}(a,b).

\begin{figure}[t]
\begin{center}
\begin{overpic}
  [height=45mm]{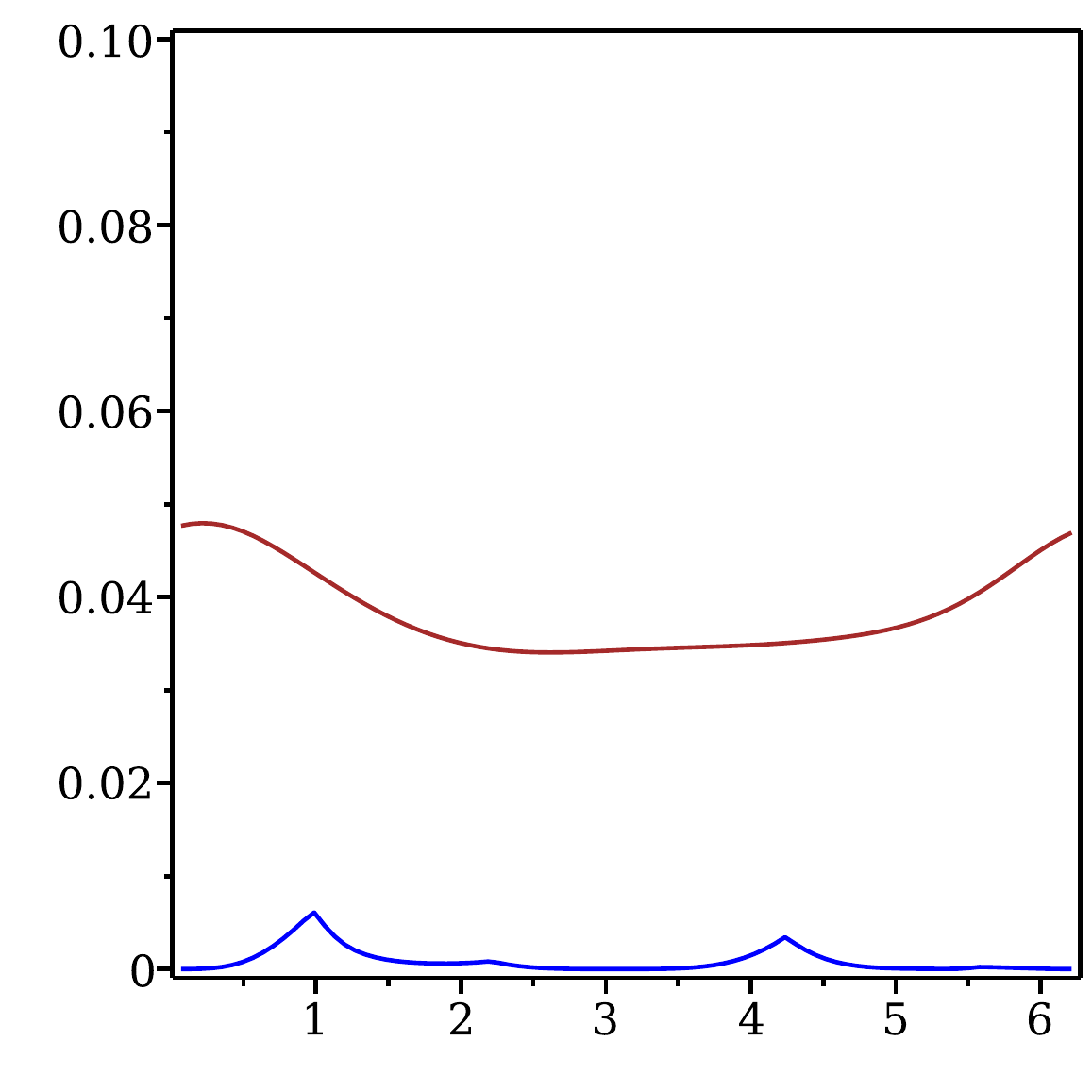}
    \begin{small}
     \put(33,-7) {(a)}  
      \put(46,-7){{$G_{\rectangleblack}^{\vartriangle}(\mathbf{K,K'})$}}  
     \put(40,48){$B_{\rectangleblack}^{\bullet}(\mathbf{K},\mathbf{K'})$}
      \end{small}
        \begin{scriptsize}   
   \put(0,50){\makebox(0,0){\rotatebox{90}{Distance}}}
     \put(85,0.5){$\phi$ (rad)}
   \end{scriptsize}
\end{overpic}
\quad
\begin{overpic}
    [width=45mm]{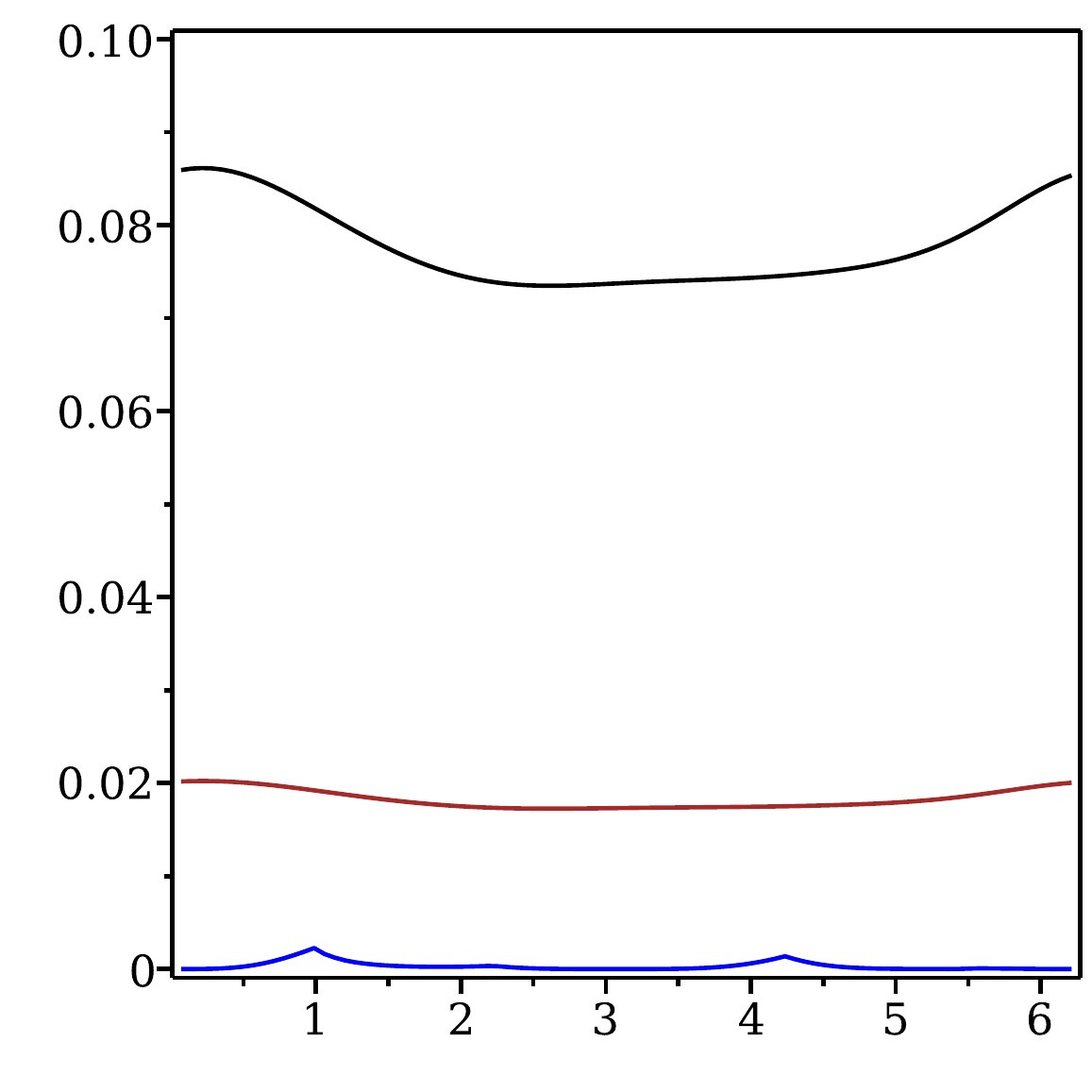}
     \begin{small}
     \put(33,-7) {(b)}
      \put(46,-7){{$G_{\blacktriangle}^{\vartriangle}(\mathbf{K,K'})$}}  
     \put(35,30) {$B_{\blacktriangle}^{\bullet}(\mathbf{K},\mathbf{K'})$}
     \put(35,85)  {$B_{\textnormal{coll}(\blacktriangle)}^{\textnormal{coll}(\vartriangle)}(\mathbf{K},\mathbf{K'})$}
     \end{small}
       \begin{scriptsize}   
   \put(0,50){\makebox(0,0){\rotatebox{90}{Distance}}}
    \put(85,0.5){$\phi$ (rad)}
   \end{scriptsize}
\end{overpic}
\quad
\begin{overpic}
    [width=45mm]{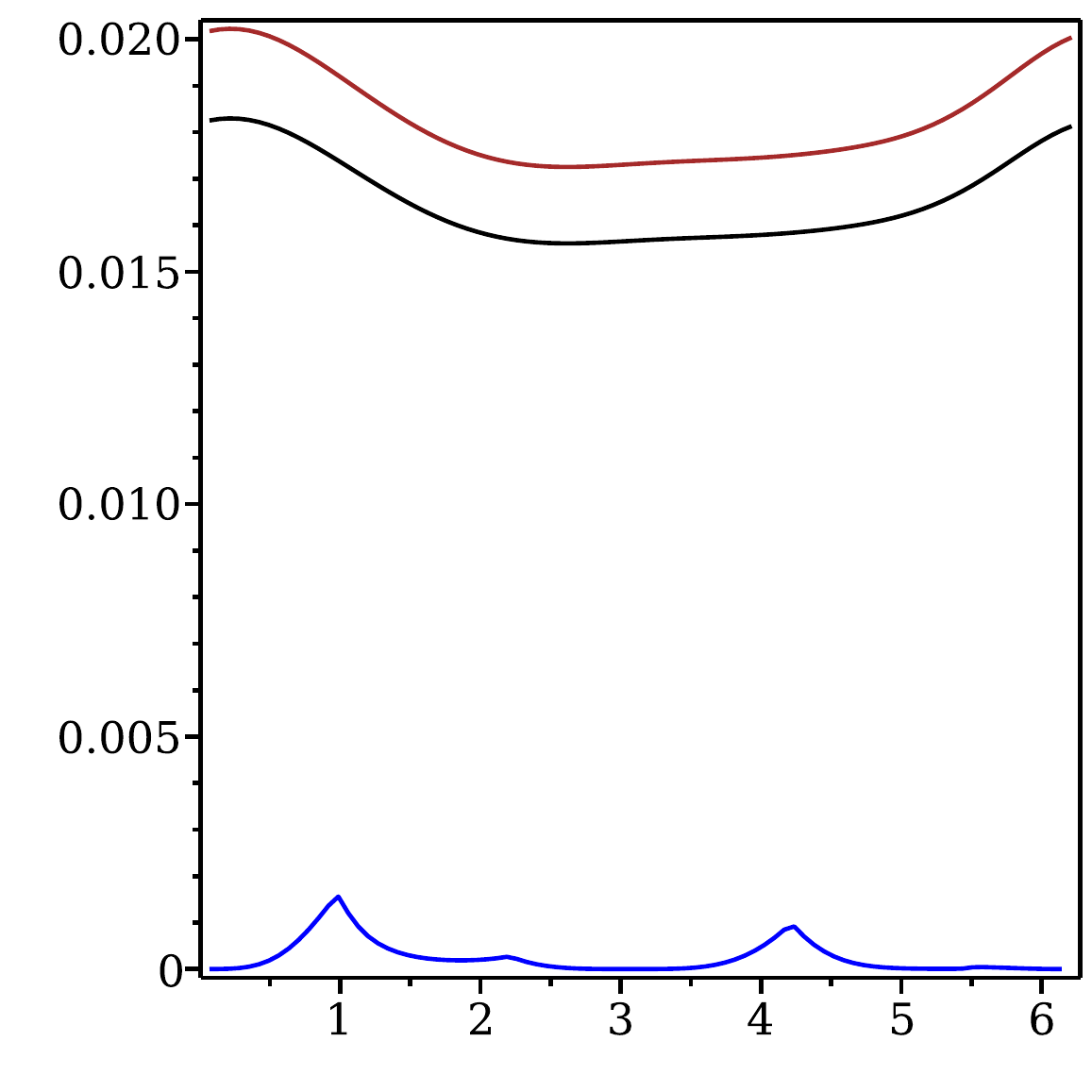}
    \begin{small}
        \put(33,-7){(c)} 
        \put(46,-7){{$G_{\vartriangle}^{\vartriangle}(\mathbf{K,K'})$}}  
        \put(36,90){$B^{\bullet }_{\vartriangle}(\mathbf{K},\mathbf{K'})$}  
    \end{small}
    \put(35,65)  {$B_{\textnormal{coll}(\vartriangle)}^{\textnormal{coll}(\vartriangle)}(\mathbf{K}, \mathbf{K'})$}
       \begin{scriptsize}   
   \put(0,50){\makebox(0,0){\rotatebox{90}{Distance}}}
    \put(85,0.5){$\phi$ (rad)}
   \end{scriptsize}
\end{overpic} 
\\ \phantom{x} \\
\begin{overpic}
    [width=45mm]{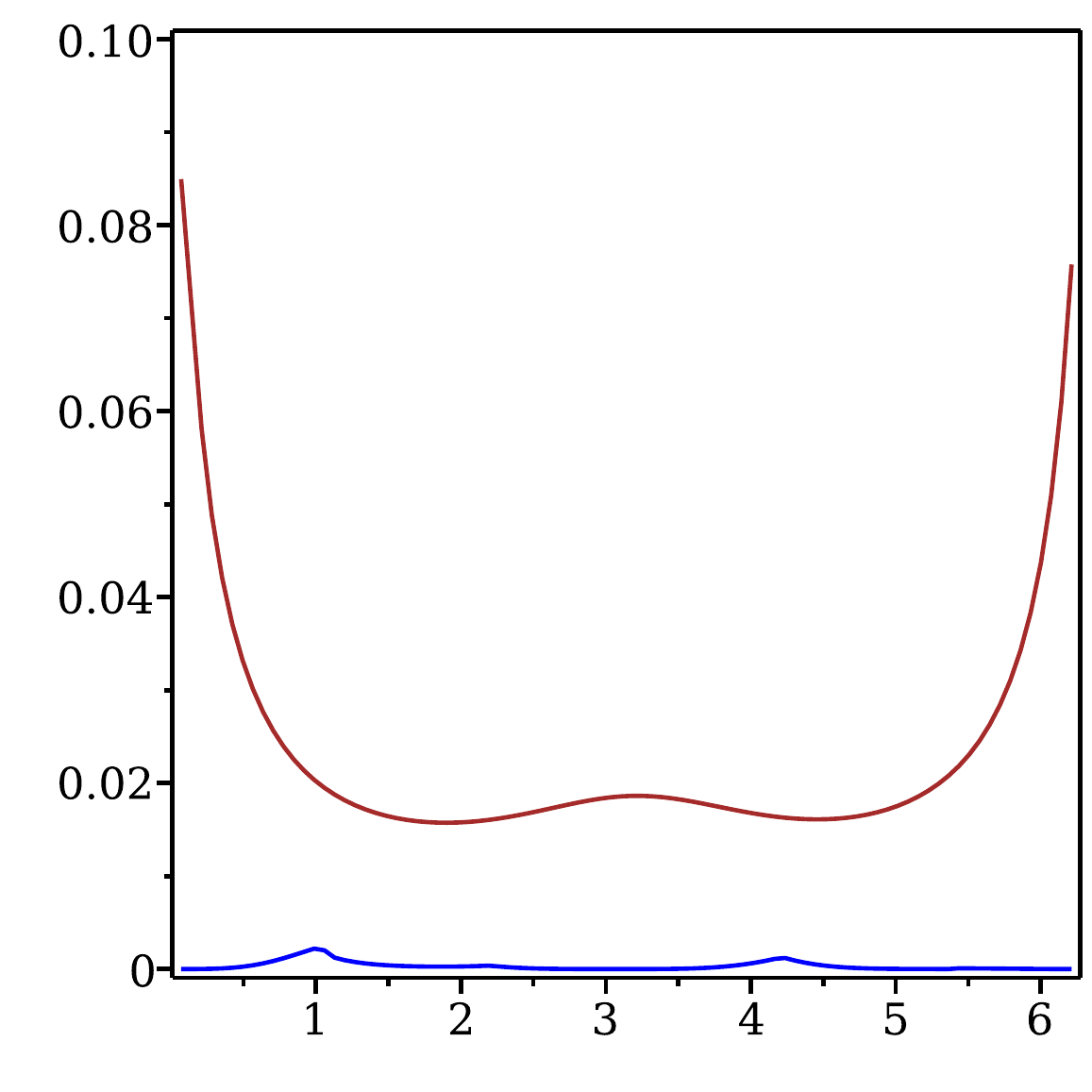}
    \begin{small}
    \put(35,-7) {(d)}
     \put(46,-7){{$G^{\rectangleblack}_{\vartriangle}(\mathbf{K,K'})$}}
     
    \put(35,32){$B_{\bullet}^{\rectangleblack}(\mathbf{K}, \mathbf{K'})$}
    \end{small}
      \begin{scriptsize}   
   \put(0,50){\makebox(0,0){\rotatebox{90}{Distance}}}
    \put(85,0.5){$\phi$ (rad)}
   \end{scriptsize}
\end{overpic}
\quad
\begin{overpic}
    [width=45mm]{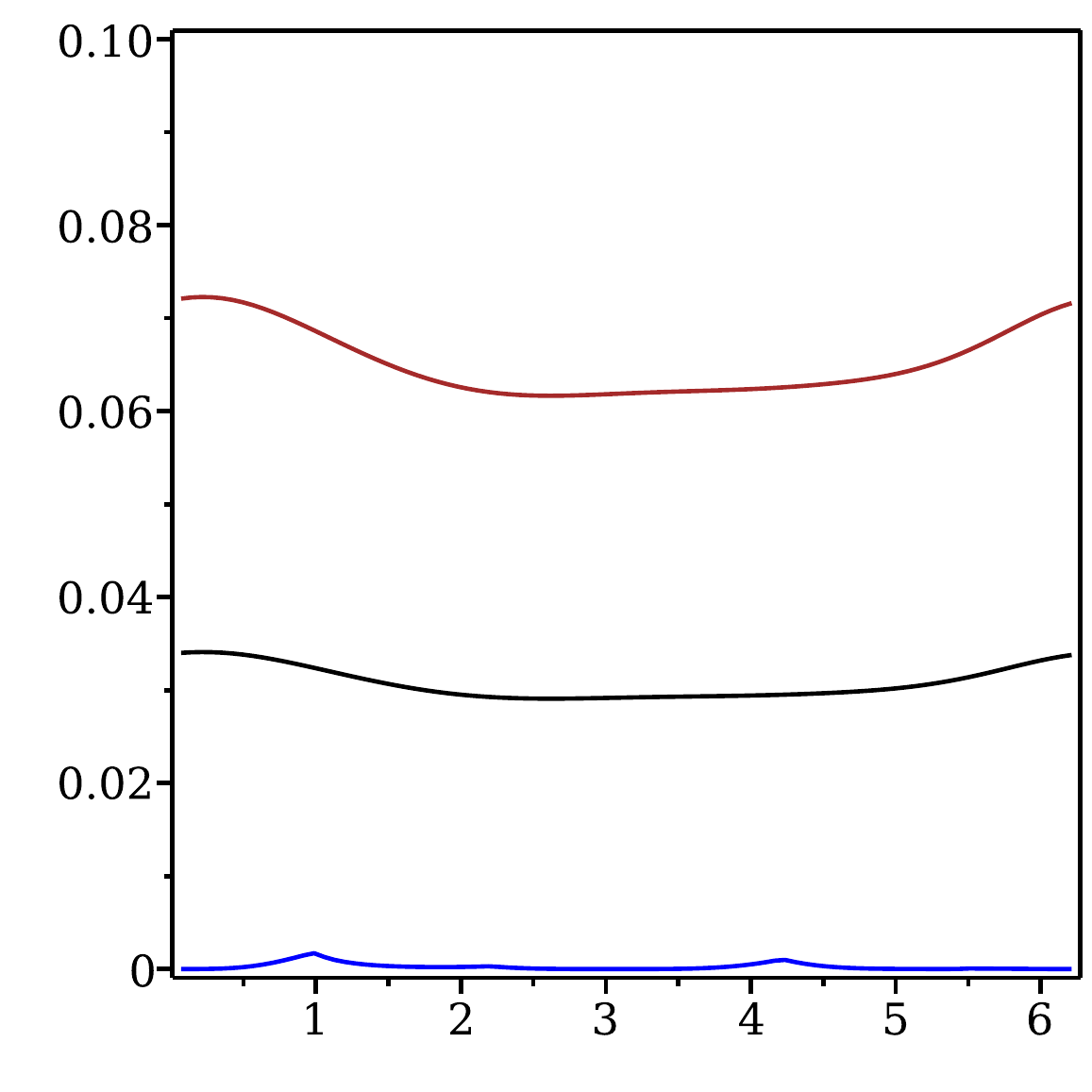}
    \begin{small}
      \put(33,-7) {(e)}
     \put(35,70){ $B^{\blacktriangle}_{\bullet}(\mathbf{K,K'})$}
      \put(46,-7){{$G^{\blacktriangle}_{\vartriangle}(\mathbf{K,K'})$}} 
     \put(35,42)  {$B^{\textnormal{coll}(\blacktriangle)}_{\textnormal{coll}(\vartriangle)}(\mathbf{K}, \mathbf{K'})$}
    \end{small}
  \begin{scriptsize}   
   \put(0,50){\makebox(0,0){\rotatebox{90}{Distance}}}
    \put(85,0.5){$\phi$ (rad)}
   \end{scriptsize}  
\end{overpic}
\quad
\begin{overpic}
   [width=45mm]{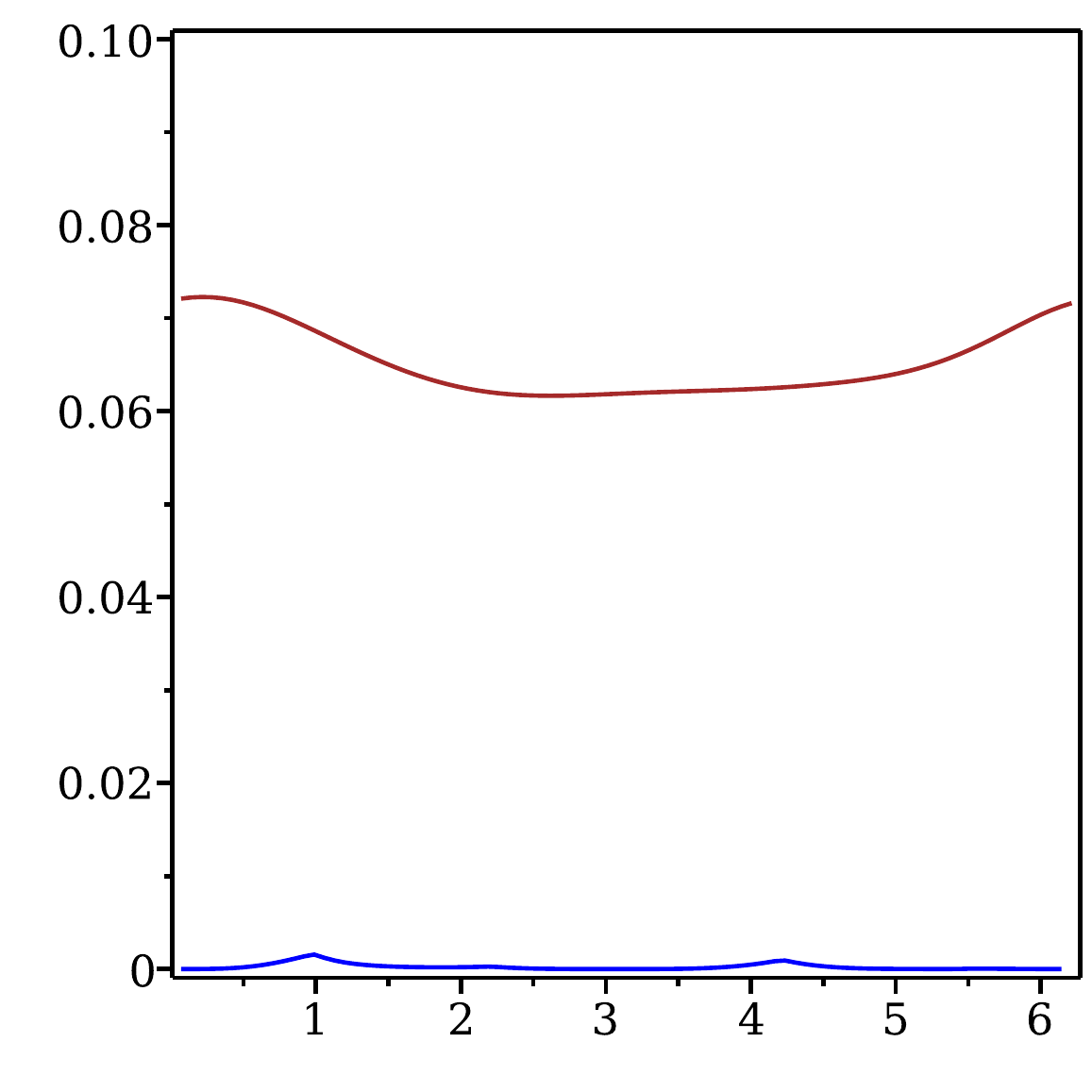}
    \begin{small}
      \put(46,68) {$B^{\vartriangle}_{\bullet}
      (\mathbf{K}, \mathbf{K'})$ }
      \put(33,-7) {(f)}
      \put(46,-7){{$G_{\vartriangle}^{\vartriangle}(\mathbf{K,K'})$}}  
    \end{small}
    \begin{scriptsize}   
   \put(0,50){\makebox(0,0){\rotatebox{90}{Distance}}}
    \put(85,0.5){$\phi$ (rad)}
   \end{scriptsize}
\end{overpic} 
\medskip
\end{center}
 \caption{Comparison of the distance to the closest regular point on $V=0$ (blue) and the lower bound for the distance to the closest singular point on the singularity/collinearity variety (black/brown).}
 \label{fig:singularity_collinearity}
  \end{figure}

\begin{figure}[t]
\begin{center}
\begin{overpic}
  [height=45mm]{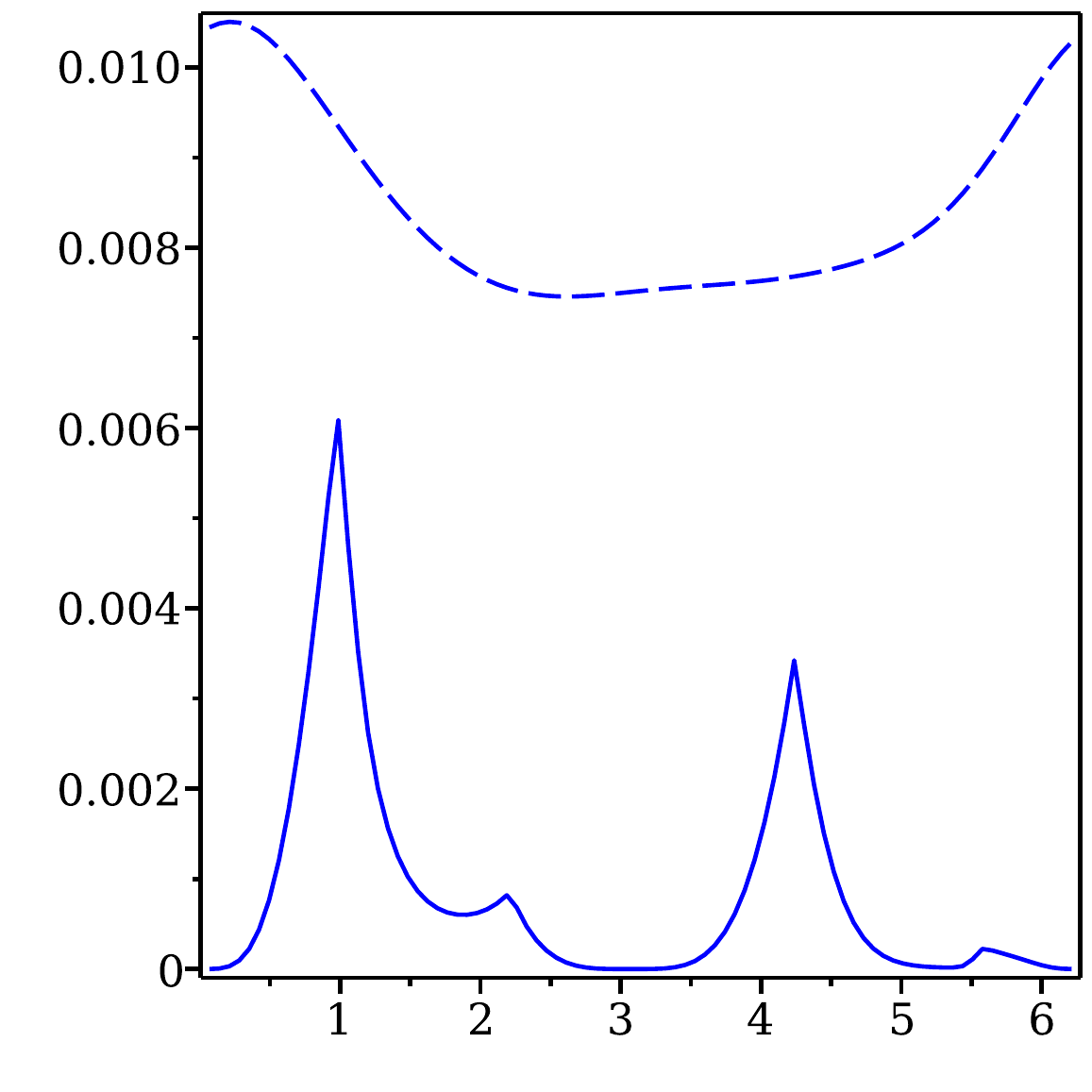}
    \begin{small}
     \put(33,-7) {(a)}  
     \put(36,83){$B_{\rectangleblack}^{\text{coll}(\vartriangle)}(\mathbf{K},\mathbf{K'})$}
     \put(46,-7){{$G_{\rectangleblack}^{\vartriangle}(\mathbf{K,K'})$}}  
      \end{small}
        \begin{scriptsize}   
   \put(0,50){\makebox(0,0){\rotatebox{90}{Distance}}}
    \put(85,0.5){$\phi$ (rad)}
   \end{scriptsize}
\end{overpic}
\quad
\begin{overpic}
    [width=45mm]{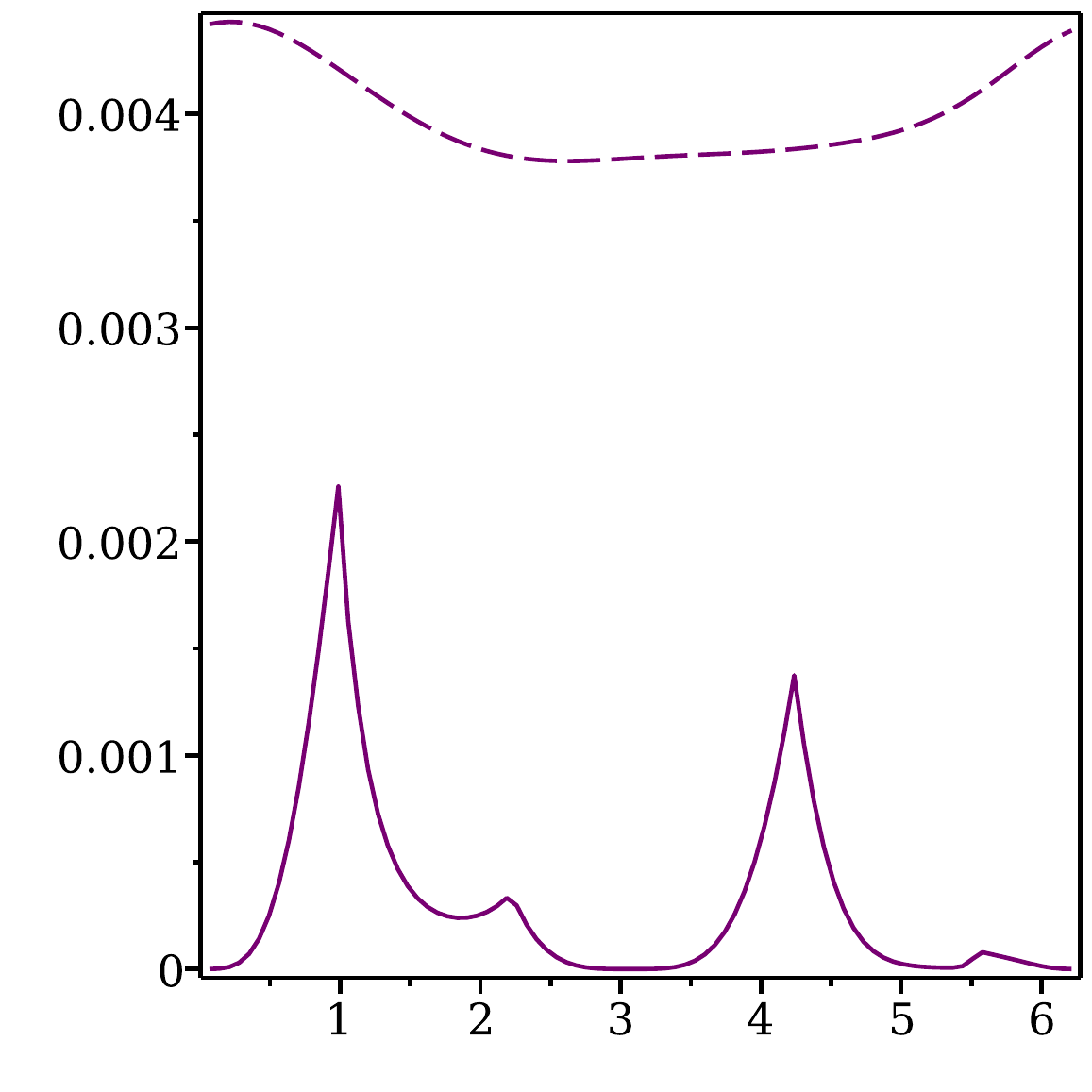}
     \begin{small}
     \put(33,-7) {(b)} 
     \put(46,-7){{$G_{\blacktriangle}^{\vartriangle}(\mathbf{K,K'})$}} 
     \put(38,90.2){$B_{\blacktriangle}^{\text{coll}(\vartriangle)}(\mathbf{K}, \mathbf{K'})$ }
     \end{small}
       \begin{scriptsize}   
   \put(0,50){\makebox(0,0){\rotatebox{90}{Distance}}}
    \put(85,0.5){$\phi$ (rad)}
   \end{scriptsize}
\end{overpic}
\quad
\begin{overpic}
    [width=45mm]{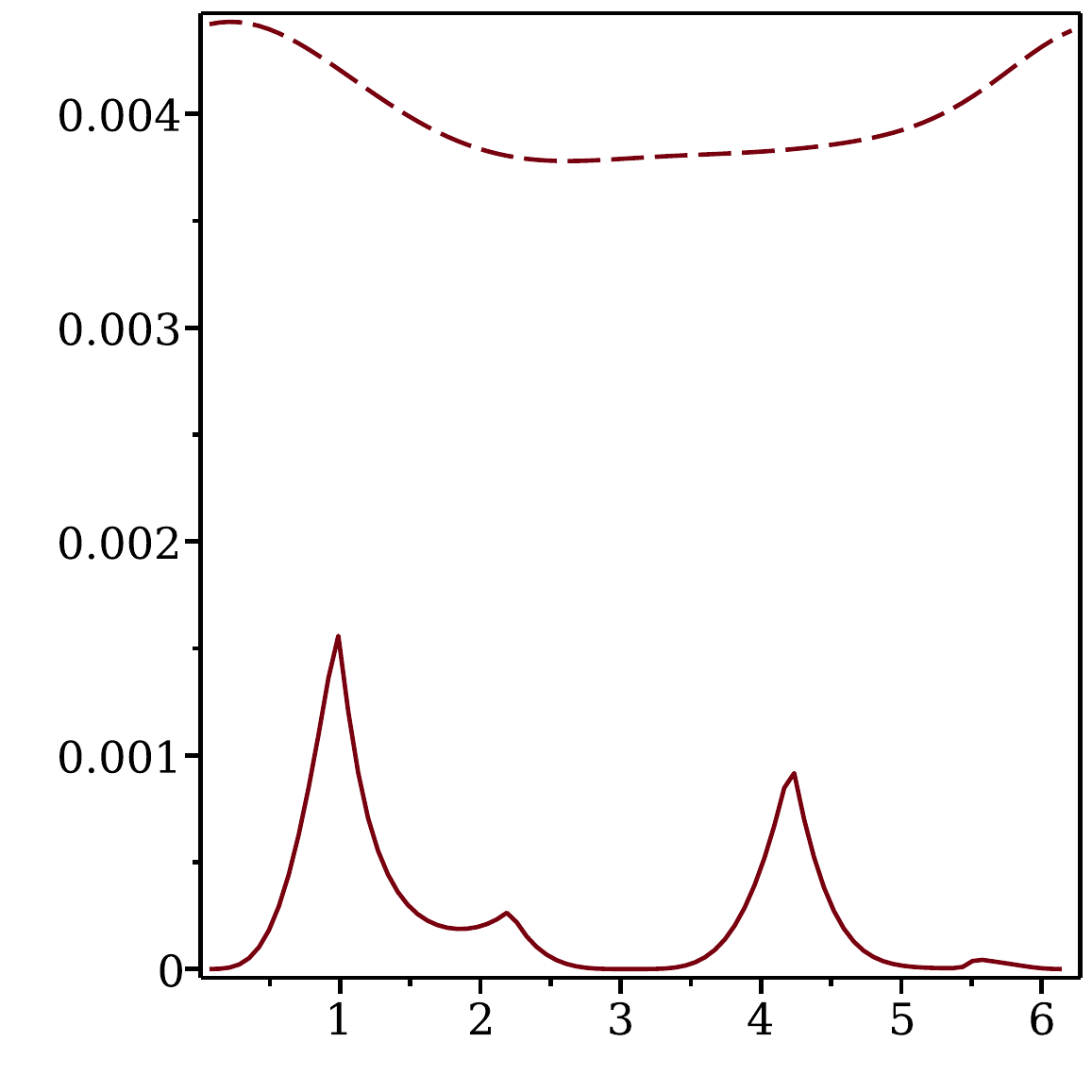}
    \begin{small}
        \put(33,-7){(c)}
        \put(46,-7){{$G_{\vartriangle}^{\vartriangle}(\mathbf{K,K'})$}}  
        \put(38,90.2) {$B^{\text{coll}(\vartriangle) }_{\vartriangle}(\mathbf{K},\mathbf{K'})$}  
    \end{small}
       \begin{scriptsize}   
   \put(0,50){\makebox(0,0){\rotatebox{90}{Distance}}}
    \put(85,0.5){$\phi$ (rad)}
   \end{scriptsize}
\end{overpic} 
\\ \phantom{x} \\
\begin{overpic}
    [width=45mm]{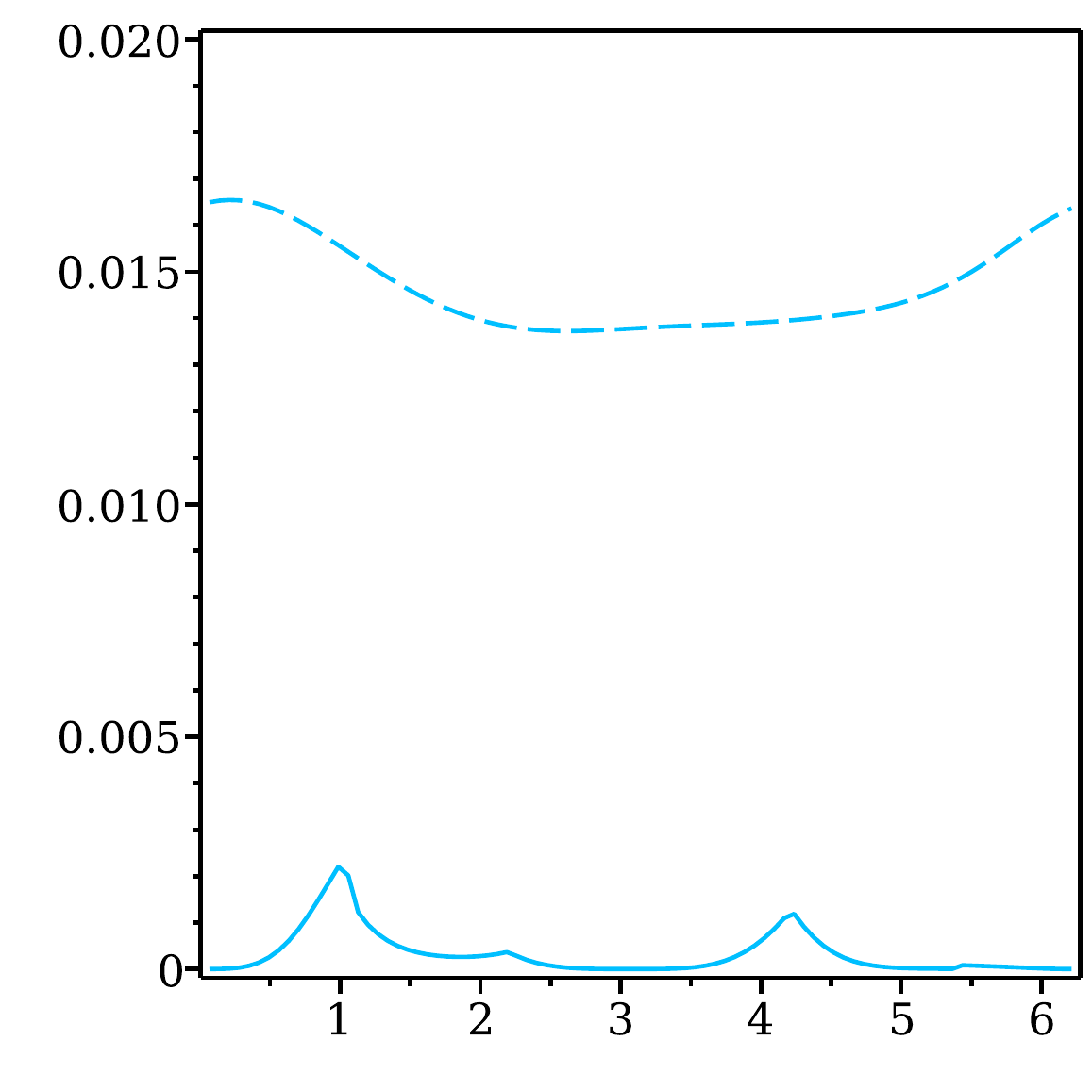}
    \begin{small}
    \put(33,-7) {(d)}
    \put(46,-7){{$G^{\rectangleblack}_{\vartriangle}(\mathbf{K,K'})$}}
    \put(36,80){$B_{\text{coll}(\vartriangle)}^{\rectangleblack}(\mathbf{K},\mathbf{K'})$}
    \end{small}
      \begin{scriptsize}   
   \put(0,50){\makebox(0,0){\rotatebox{90}{Distance}}}
    \put(85,0.5){$\phi$ (rad)}
   \end{scriptsize}
\end{overpic}
\quad
\begin{overpic}
    [width=45mm]{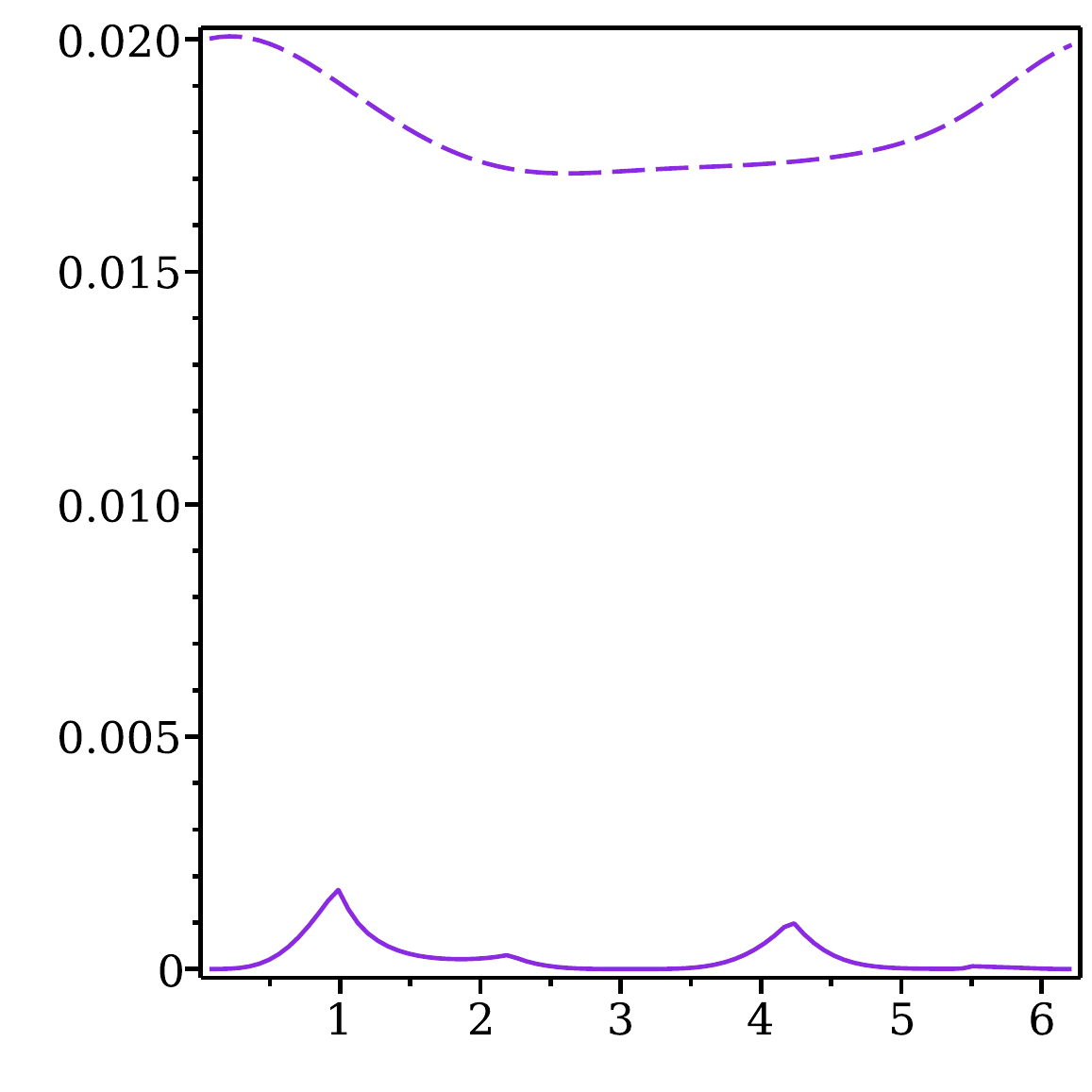}
    \begin{small}
      \put(33,-7) {(e)} 
      \put(38,90.2){$B^{\blacktriangle}_{\text{coll}(\vartriangle)}(\mathbf{K},\mathbf{K'})$}
       \put(46,-7){{$G^{\blacktriangle}_{\vartriangle}(\mathbf{K,K'})$}} 
    \end{small}
  \begin{scriptsize}   
   \put(0,50){\makebox(0,0){\rotatebox{90}{Distance}}}
    \put(85,0.5){$\phi$ (rad)}
   \end{scriptsize}  
\end{overpic}
\quad
\begin{overpic}
   [width=45mm]{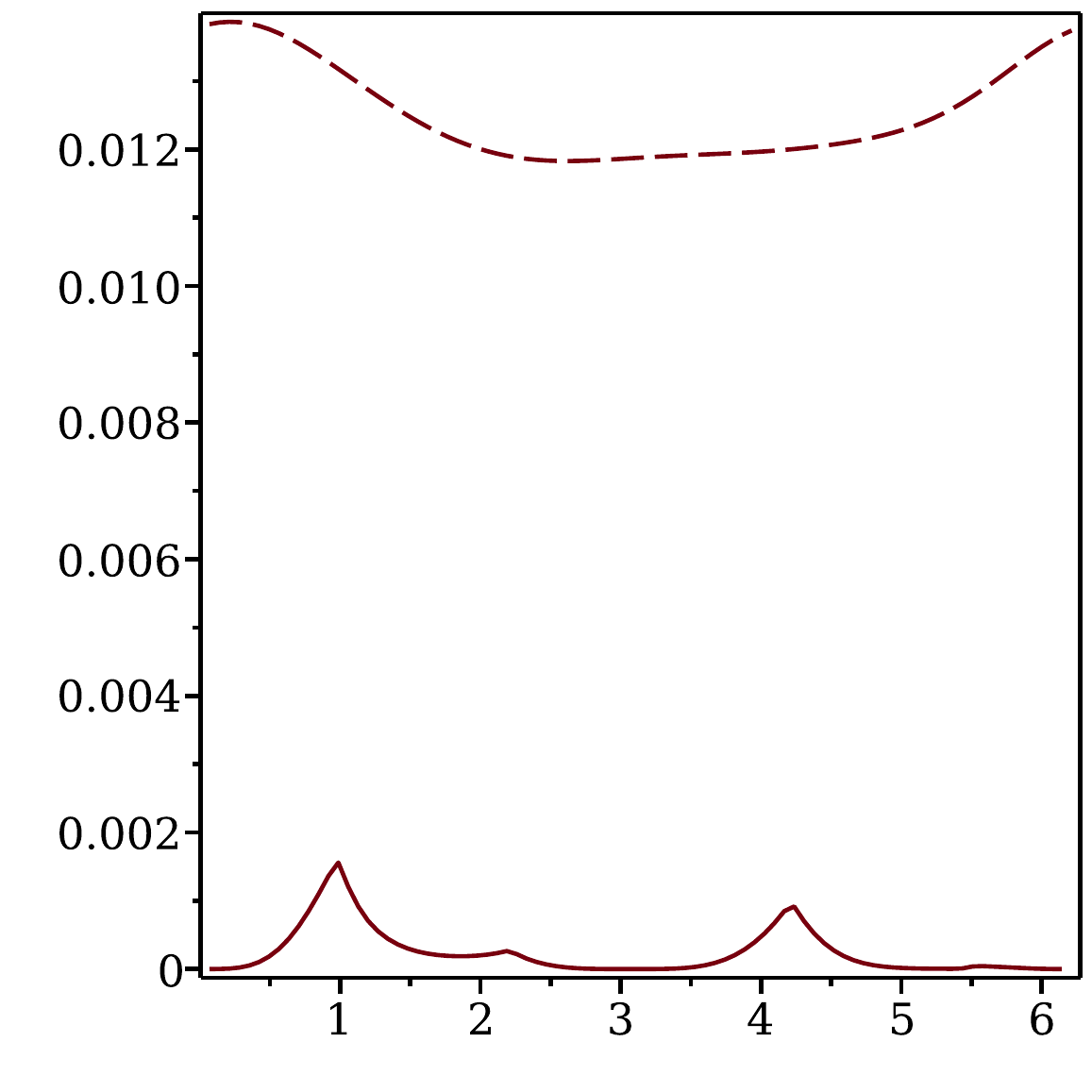}
    \begin{small}
      \put(33,-7) {(f)} 
      \put(38,90.5){$B^{\vartriangle}_{\text{coll}(\vartriangle)}(\mathbf{K},\mathbf{K'})$ }
      \put(46,-7){{$G_{\vartriangle}^{\vartriangle}(\mathbf{K,K'})$}}  
    \end{small}
    \begin{scriptsize}   
   \put(0,50){\makebox(0,0){\rotatebox{90}{Distance}}}
    \put(85,0.5){$\phi$ (rad)}
   \end{scriptsize}
\end{overpic} 
\medskip
\end{center}
 \caption{Comparison of the distances to the closest regular point on $V=0$ (solid) lower bounds for the distances to regular points of the collinearity variety (dashed).}  
 \label{fig:result3}
  \end{figure}

In Fig.\ \ref{bounds1}(c) and Fig.\ \ref{fig:singularity_collinearity}
 the lower bounds for the distance to the singular points of the singularity/collinearity variety (black/brown lines) are illustrated, which are given in Eqs.~(\ref{b7})--(\ref{cb3}).
 Fig.\ \ref{fig:result3} displays the lower bounds for the distance to regular points of the collinearity variety calculated according to Eqs.~(\ref{b1})--(\ref{b4}). 
 Notably, all these lower bounds are higher 
 compare to the singularity distances visualized in Fig.~\ref{measure2}, thus no further computations have to be performed. Moreover, by comparing Figs.\ \ref{fig:singularity_collinearity} and \ref{fig:result3} it can also be verified that the bound regarding the singular points of the collinearity variety is always higher than the bound regarding the regular points of the collinearity variety.

\subsection{Illustrative example to Section \ref{sec:erratum}}\label{sec:ex_errat}
The approach for computing the singularity distance, as described in Section \ref{sec:erratum}, is demonstrated by considering the interpretation $G_{\hrectangleblack}^{\vartriangle}(\mathbf{K})$, which is illustrated in  Fig.~\ref{zeinsingular}(c). We have a closer look at the pose $\phi=1.20015899154$ radians, which is indicated by the black line in Fig.~\ref{zeinsingular}(c). 
 At this pose, solving the embedding problem yields 12 solutions. Besides the real solution corresponding to the given pose of the manipulator, denoted as $G_{\hrectangleblack}^{\vartriangle}(\mathbf{K}_1)$ in Table~\ref{counter2}, there are two complex conjugate solutions, denoted by $G_{\hrectangleblack}^{\vartriangle}(\mathbf{K}_{2,3})$ in Table~\ref{counter2},
 which are of interest. 
If we track the paths of the three mentioned solutions  
implied by the homotopy given in Eq.\ (\ref{eq:path}), we can see that they all end up in the same real configuration denoted by   $G_{\hrectangleblack}^{\vartriangle}(\mathbf{K}')$ in Table~\ref{counter2}, which is a singular one with $k=2$.

\begin{table}
\caption{Numerical data of the example discussed in Section \ref{sec:ex_errat}.}
\centering
\begin{tabular}{|l|l|l|}
\hline
$G_{\hrectangleblack}^{\vartriangle}(\mathbf{K}_{1})$ & $G_{\hrectangleblack}^{\vartriangle}(\mathbf{K}_{2,3})$ &  $G_{\hrectangleblack}^{\vartriangle}(\mathbf{K}')$ \\ \hline
2.703709942 & -2.571965406$\pm$1.642673261I & 0.377950101 \\ \hline
0.956685654 & 2.626174234$\pm$1.608765617I & 0.262619654 \\ \hline
3.790338633 & 1.842974136$\pm$1.496551157I & 0.315726176 \\ \hline
3.752975711 & 2.825168520$\pm$4.850669104I  & 0.373189870 \\ \hline
1.201726134 & -1.23298175$\pm$0.567303095I & 0.285678598  \\ \hline
2.613201467 & 5.635798691$\pm$2.591985374I & 0.469363370 \\ \hline
\end{tabular}
\label{counter2} 
\end{table}

\begin{table}[ht!]
\caption{Numerical data of the example discussed in Section \ref{zeincomparison}.}
\centering
\begin{tabular}{|l|l|l|l|l|l|}
\hline
$G_{\rectangleblack}^{\rectangleblack}(\mathbf{K}_{1})$ & $G_{\rectangleblack}^{\rectangleblack}(\mathbf{K}_{2})$ & $G_{\rectangleblack}^{\rectangleblack}(\mathbf{K}_{1}')$ \cite{zein} & $G_{\rectangleblack}^{\rectangleblack}(\mathbf{K}_{2}')$ \cite{zein} &  $G_{\rectangleblack}^{\rectangleblack}(\mathbf{K}')$ Eq.~(\ref{eq:lagrangeplus}) & $G_{\rectangleblack}^{\rectangleblack}(\mathbf{K}')$ \cite{NAWRATIL2022104510}  \\ \hline
0.96973480 & -26.68181043 & -21.46607391  & -17.55608464 & -19.87544138  & -17.31188631 \\ \hline
29.98432283 & -13.71426231 & 24.41132358 & 27.358525950  & 18.59890609 & 15.03840255 \\ \hline
-1.61948126 & -23.71472529 & -36.94992514 & -34.26226994 & -36.82134659 & -34.34734357 \\ \hline
46.82645928 & -30.49395258 & 17.29711613 & 24.00219270 & 16.81063977 & 14.64496133 \\ \hline
-16.95132098 & -8.52622187 & -26.77327531 & -27.36262140  & -31.34945104 & -30.17306953 \\ \hline
40.62111555 & -23.9455967 & 4.25842767 & 8.97001222 & 1.20198967 & -1.35963289 \\ \hline
\end{tabular}
\label{zeinsingular}
\end{table}

\begin{figure}[ht!]
  \centering
 \begin{overpic}
     [width=50mm]{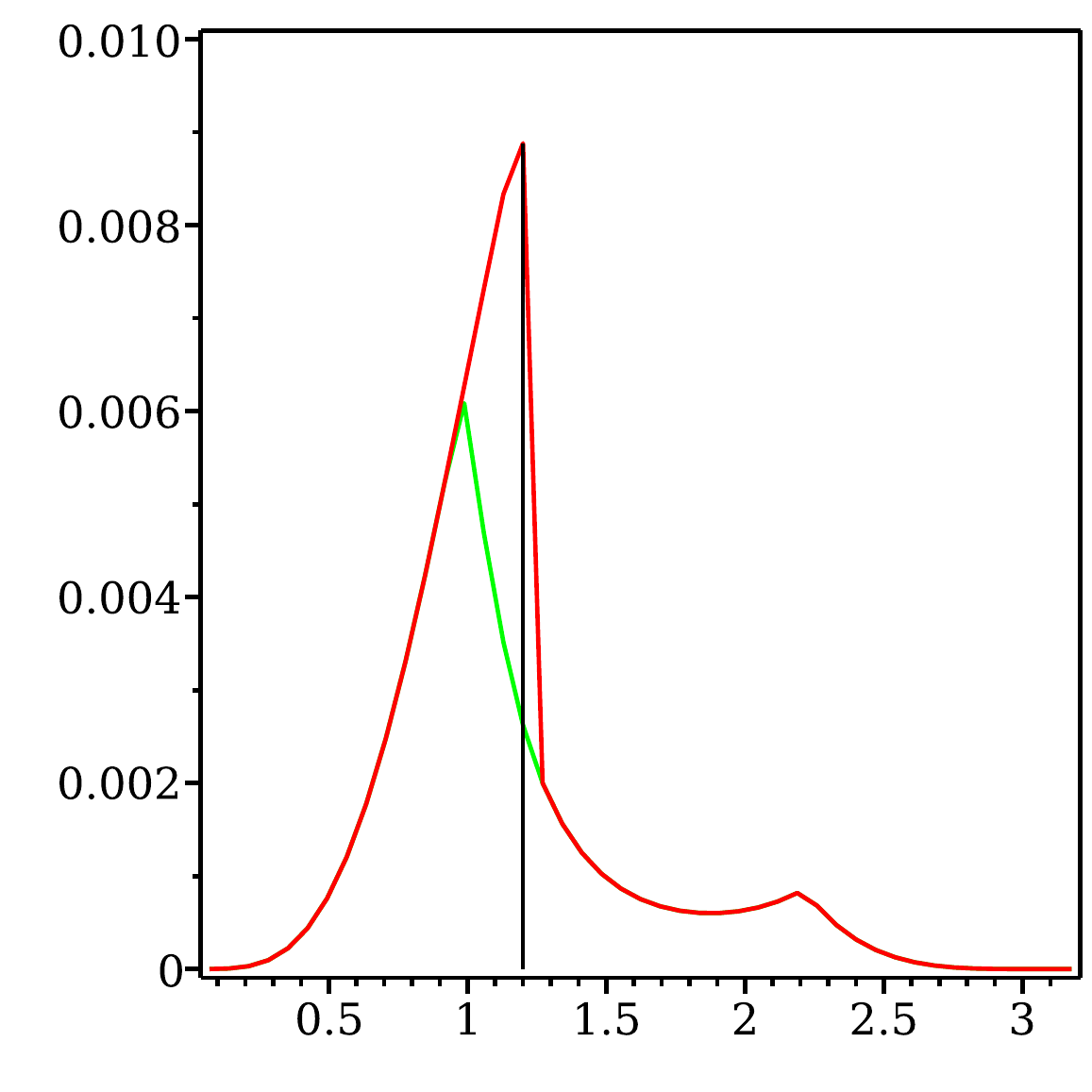}
     \begin{small}
     \put(45,-7) {(a)} 
    \end{small}
     \begin{scriptsize} 
     \put(82,1){$\phi$ (rad)}
      \put(0,50){\makebox(0,0){\rotatebox{90}{Distance}}}
      \end{scriptsize}
\end{overpic}
\quad
 \begin{overpic}
      [width=50mm]{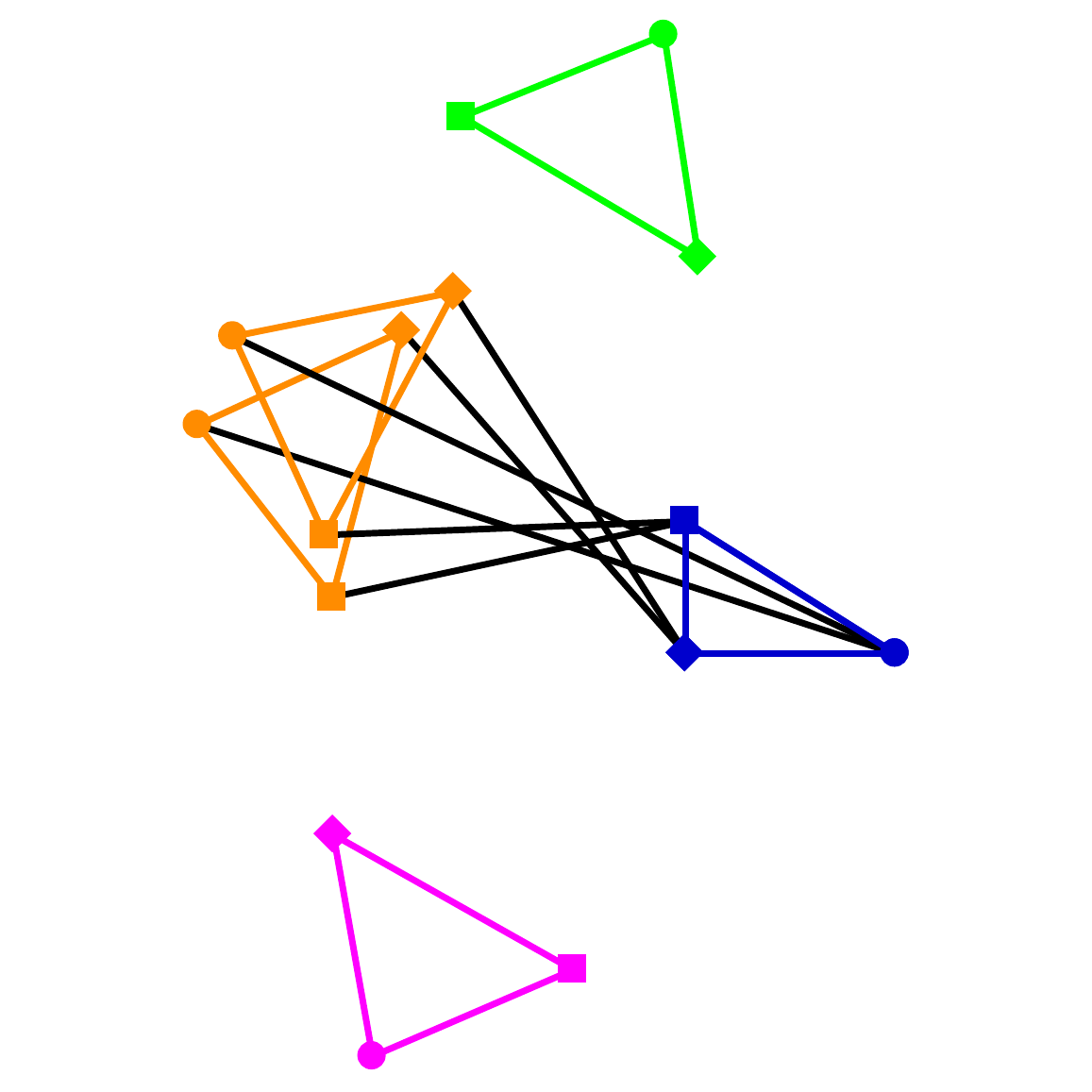}
     \begin{small}
     \put(45,-7) {(b)}
     \put (55,35){$\mathbf{k}_{1}$}
      \put (82,33){$\mathbf{k}_{2}$}
        \put (57,57){$\mathbf{k}_{3}$}
         \put (58,100){$\mathbf{k}_{5}$}
          \put (58,72){$\mathbf{k}_{4}$}
          \put (34,85){$\mathbf{k}_{6}$}
            \put (64,86) {$G_{\rectangleblack}^{\rectangleblack}(\mathbf{K}_{1})$}
            \put (45,20) {$G_{\rectangleblack}^{\rectangleblack}(\mathbf{K}_{2})$}
             \put (16,75.5) {$G_{\rectangleblack}^{\rectangleblack}(\mathbf{K}_{1}')$}
             \put (2,48) {$G_{\rectangleblack}^{\rectangleblack}(\mathbf{K}_{2}')$} 
          \end{small}
\end{overpic}
\begin{overpic}
      [width=50mm]{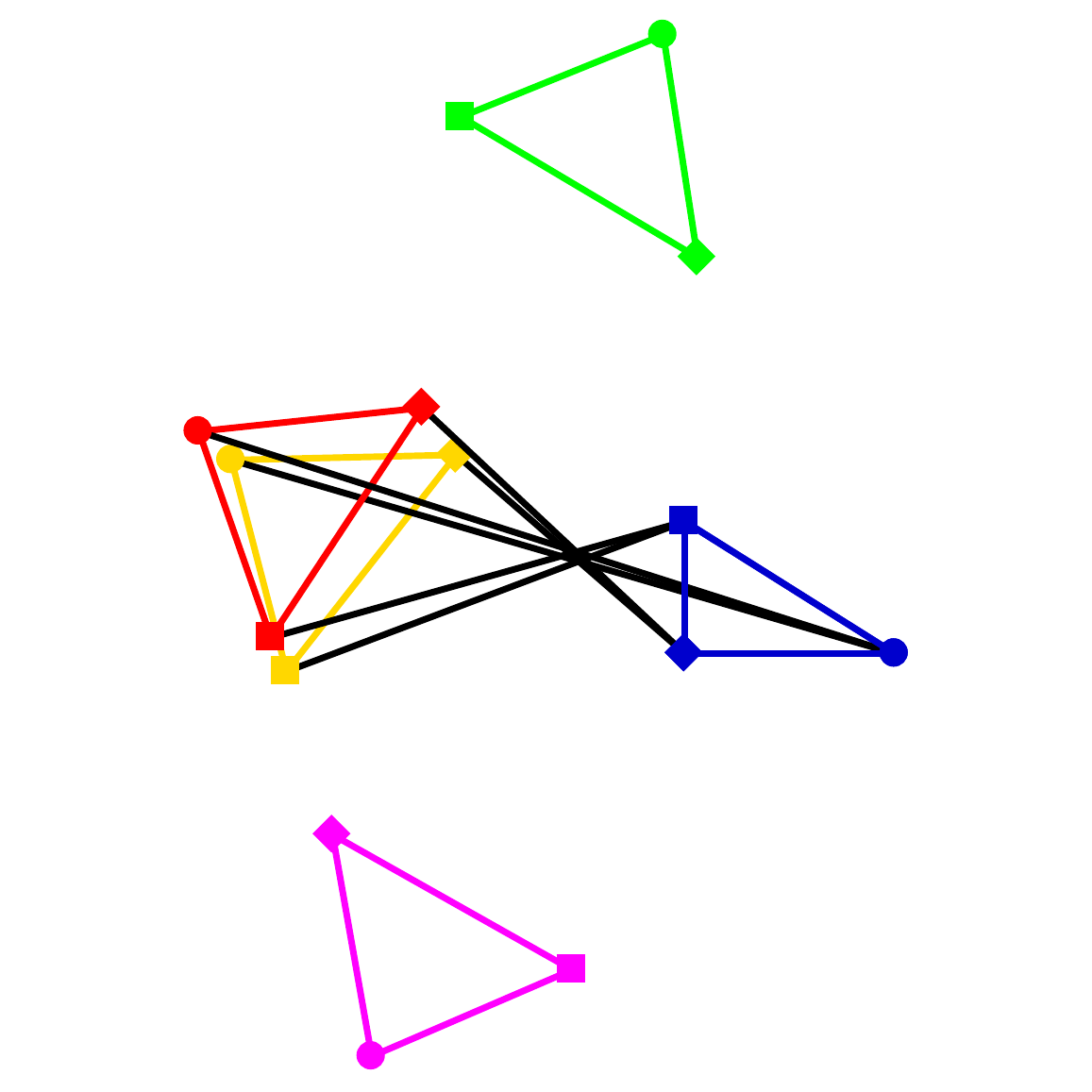}
     \begin{small}
     \put(45,-7) {(c)}
      \put (55,35){$\mathbf{k}_{1}$}
      \put (82,33){$\mathbf{k}_{2}$}
        \put (57,57){$\mathbf{k}_{3}$}
         \put (58,100){$\mathbf{k}_{5}$}
          \put (58,72){$\mathbf{k}_{4}$}
          \put (34,85){$\mathbf{k}_{6}$}
            \put (64,86) {$G_{\rectangleblack}^{\rectangleblack}
            (\mathbf{K}_{1})$}
            \put (45,20) {$G_{\rectangleblack}^{\rectangleblack}(\mathbf{K}_{2})$}
            \put (16,67) {$G_{\rectangleblack}^{\rectangleblack}(\mathbf{K}')$}
             \put (0,40) {$G_{\rectangleblack}^{\rectangleblack}(\mathbf{K}')$}
      \end{small}
\end{overpic}
\medskip
\caption{(a) Illustrative example to Section \ref{sec:erratum} discussed in Section \ref{sec:ex_errat}: The red graph indicates without fail only the distance to singularities with odd $k$ and the green takes all singularities (also with even $k$) into account. Illustration of the two real solutions of the direct kinematics
$G_{\rectangleblack}^{\rectangleblack}(\mathbf{K}_{1})$ (green) and $G_{\rectangleblack}^{\rectangleblack}(\mathbf{K}_{2})$ (magenta) together with (b) the approximate singular configurations $G_{\rectangleblack}^{\rectangleblack}(\mathbf{K}'_{1})$ and 
$G_{\rectangleblack}^{\rectangleblack}(\mathbf{K}'_{2})$ in orange, (c) the closest singularity $G_{\rectangleblack}^{\rectangleblack}(\mathbf{K}')$
with respect to the intrinsic metric given in Eq.\ (\ref{eq:lagrangeplus}) (red) and \cite[Eq.\ (44)]{NAWRATIL2022104510}] (gold), respectively.
In order not to overload the figures we abstained from drawing the legs for the direct kinematics solutions but did it only for the (approximate) singular configurations to show that the legs belong (approximately) to a pencil of lines. Note that the corresponding anchor points in the platform and base are flagged by diamonds, circles, and boxes, respectively.}
\label{comparsion}
\end{figure}

\subsection{Comparison of $D_{\rectangleblack}^{\rectangleblack}(\mathbf{K},\mathbf{K}')$ with already existing methods}\label{zeincomparison}
 
In this section, we compare the closest singular configuration with respect to $D_{\rectangleblack}^{\rectangleblack}(\mathbf{K},\mathbf{K}')$ with the methods given in \cite{zein} and 
\cite[Eq.\ (44)]{NAWRATIL2022104510}, respectively (cf.\ Section \ref{sec:review}). 
To do so, we use~\cite[Example 2]{zein} with the following numerical data provided by \cite{personal1}:
The leg lengths in the given non-singular configuration read as
$\ell(\mathbf{k}_{1},\mathbf{k}_{4})=\tfrac{38\,125}{1000}$, 
$\ell(\mathbf{k}_{2},\mathbf{k}_{5})=50$,  
$\ell(\mathbf{k}_{3},\mathbf{k}_{6})=33$. The leg lengths of the 
closest singularity according to the metric of \cite{zein} are given by $\ell(\mathbf{k}'_{1},\mathbf{k}'_{4})=\tfrac{32\,507}{1000}$, 
$\ell(\mathbf{k}'_{2},\mathbf{k}'_{5})=\tfrac{27\,809}{500}$,  
$\ell(\mathbf{k}'_{3},\mathbf{k}'_{6})=\tfrac{13\,691}{500}$. 
The corresponding singularity distance equals $\tfrac{e}{2}=\frac{5618}{1000}$.

First, we recompute the given non-singular configuration by solving the direct kinematics problem, which has only two real solutions $G_{\rectangleblack}^{\rectangleblack}(\mathbf{K}_1)$ and $G_{\rectangleblack}^{\rectangleblack}(\mathbf{K}_2)$ (cf.\ Table~\ref{zeinsingular} and Fig.\ \ref{comparsion}(a)). 
If we do the same procedure for determining the closest-singular configuration we expect to get a real double solution, but we get again two ordinary real solutions $G_{\rectangleblack}^{\rectangleblack}(\mathbf{K}'_1)$ and 
$G_{\rectangleblack}^{\rectangleblack}(\mathbf{K}'_2)$, which are close to each other (cf.\ Table~\ref{zeinsingular} and Fig.\ \ref{comparsion}(a)).  The reason for this is numerical inaccuracies in the input data of the leg lengths arising from the slicing approach used~\cite{zein}.

For the given non-singular configurations
$G_{\rectangleblack}^{\rectangleblack}(\mathbf{K}_1)$ and $G_{\rectangleblack}^{\rectangleblack}(\mathbf{K}_2)$
We also computed the closest singular configuration using the metrics given in Eq.~(\ref{eq:lagrangeplus}) and \cite[Eq.\ (44)]{NAWRATIL2022104510}, respectively, by following the procedures given in Sections \ref{non} and \ref{postprocess}.  Note that for both metrics the obtained closest singularity $G_{\rectangleblack}^{\rectangleblack}(\mathbf{K}')$ (cf.\  Table~\ref{zeinsingular} and Fig.\ \ref{comparsion}(b)) is the same for both input configurations 
$G_{\rectangleblack}^{\rectangleblack}(\mathbf{K}_1)$ and $G_{\rectangleblack}^{\rectangleblack}(\mathbf{K}_2)$. 
For reasons of completeness, we also give the corresponding singular distances which are $0.0082259948$ and $490\,387.2090$, respectively.

\subsection{Comparison with extrinsic metric distance computations}\label{eg:comparison}

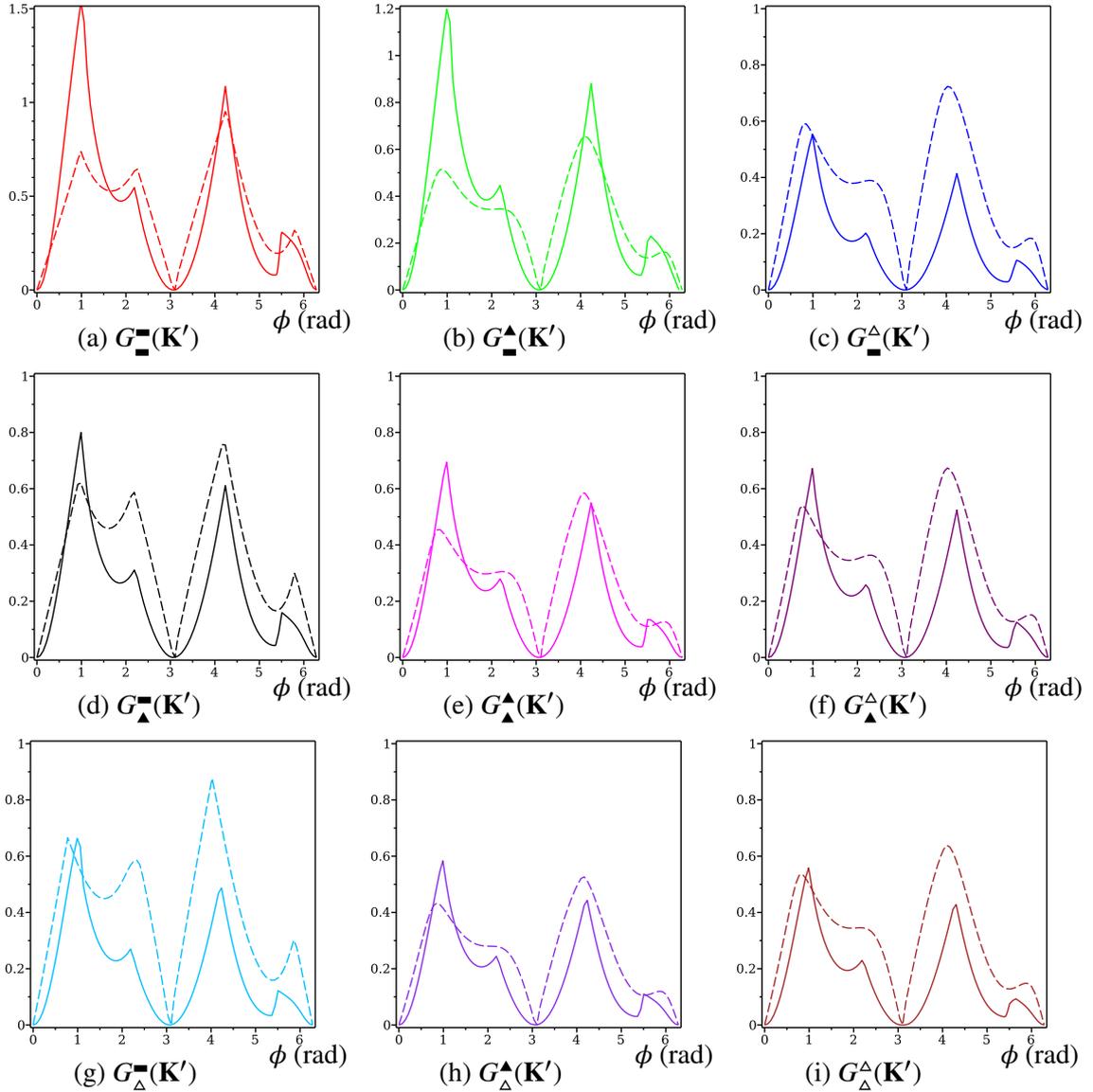
\begin{figure}[ht!]
\begin{center}
\begin{overpic}
  [height=46mm]{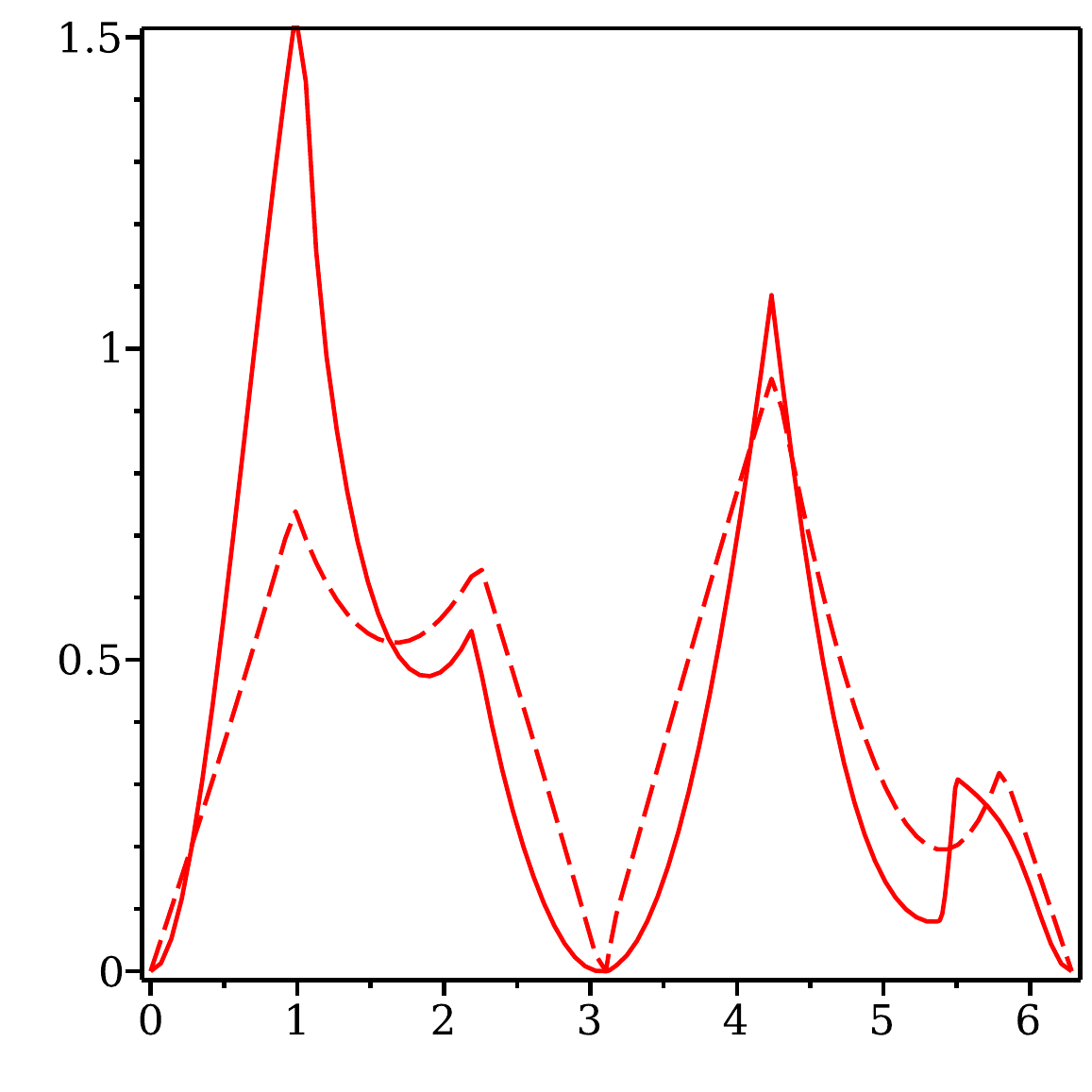}
  \begin{small}
   \put(85,-0.75){$\phi$ (rad)}
    \put (26,-6){(a) $G_{\rectangleblack}^{\rectangleblack}(\mathbf{K'})$}
    \end{small}
\end{overpic}
\quad
\begin{overpic}
  [height=46mm]{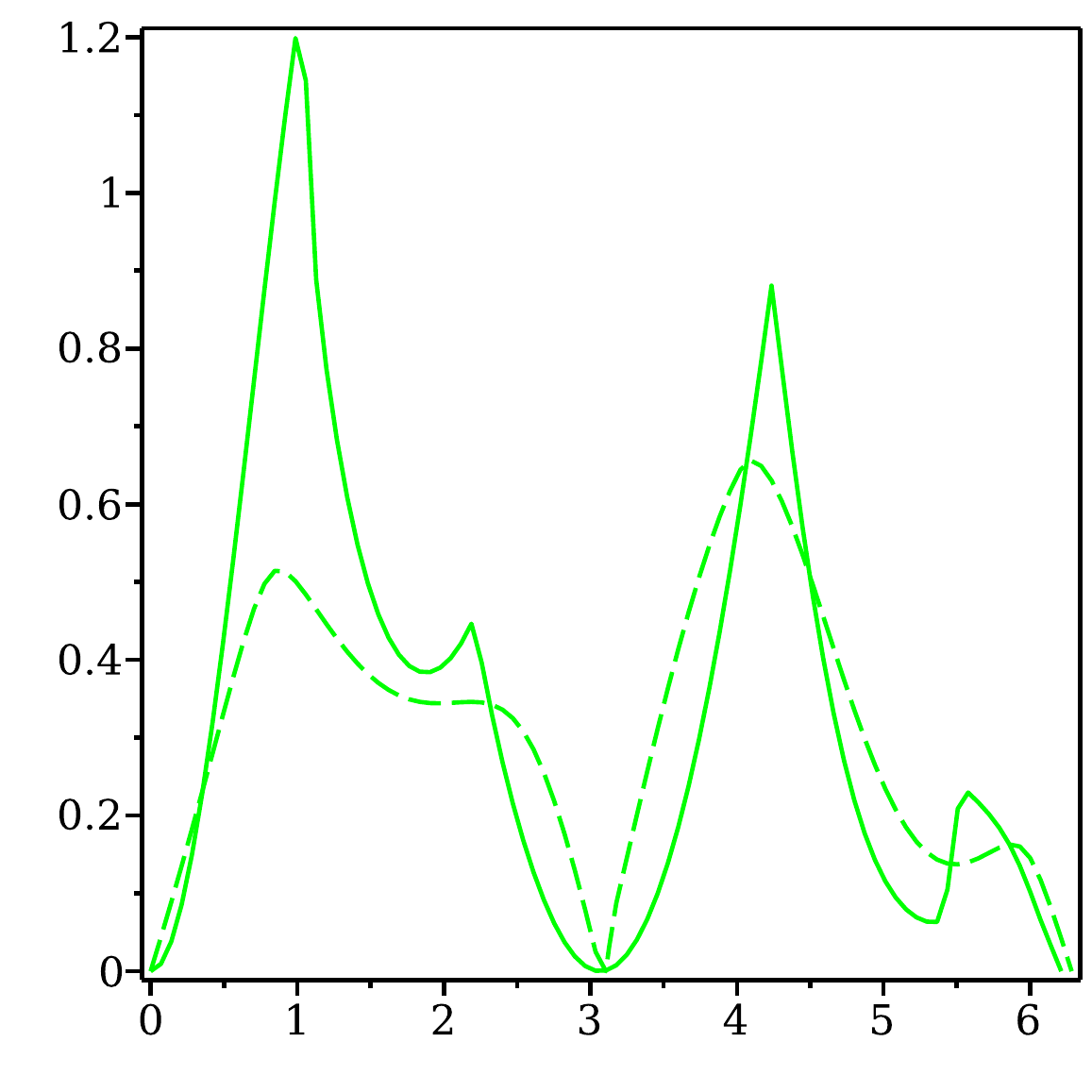}
  \begin{small}
  \put(85,-0.75){$\phi$ (rad)}
    \put (26,-6){(b) $G_{\rectangleblack}^{\blacktriangle}(\mathbf{K'})$} 
    \end{small}
\end{overpic}
\quad
\begin{overpic}
  [height=46mm]{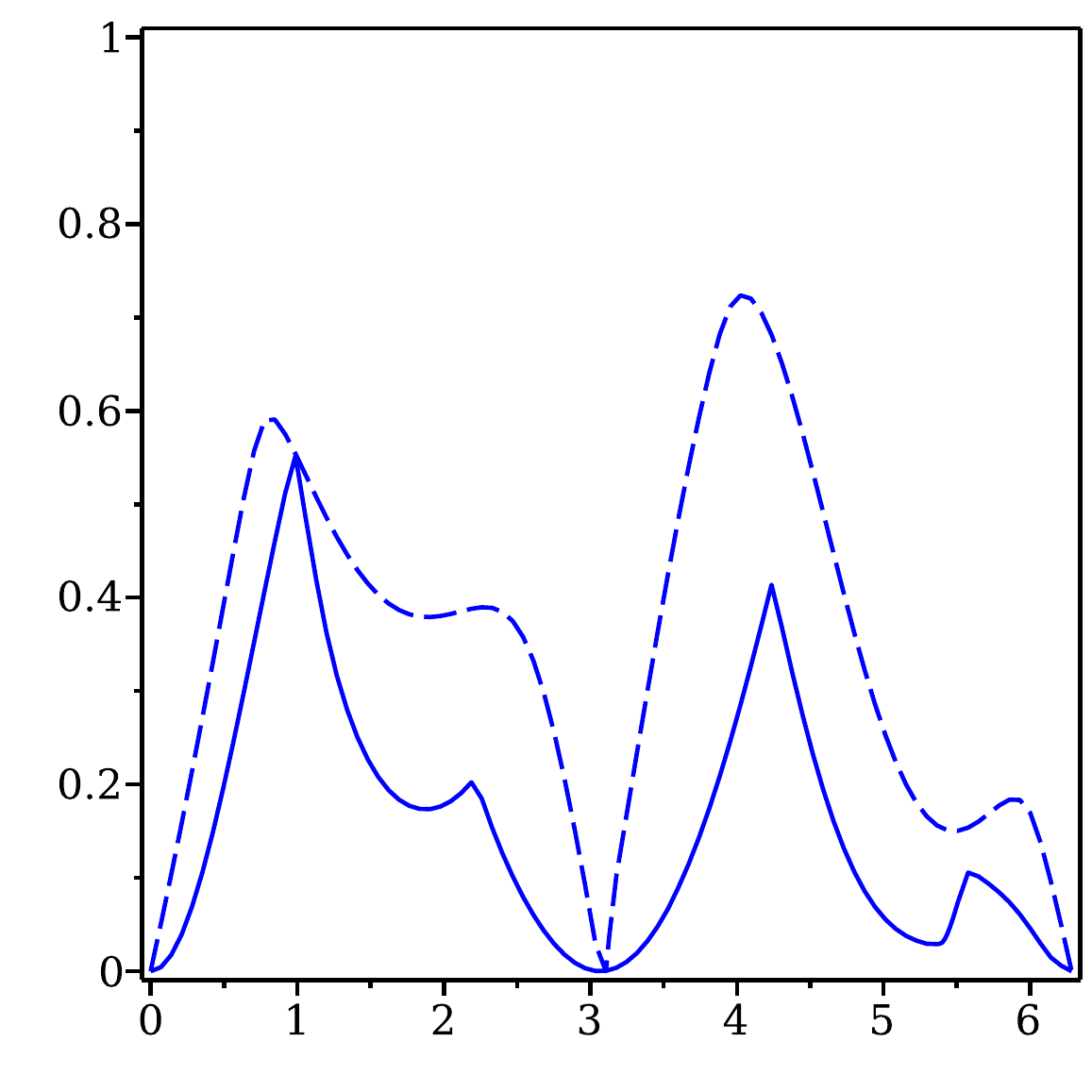}
  \begin{small}
  \put(85,-0.75){$\phi$ (rad)}
    \put (26,-6){(c) $G_{\rectangleblack}^{\vartriangle}(\mathbf{K'})$}
    \end{small}
\end{overpic}
\\ \phantom{x} \\
\begin{overpic}
  [height=46mm]{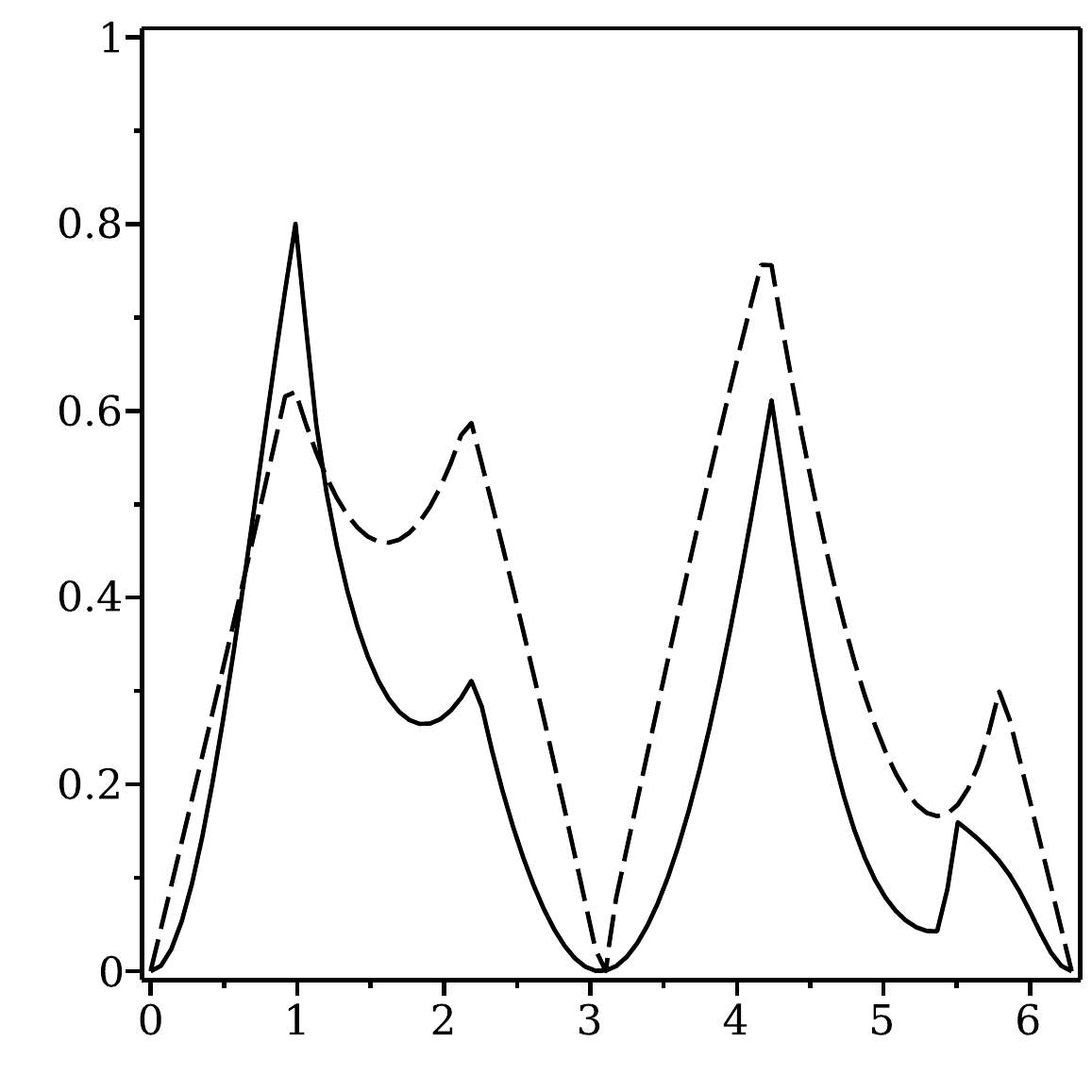}
  \begin{small}
   \put(85,-0.75){$\phi$ (rad)}
      \put (26,-6){(d) $G^{\rectangleblack}_{\blacktriangle}(\mathbf{K'})$}
     \end{small}
     \end{overpic}
     \quad
     \begin{overpic}
  [height=46mm]{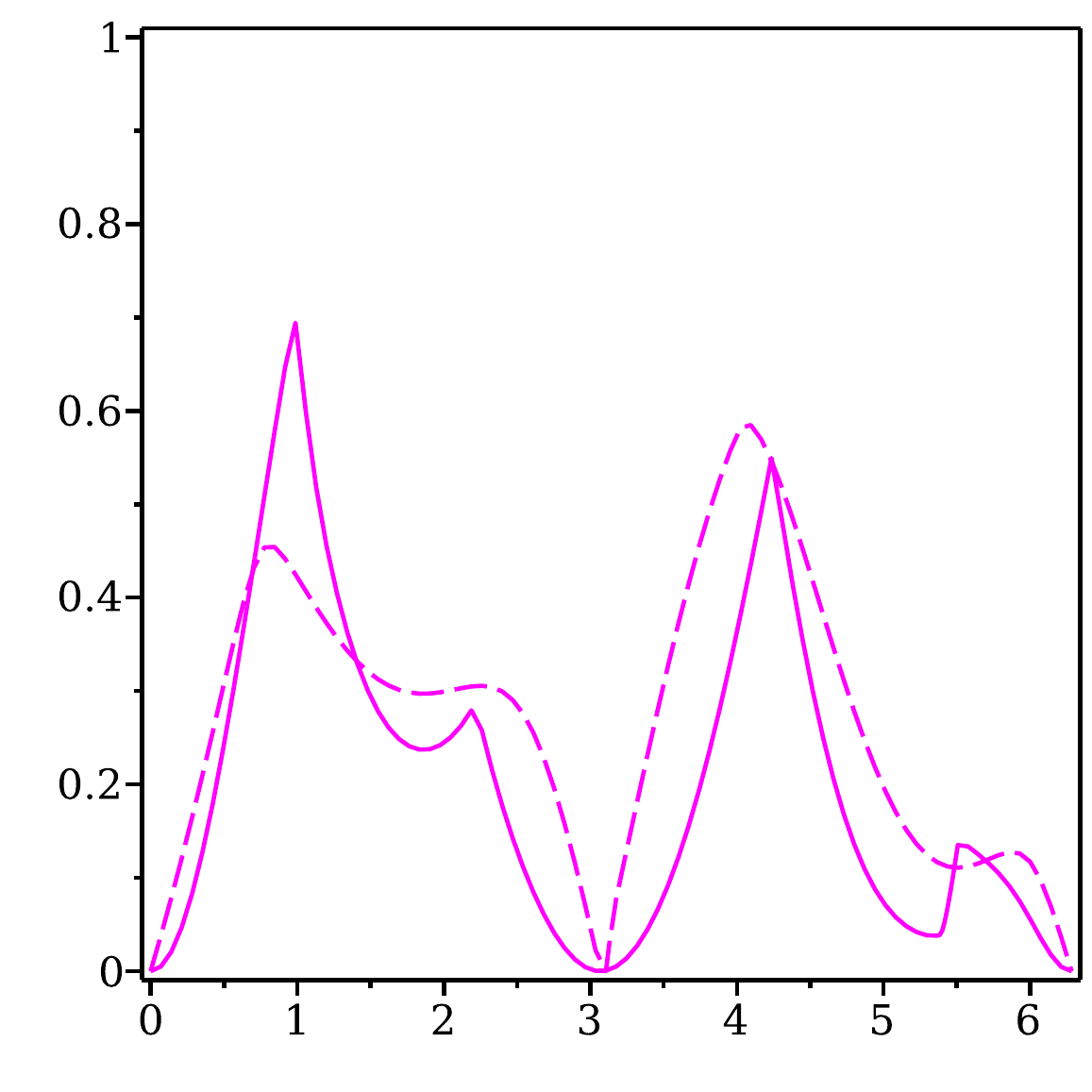}
  \begin{small}
  \put(85,-0.75){$\phi$ (rad)}
      \put (26,-6){(e) $G^{\blacktriangle}_{\blacktriangle}(\mathbf{K'})$}
     \end{small}
     \end{overpic}
\quad     
\begin{overpic}
  [height=46mm]{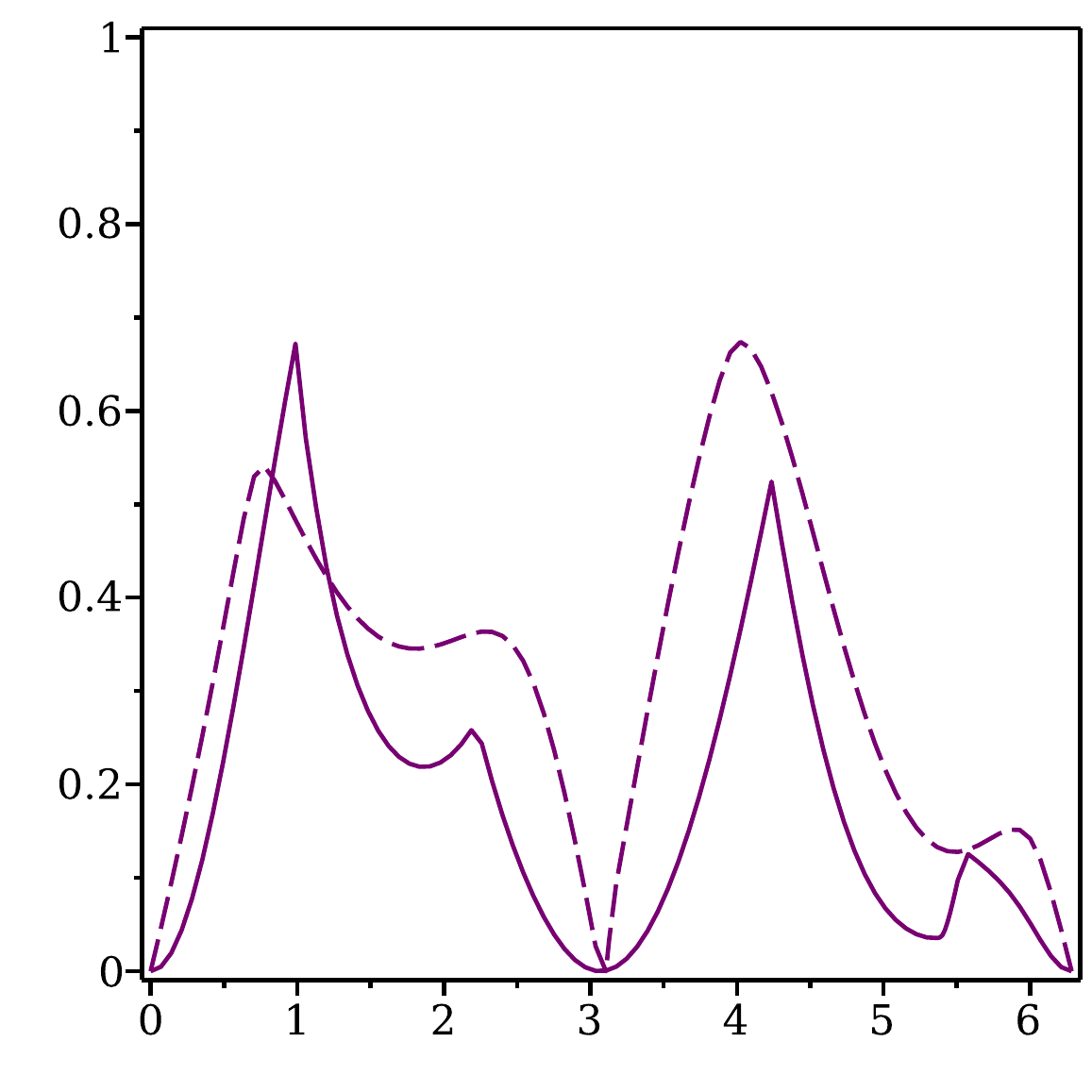}
  \begin{small}
  \put(85,-0.75){$\phi$ (rad)}
      \put (26,-6){(f) $G^{\vartriangle}_{\blacktriangle}(\mathbf{K'})$}
     \end{small}
     \end{overpic}
\\ \phantom{x} \\
 \begin{overpic}
  [height=46mm]{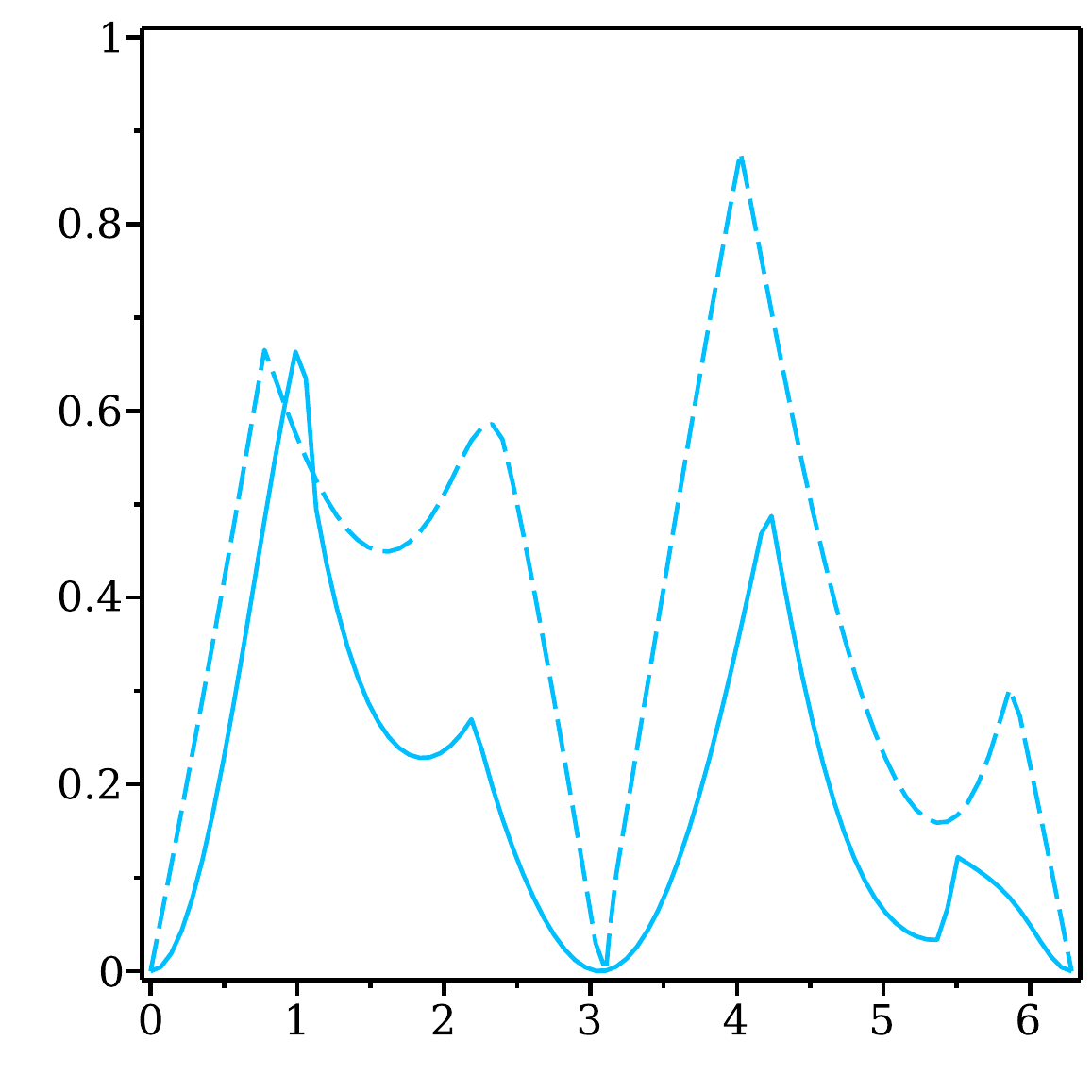}
  \begin{small}
   \put(85,-0.75){$\phi$ (rad)}
      \put (26,-6){(g) $G^{\rectangleblack}_{\vartriangle}(\mathbf{K'})$}
     \end{small}
     \end{overpic}
 \quad
  \begin{overpic}
  [height=46mm]{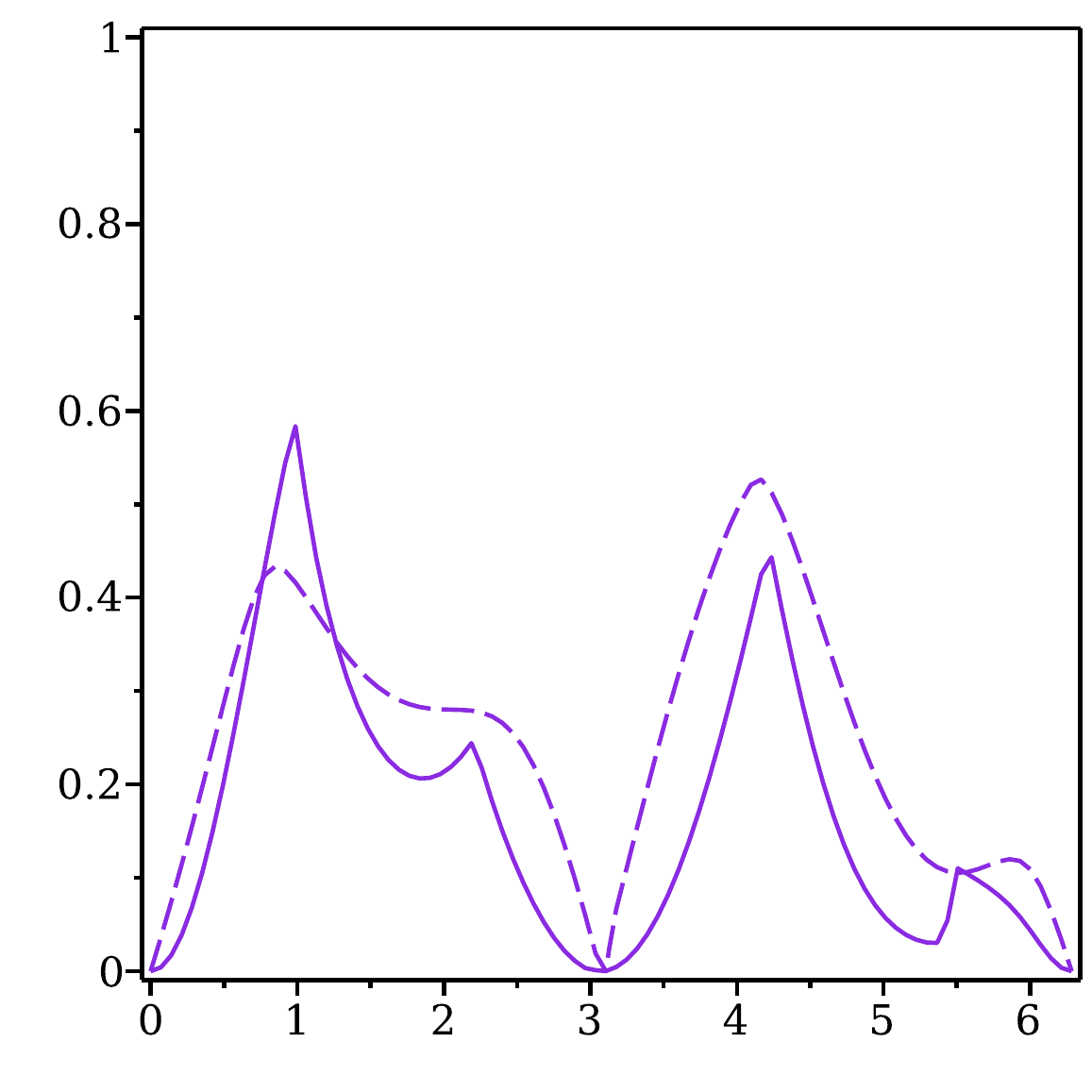}
  \begin{small}
   \put(85,-0.75){$\phi$ (rad)}
      \put (26,-6){(h) $G^{\blacktriangle}_{\vartriangle}(\mathbf{K'})$}
     \end{small}
     \end{overpic}
     \quad
\begin{overpic}
  [height=46mm]{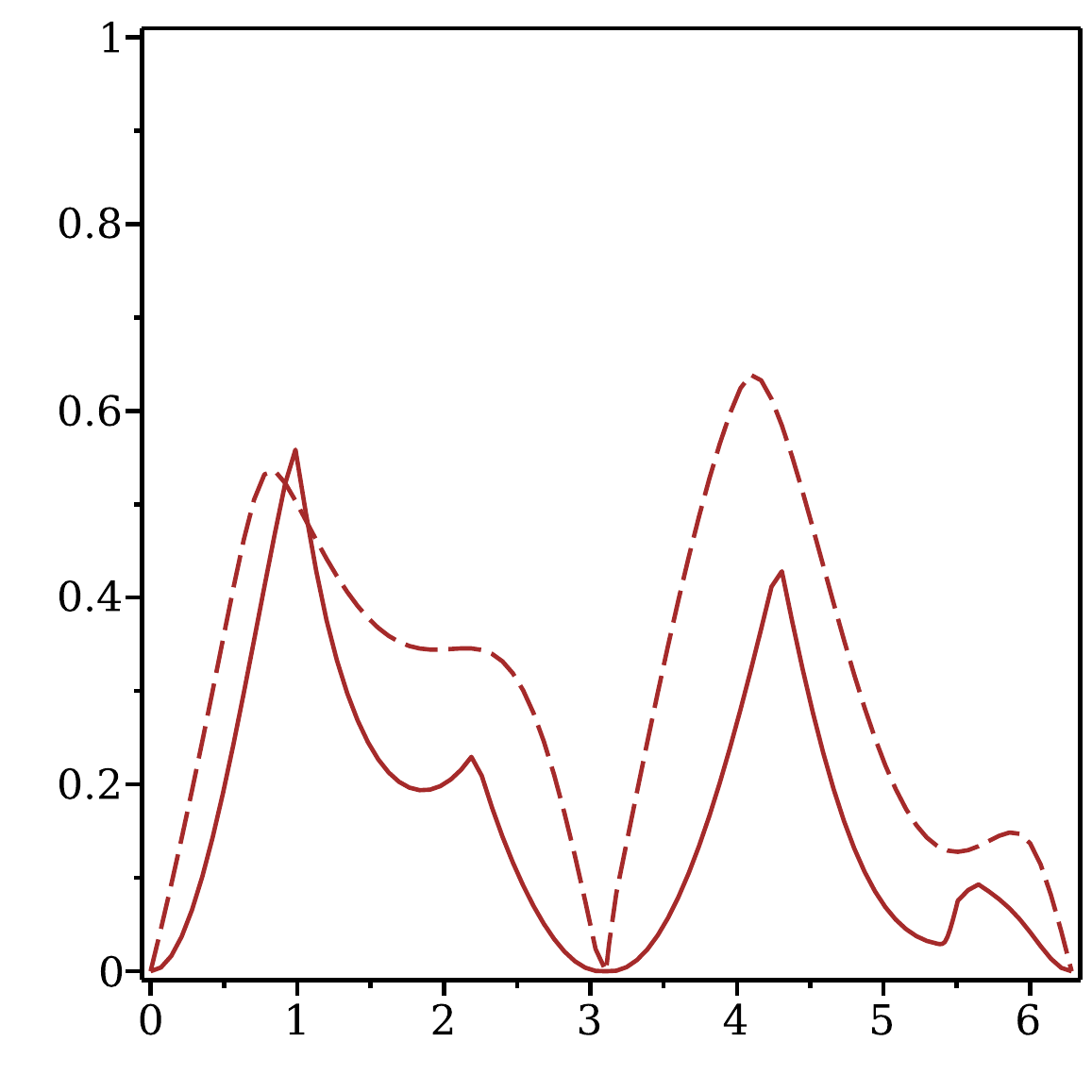}
   \begin{small}
   \put(85,-0.75){$\phi$ (rad)}
    \put (26,-6){(i) $G^{\vartriangle}_{\vartriangle}(\mathbf{K'})$}
     \end{small}
\end{overpic} \medskip     
\end{center}
\caption{Comparison of 
$s\sqrt{D_\star^\circ(\mathbf{K},\mathbf{K}')}$ 
 (solid line) and the corresponding extrinsic (dashed line) distance to the closest regular point on $V=0$.}
\label{fig:compare1}
\end{figure}

In the following we want to compare the computed intrinsic singularity-distances given in  Fig.~\ref{measure2} with the corresponding extrinsic ones presented in \cite{akapilavai2022}. To do so, we have to take care of the physical dimensions which are energy density in the intrinsic case (cf.\ Section \ref{sec:metrics}) and length for the extrinsic approach. In order to compare like and like we want to convert the energy density to a length which can be done using the following idea:

Let us assume that we have elastically deformed the manipulator from the given configuration into the closest singular one. Now we let the manipulator spring back, where the stored elastic energy is transformed into kinetic energy $U_K$ for which the formula $U_K=\tfrac{1}{2}mv^2$ holds true, where $m$ denotes the mass and $v$ the velocity. From that we can express 
$v=\sqrt{\tfrac{2}{d}}\sqrt{\tfrac{U_K}{V}}$
as the mass is computed by volume $V$ times density $d$. In this formula, we can substitute the energy density by our values  $D_\star^\circ(\mathbf{K},\mathbf{K}')$. 
In order to convert the velocity into a length we only have to multiply it with a factor of time $t$. Therefore, $s\sqrt{D_\star^\circ(\mathbf{K},\mathbf{K}')}$
can be interpreted as a length where 
$s:=t\sqrt{\tfrac{2}{d}}$ plays only the role of a scaling factor. For the graphs displayed in Fig.~\ref{fig:compare1} we have chosen $s=10\sqrt{2}$.

Surprisingly, from Fig.~\ref{fig:compare1}, we observe that the graphs exhibit similar trends, which was unexpected (apart from the behavior in the neighborhood of the two singular poses). They have nearly identical positions for the local maxima and local minima. Clearly, one can compute an upper bound for the intrinsic metric from the closest singular configuration with respect to the extrinsic metric, and vice versa, but this alone does not explain the similar courses of the graphs. 

\begin{rmk}
For a detailed comparison with singularity closeness indices along the same one-parametric motion we refer to
\cite[Fig.\ 15]{akapilavai2022}. 
\hfill $\diamond$    
\end{rmk}

\section{Conclusion and final remarks}\label{end}
We presented in Section \ref{terminology} intrinsic metrics for computing the closest singular configuration of a 3-RPR parallel manipulator. These metrics take into consideration various design options for the base and platform. In total, we end up with nine interpretations of a  3-RPR mechanism, where each one implies an intrinsic metric. 
To formulate the constrained optimization problem of finding the closest singular configuration, we employed the Lagrangian approach, as discussed in Section \ref{generalcase}. For the computation, we developed a computational algorithm in Section \ref{results}, utilizing the software \texttt{Maple} and \texttt{Bertini}, which can calculate the singularity distance along a one-parametric motion of the manipulator.  
To validate the proposed approach, a numerical example was presented in Section \ref{sec:example}. Additionally, we compared our method with the existing intrinsic singularity distances mentioned in the literature  (cf.\ Section \ref{zeincomparison})
Moreover, in Section \ref{eg:comparison}, we compared our intrinsic singularity distances with the corresponding extrinsic ones presented in \cite{akapilavai2022}. The observed relation between these distances is not fully understood and is open to future research.

There also remain open computational issues as we cannot guarantee the completeness of the solution set for some of the optimization problems of the ab-initio phase (cf.\ Tables \ref{rootcount} and \ref{collinearityresult}). They may be resolved by using isotropic coordinates, which facilitate the numerical solution of problems in planar kinematics \cite{wampler}. However, we abstain from trying this approach, as the discussion of the concrete example in Section \ref{sec:example} shows that all nine presented intrinsic metrics have a similar course (cf.\ Fig.\ \ref{measure2}). Therefore, we recommend to use $D_{\rectangleblack}^{\rectangleblack}(\mathbf{K},\mathbf{K}')$ of Eq.~(\ref{eq:lagrangeplus}) or the one given in \cite[Eq.\ (44)]{NAWRATIL2022104510} as they 
\begin{enumerate}[(a)]
    \item live in the joint space of the manipulator (like the metric of Zein et al. \cite{zein}),
    \item can be computed very efficiently (cf.\ Table \ref{rootcount}), and have the potential to be calculated in real-time using parallel computing\footnote{The average computational time (5runs) for tracking $76$ solutions given in Table~\ref{rootcount} for a single pose is $\approx 13$ secs.}.
\end{enumerate}
The latter is also of importance for the application mentioned in  \cite{zein} of planning singularity-free trajectories in the joint space. However, in contrast to the approach of \cite{zein}, our method can deal with the problem that arises from the fact that the mapping from the configuration space to the joint space is 6 to 1 (cf.\ Section \ref{sec:review}). 
The distances obtained with our method imply guaranteed singularity-free balls, including the contact point with the singularity variety (i.e.\ closest singular configuration). The latter information can be exploited to optimize a singularity-free trajectory between a start- and an end-pose by maximizing the intrinsic distance to the singularity variety. This can be achieved by utilizing similar ideas (variational path optimization) as presented in  \cite{RASOULZADEH2020104002}, but applied to the joint space instead of the configuration space.

According to \cite{zein}, another application of intrinsic metrics is the determination of the center/home configuration of the robot as the one that has the largest singularity distance within the joint space. Moreover, 
the maximization of the latter distance can be used as an objective for optimizing the design of the manipulator.

The usage of the paired intrinsic and extrinsic metrics, which were compared in Section \ref{eg:comparison}, at the level of velocities is also expected to contribute to
sensitivity analysis~\cite{caro2009sensitivity,goldsztejn2016three}, as in this way one can quantify the change in the shape of the manipulator implied by variations of the inner geometry. This topic is dedicated to future research.

Note that the presented methods and applications are not restricted to 3-RPR manipulators used for the proof of concept in the paper at hand, but can be applied to any mechanism/robot which can be considered as pin-jointed body-bar framework according to \cite{NAWRATIL2022104510}, where e.g.\ Stewart Gough platforms were discussed in Section 6.

%%%%%%%%%%%%%%%%%%%%
\section*{Acknowledgments}
This research is supported by Grant No.\ P 30855-N32 from the Austrian Science Fund (FWF). The first author would like to express gratitude to Silviana Amethyst from the Department of Mathematics at the University of Wisconsin Eau Claire and Professor Jonathan Hauenstein from the Department of Applied and Computational Mathematics and Statistics at the University of Notre Dame for their valuable suggestions and technical discussions on $\mathtt{Bertini}$ and $\mathtt{Paramotopy}$. Additionally, the first author would like to thank Paul Breiding for providing assistance in setting up polyhedral homotopy computations in \texttt{HC.jl} and for clarifying questions regarding B\'{e}zout bounds.
Furthermore, the first author would like to acknowledge Aravind Baskar from the University of Notre Dame for his assistance in sharing the algorithm for finding the best grouping for minimal B\'{e}zout bounds.

\bibliography{bibfile1}

\end{document}